\documentclass[runningheads]{llncs}

\usepackage{graphicx}
\usepackage{amsmath,amssymb} %
\usepackage{color}

\begin{document}

\title{End-to-End Deep Structured Models for Drawing Crosswalks}

\titlerunning{Structured Models for Drawing Crosswalks}

\author{Justin Liang\inst{1,2} \and Raquel Urtasun\inst{1,2}}

\authorrunning{J. Liang and R. Urtasun}

\institute{Uber Advanced Technologies Group \and University of Toronto \\
\small\texttt{justin.j.w.liang@gmail.com, urtasun@cs.toronto.edu}
}

\maketitle              

\begin{abstract}
In this paper we address the problem of detecting crosswalks from LiDAR and camera imagery. Towards this goal, given multiple LiDAR sweeps and the corresponding imagery, we project both inputs onto the ground surface to produce a top down view of the scene. We then leverage convolutional neural networks to extract semantic cues about the location of the crosswalks. These are then used in combination with road centerlines from freely available maps (e.g., OpenStreetMaps) to solve a structured optimization problem which draws the final crosswalk boundaries. Our experiments over  crosswalks in a large city area show that 96.6\% automation can be achieved. 

\end{abstract}
\section{Introduction}
\label{introduction}

\begin{figure}[t]
\vskip 0.2in
\begin{center}
\centerline{\includegraphics[width=\columnwidth]{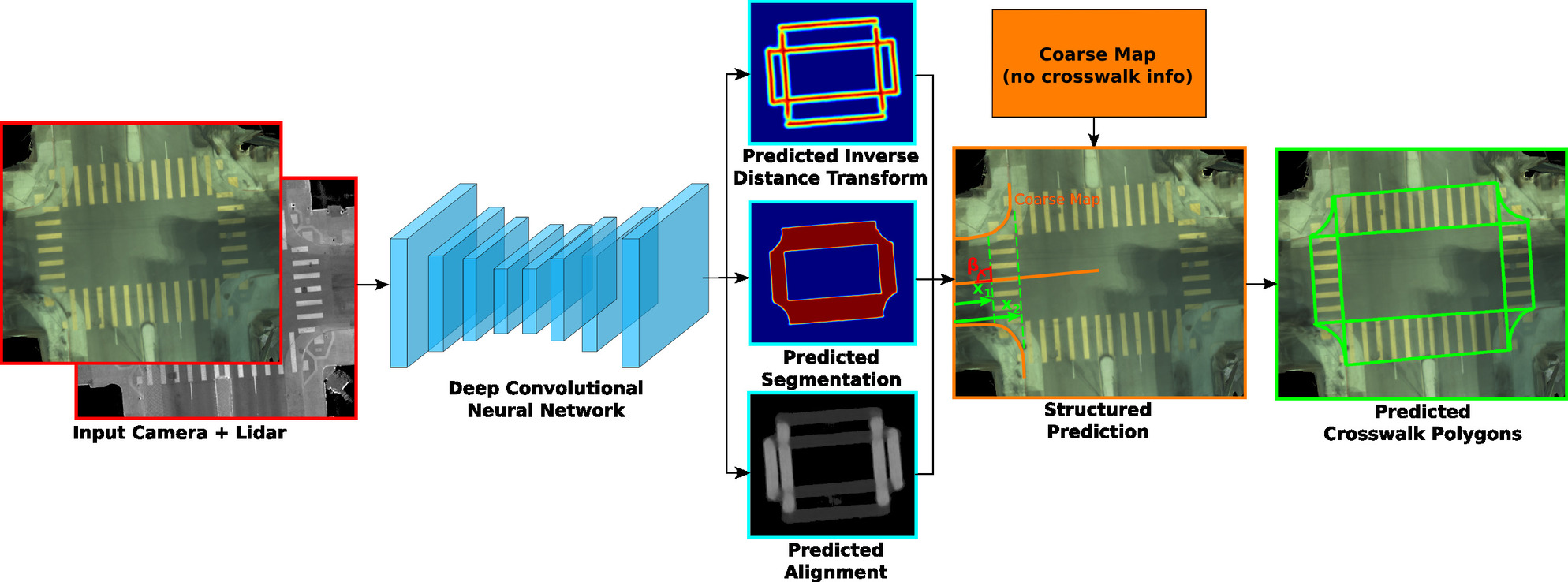}}
\vspace{-2mm}
\caption{Overview of our model. LiDAR points and camera images are projected onto the ground to produce input images from bird's eye view. These are then fed into a convolutional neural network (CNN) to produce three feature maps. Next, we perform inference using the three feature maps along with a coarse map which provides the road centrelines and intersection polygons. This is fed into the structured prediction module that finds the best two boundaries $x_1$ and $x_2$ along with the best angle $\beta$ by maximizing a structured energy function. }
\label{fig:overview}
\end{center}
\vskip -0.2in
\vspace{-5mm}
\end{figure} 

Autonomous vehicles have many potential benefits. Every year, more than 1.2 million people die in traffic accidents. Furthermore, accidents are caused by human factors (e.g., driver distraction) in 96\% of the cases. Urban congestion is also changing the landscapes of our cities, where more than 20\% of the land is typically dedicated to parking. 
In recent years, tremendous progress has been made in the field of autonomous vehicles. This is the result of 
major breakthroughs in artificial intelligence, hardware (e.g., sensors, specialized compute) as well as heroic engineering efforts.  

Most autonomous driving teams in both industry and academia utilize detailed annotated maps of the environment to drive safely. These maps capture the static information about the scene. As a consequence, very strong prior knowledge can be used to aid perception, prediction and motion planning when the autonomous vehicle is accurately localized. 
Building such maps is, however, an extremely difficult task. 
High definition (HD) maps typically contain both geometric and semantic information about the scene. SLAM-based approaches are typically employed to create dense point cloud representations of the world, while human labellers are used to draw the semantic components of the scene such as lanes, roads, intersections, crosswalks, traffic lights, traffic signs, etc. 

Most map automation efforts focus on automatically estimating the lanes \cite{lane_est1,lane_est2,lane_est3,lane_est4,lane_est5,gellert2}. Approaches based on cameras \cite{lane_est1,lane_est2}, LiDAR \cite{lane_est4,lane_est5} as well as aerial images \cite{gellert2,lane_est6} have been proposed. 
On the other hand, very little to no attention has been paid to other semantic elements.

In this paper, we tackle the problem of accurately drawing crosswalks. Knowledge about where they are is vital for navigation, as it allows the autonomous vehicle to plan ahead and be cautious of potential pedestrians crossing the street. 
Existing approaches focus on predicting the existence of a crosswalk, but do not provide an accurate localization.  Instead, crosswalks are typically crowd-sourced and manually drawn.

Drawing crosswalks is not an easy task.
As shown in our experiments, crosswalks come in a variety of shapes and styles even within the same city. Furthermore, paint quality of the crosswalk markings can often be washed out, making the task hard even for humans.  Framing the task as semantic segmentation or object detection does not provide the level of reliability that is necessary for autonomous driving. Instead a more structured representation is required.  

In this paper, we propose to take advantage of road centerlines and intersection polygons that are typically available in publicly available maps such as the OpenStreetMap (OSM) project. %
This allows us to parameterize the problem in a structured way, where our crosswalks have the right topology and shape. 
Towards this goal, we derive a deep structured model that is able to produce accurate estimates and exploit multiple sensors such as LiDAR and cameras.  
In particular, we use a convolutional net to predict semantic segmentation, semantic edge information, as well as crosswalk  directions. %
These outputs are then used to form a structured prediction problem, whose inference results are our final crosswalk drawings. By leveraging distance transforms and integral accumulators, efficient exact inference is possible.

We demonstrate the effectiveness of our approach in a variety of scenarios, where LiDAR and/or cameras are exploited to build bird's eye view representations of the road on which our model operates.
Our approach shows that 96.6\% automation is possible when building maps offline and 91.5\% when building the map online (as we drive). For comparison, human disagreement is around 0.6\%.

\section{Related Work}
\label{related-work}

\subsubsection{Crosswalk Detection:}

In \cite{mobile-cw,mobile-cw2,streetlevel-cw,zebrac,zebrarecog}, methods were developed to detect crosswalks at the street level. Moreover, \cite{aerial-cw} proposes a model for crosswalk detection in aerial images. However, these methods employ manually created feature extraction techniques, and can only handle zebra-style crosswalks. More recent methods have used deep convolutional neural networks (CNNs) to detect crosswalks. For example, the authors of \cite{aerial-cw2} use deep CNNs to detect the crosswalks in aerial imagery. However, they do no draw the crosswalks. Instead, they only produce the locations of the crosswalks. Similarly, the authors of \cite{cw-classify} use deep CNNs to detect crosswalks in satellite imagery, but only predicts whether or not a crosswalk exists in the image. Crosswalk detection is performed for driver assistance systems in \cite{cw-driving}. In this paper, they draw the crosswalk in front of a vehicle. However, the method is limited in the sense that there is a maximum distance in which a crosswalk can be detected. Furthermore, the method only works on camera imagery taken at the vehicle level.

\subsubsection{Automatic Mapping:}
There are many methods used to automatically generate different elements of a map. For example, the automatic extraction and segmentation of roads has been tackled in \cite{gellert1,gellert2,gellert3,WegnerMS13} using techniques such as Markov random fields and deep CNNs. In \cite{building_recon1,building_recon2}, they use LiDAR data in combination with aerial images and/or building address points to perform building shape reconstruction. Reconstruction of both the 2D footprints and the 3D shape of the buildings is tackled in these papers. Recently, the TorontoCity dataset \cite{tcity} was released, and provides a multitude of map related benchmarks such as building footprints reconstruction, road centerline and curb extraction, road segmentation and urban zoning classification. In \cite{semantic-map}, a bird's eye view semantic map is produced from multi-view street-level imagery. Here, they perform semantic segmentation on street-level imagery and project this onto the ground plane in overhead view. In \cite{robocodes}, they develop a generative algorithm to automatically label regions, roads and blocks with street addresses by extracting relevant features in satellite imagery using deep learning. Many mapping methods have utilized LiDAR data to perform automatic mapping. Examples of this can be seen in \cite{3dlidar,3dpavement,3dlidar2,semantic3d,3dlidar3}. In these papers, they utilize LiDAR data to create 3D models of cities, automatically extract pavement markings, and perform semantic segmentation on urban maps to classify features.

\subsubsection{Semantic Segmentation:}
In semantic segmentation, the goal is to label every pixel in an image with a class. Methods involving recurrent neural networks (RNNs) have been proposed \cite{rnnseg1,rnnseg2}, however, the RNNs themselves can be computationally expensive to run. In \cite{fully-conv}, the authors introduced fully convolutional networks (FCNs) which uses skip connections to combine semantic information from feature volumes of various spatial dimensions within the CNN. It utilizes bilinear upsampling to perform semantic segmentation. After this, many variants of FCNs were released. For example, in \cite{crfseg}, a deep deconvolutional network followed by a conditional random field (CRF) were used to fine-tune the output segmentation. Similarly, \cite{residual-learning} builds upon this idea and uses a deeper network with residual layers and shortcut connections to learn an identity mapping. \cite{feature-pyramid,linknet,segnet1,segnet2,segnet3} further expands on these concepts, and use an encoder-decoder network with skip connections. This encoder-decoder architecture is inherently represented as a pyramid which produces a multi-scale feature representation. Since the representation is inherent to the shape of a CNN, inference is less memory and computationally expensive. In \cite{dilated-conv}, they introduce dilated convolutions to aggregate multi-scale contextual information. They show with their method they can expand the receptive field with no loss in resolution and coverage.
\section{Deep Structured Models for Mapping Crosswalks}
\label{method}

High definition (HD) maps typically contain both geometric and semantic information about the scene. SLAM-based approaches are typically utilized to create dense point cloud representations of the world, while human labellers are typically employed to draw the semantic components of the scene, e.g., lanes, crosswalks, rules at intersections. 
In this paper, we focus on automatically drawing crosswalks. Towards this goal, we derived a deep structured model that is able to produce accurate estimates and exploit multiple sensors such as LiDAR and cameras.  
In particular, we exploit a CNN to predict semantic segmentation, semantic edge information as well as crosswalk  directions. %
These outputs are then used to form a structured prediction problem, whose inference results are our final crosswalk drawings. By leveraging distance transforms and integral accumulators, efficient exact inference is possible. 
In the remainder of the section, we first define our convolutional potentials, follow by our structured prediction framework.

\subsection{Computing Deep Semantic Features}
\label{prediction}

We leverage both images and LiDAR to automatically draw crosswalks. Towards this goal, for each sensor modality we create an overhead view of each intersection. 
We refer the reader to Fig. \ref{fig:overview} for an example of the overhead representation of both LiDAR as well as images. 
Note that determining where an intersection happens is trivial given the topological graphs of existing freely available coarse maps, such as OpenStreetMaps.

Both LiDAR and Camera overhead images are then concatenated to create our input representation of the scene. This forms a 4-channel input, with 3 dimensions for RGB and one for LiDAR intensity. 
This 4-channel image is then fed to a multi-task CNN that is trained to produce semantic segmentation, semantic contour detection as well as angles defining the crosswalk direction. 
In particular,  the first output feature map is a pixel-wise foreground/background segmentation of the crosswalks. The second output  map is an inverse distance transform from the boundaries of the crosswalks thresholded at a value of 30 pixels (i.e., 1.2m). By predicting an inverse distance transform, the network learns about the relative distance to the boundaries which makes learning more effective as it contains more supervision than simply predicting the location of the edge. The third output feature map encodes the angles of each crosswalk boundary dilated to a diameter of 30 pixels. We encode this with two outputs per pixel, which correspond to the x and y components of the directional unit vector of the angle. %
Thus, simply taking the arc tangent of this would produce the predicted angle. 

\subsubsection{Network Architecture: } 
\begin{figure*}[t]
\begin{center}
\includegraphics[width=\textwidth]{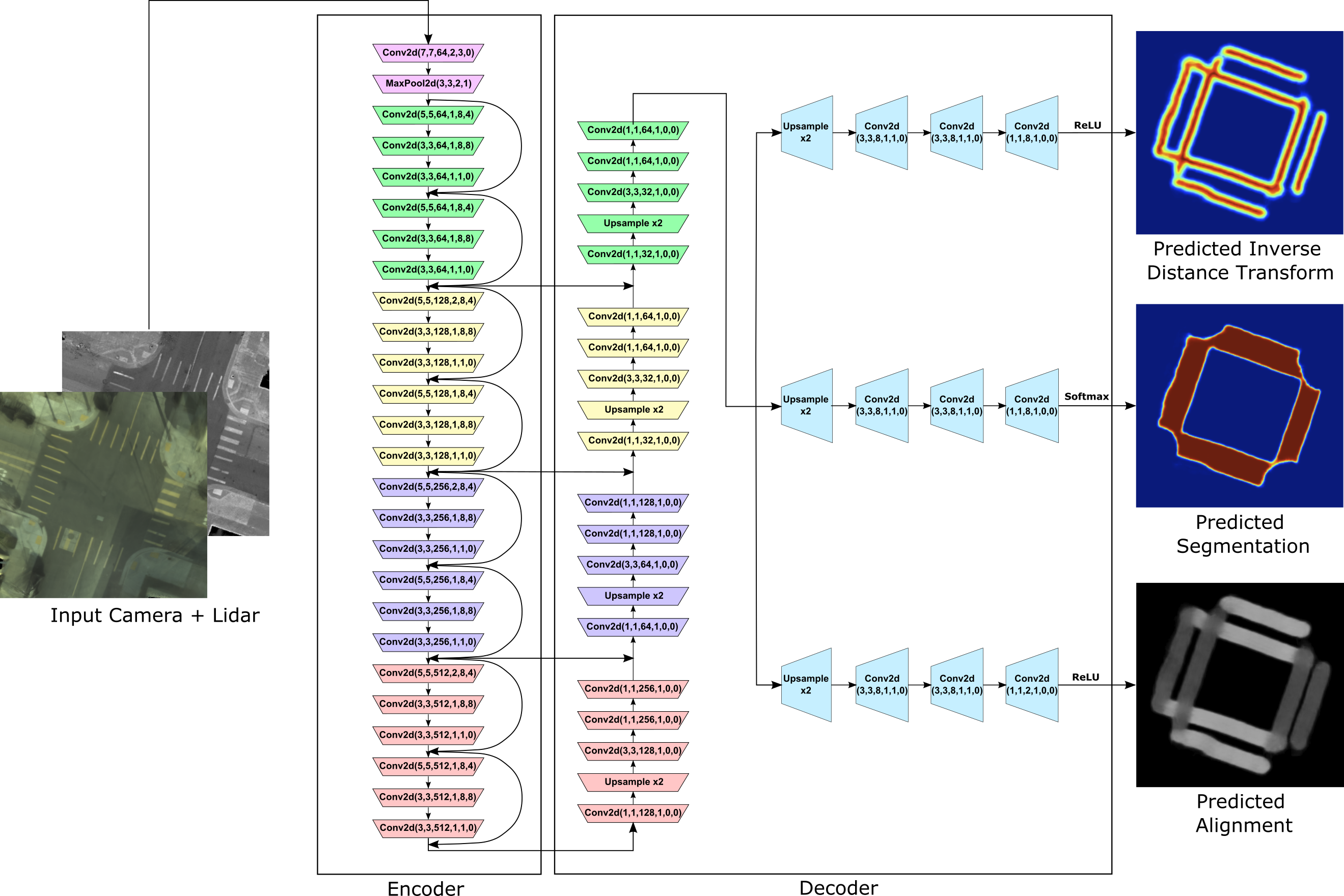} 
\vspace{-2mm}
\caption{Overview of our prediction network. Here we use MaxPool2d(kernel width, kernel height, stride, padding) and Conv2d(kernel  width, kernel height, out channels, stride, padding, dilation).}
\label{fig:model}
\end{center}
\vspace{-7mm}
\end{figure*}

We use an encoder-decoder architecture with skip connections and residual layers based on the feature pyramids networks in \cite{residual-learning,feature-pyramid,linknet} to output the three feature maps. We refer the reader to Fig.  \ref{fig:model} for a detail visualization of our network architecture. %
Note that before each convolutional layer we use batch normalization \cite{batchnorm} followed by a ReLU non-linearity \cite{relu}. In the encoder network, each residual block consists of three convolutional layers. Because the images can be quite large, we need to ensure the network has a large receptive field, thus, we leverage dilated convolutions \cite{dilated-conv} in the residual blocks. In the decoder network, we perform nearest neighbor upsampling to upsample back to the original image size. We then split the output into three branches, one for each feature map. To predict the inverse distance transform, we apply a ReLU non-linearity at the end to restrict the output to a positive value. To predict the segmentation, we apply a softmax over the output to get a probability map. To predict the alignment we apply a ReLU non-linearity to restrict the output to a positive value. 

\subsubsection{Learning:} We treat the distance transform and angle predictions as regression and the segmentation as pixel-wise classification tasks. 
To train our network, we minimize the sum of losses over the three prediction tasks:
\begin{equation}
  l(\mathcal{I}) = l_{seg}(\mathcal{I}) + l_{dt}(\mathcal{I}) + \lambda_\ell l_{a}(\mathcal{I})
\end{equation}
where $\lambda_\ell$ is a weighting for the alignment loss. In practice we use  $\lambda_\ell = 100$ which we found through cross-validation.
We define the segmentation loss $l_{seg}$ to be binary cross entropy:
\begin{eqnarray}
  l_{seg}(\mathcal{I}) & = & \frac{1}{N} \sum_{p } \left( \hat{y}_{p} log(y_{p}) + \right. 
  \left. (1 - \hat{y}_{p}) log(1 - y_{p}) \right)
\end{eqnarray}
where $N$  is the number of pixels in the bird's eye view image, $\hat{y}_{p}$ represents the ground truth pixel $p$'s value and $y_{p}$ represents the predicted probability of the $p$ being a crosswalk.

We define the boundary loss $I_{dt}$ to be the mean squared loss:
\begin{equation}
  l_{dt}(\mathcal{I}) = \frac{1}{N} \sum_{p}  || d_{p} - \hat{d}_{p} ||^2
\end{equation}
where $d_p$ is pixel $p$'s value in the inverse distance transform feature map $\phi_{dt}$.

Finally, we define the alignment loss $I_{a}$ as the mean squared loss:
\begin{equation}
  l_{a}(\mathcal{I}) = \frac{1}{N}  \sum_{p } || \textrm{atan} \left(\frac{v_{p, y}}{v_{p, x}}\right) - \hat{\alpha}_{p} ||^2
\end{equation}
where $v_{p, y}$ and $v_{p, x}$ are the y and x components of the unit vector corresponding to the predicted angle, and $\alpha_{p, gt}$ is the ground truth angle. Since a single crosswalk boundary can be represented with multiple angles, we restrict our output to be between $(0, \pi)$.

\subsection{Structured Prediction}
\label{inference}
During inference, we seek to draw the polygon that define each crosswalk. Our approach takes as input the road centerlines, the intersection polygon, as well as the three feature maps predicted by our multi-task convolutional network. Inspired by how humans draw crosswalks, we frame the problem as a 2D search along each centerline to find the two points  that describe the boundaries of the crosswalk. This structured representation of the crosswalk drawing problem allows us to produce output estimates that are as good as human annotations. 

We use the angle prediction to define a set of candidate hypothesis including the road centerline angle, the mode of the prediction  as well as $\pm 2^{\circ}$ and $\pm 5^{\circ}$ angles around that prediction. 
We then formulate the problem as an energy maximization problem, where potentials encode the agreement with the segmentation and boundary semantic features. Here, the inverse distance transform values are maximum right on the boundary, thus, our formulation will favor predicted boundaries that are right on it. The segmentation potential is used to ensure the two boundaries maximize the number of crosswalk pixels inside and maximize the number of background pixels outside. Our energy maximization formulation is below:
\begin{equation}
 \max_{x_1, x_2, \beta}  \lambda_I (\vec{\phi}_{seg, \vec{\ell}, \beta}(x_2) - \vec{\phi}_{seg, \vec{\ell}, \beta}(x_1)) + \left. (1-\lambda_I)(\vec{\phi}_{dt, \vec{\ell}, \beta}(x_2) + \vec{\phi}_{dt, \vec{\ell}, \beta}(x_1) \right)
\end{equation}

where  $\vec{\phi}_{seg}$ and $\vec{\phi}_{dt}$ are the output feature maps of the segmentation and semantic edge tasks. $x_1$ and $x_2$ are the two points on the centreline that define the crosswalk. $\beta$ is the boundary angle. $\lambda_I$ is the weighting used to balance between the segmentation and semantic edge feature maps. $\vec{\ell}$ is the road centreline. 
Exhaustive search can be computed very efficiently by using non-axis align integral accumulators. In particular we can convert the $\vec{\phi}_{seg}$ to  a 1D integral image along the road centreline which allows us to easily calculate the number of enclosed crosswalk pixels inside the boundaries defined by $x_1$ and $x_2$.

\section{Experimental Evaluation}
\label{experimental-evaluation}

\begin{table}[!t]
\centering
  \begin{tabular}{|l|*{12}{c|}}
  \cline{5-13}
  \multicolumn{1}{c}{} &  \multicolumn{3}{c|}{}& \multicolumn{4}{|c}{Precision at (cm)} & \multicolumn{4}{|c|}{Recall at (cm)} & \multicolumn{1}{|c|}{IOU} \\ 
  \cline{2-13}
  \multicolumn{1}{c|}{} & N & C & L&
\multicolumn{1}{|c}{20} & \multicolumn{1}{|c}{40} & \multicolumn{1}{|c}{60} & \multicolumn{1}{|c}{80} & \multicolumn{1}{|c}{20} & \multicolumn{1}{|c}{40} &   \multicolumn{1}{|c}{60} & \multicolumn{1}{|c}{80} & \multicolumn{1}{|c|}{} \\ 
  \hline 
  
  \textrm{NN} & {Mult} & {\checkmark} & {\checkmark}
    &$21.4\%$ &$24.8\%$ &$25.2\%$  &$25.4\%$ 
    &$19.4\%$ &$22.3\%$ &$22.7\%$ &$43.1\%$ 
    &$35.9\%$ \\      %
  \textrm{Seg} & {Mult} & {\checkmark} & {\checkmark}
    &$80.1\%$ &$93.1\%$ &$94.5\%$  &$95.0\%$ 
    &$77.1\%$ &$91.9\%$ &$95.2\%$ &$97.1\%$ 
    &$88.7 \%$ \\      %
  \hline
    \textrm{Ours} & {1} & {\checkmark} & {-}
    &$78.8\%$ &$91.2\%$ &$93.8\%$  &$94.9\%$ 
    &$78.6\%$ &$90.5\%$ &$92.9\%$ &$93.8\%$ 
    &$86.9 \%$ \\      
  \textrm{Ours} & {1} & {-} & {\checkmark}
  &$77.2\%$ &$90.6\%$ &$93.1\%$  &$94.1\%$ 
    &$76.8\%$ &$89.7\%$ &$91.9\%$ &$92.8\%$ 
    &$85.7 \%$ \\      
  \textrm{Ours} & {1} & {\checkmark} & {\checkmark}
  &$79.8\%$ &$91.5\%$ &$93.6\%$  &$94.6\%$ 
    &$79.9\%$ &$91.3\%$ &$93.2\%$  &$93.9\%$ 
    &$87.1 \%$ \\       
  \hline
  \textrm{Ours} & {Mult} & {\checkmark} & {-}
  &$83.4\%$ &$94.9\%$ &$96.6\%$  &$97.3\%$ 
    &$83.3\%$ &$94.6\%$ &$96.2\%$ &$96.8\%$ 
    &$90.2 \%$ \\ 
  \textrm{Ours} & {Mult} & {-} & {\checkmark}
  &$84.5\%$ &$95.8\%$ &$97.6\%$  &$98.4\%$ 
    &$85.0\%$ &$96.1\%$ &$97.8\%$ &$98.3\%$ 
    &$91.8 \%$ \\   
  \textrm{Ours} & {Mult} & {\checkmark} & {\checkmark}
  &$\textbf{85.6\%}$ &$\textbf{96.6\%}$ &$\textbf{98.1\%}$  &$\textbf{98.8\%}$ 
    &$\textbf{86.1\%}$ &$\textbf{96.8\%}$ &$\textbf{98.2\%}$ &$\textbf{98.7\%}$ 
    &$\textbf{92.4\%}$ \\ 
  \hline
  \textrm{Human} & {-} & {-} & {-}
  &$88.3\%$ &$99.4\%$ &$99.7\%$  &$99.8\%$ 
    &$87.3\%$ &$98.3\%$ &$98.8\%$ &$98.8\%$ 
    &$95.3 \%$  \\ 
    
  \hline 
    
  \end{tabular}
  \caption{This table shows the performance of our model using various inputs. We use the columns N, C and L to denote the Number of passes, camera input and LiDAR input. Here, (Mult) denotes multiple car passes over for offline mapping and (1) denotes a single car pass for online mapping. The first baseline (NN) is a nearest neighbor algorithm on top of VGG features.  The second baseline (Seg) is the segmentation output from the model trained on multiple passes of the ground camera and LiDAR. Furthermore, we annotate 100 intersections ourselves and compare these results with the ground truth human annotation.}
  \label{tab:results}
  \vspace{-10mm}
\end{table}

\subsubsection{Dataset:}
We collected a large dataset in a North American city and use all crosswalks in this city with an area of $100km^2$. In total, 9502 km were driven to create this dataset. Our dataset consists of 1571 training images, 411 validation images and 607 test images. In total, there are 2559 intersections with 8526 crosswalks. This results in 5203 training, 1412 validation and 1911 test crosswalks. Each image represents an intersection with at least one crosswalk, and has a resolution of 4cm per pixel. 

\subsubsection{Metrics:}
We use precision and recall as our base metrics. For precision, the true positive equals the set of predicted crosswalks with a minimal distance smaller than $\tau$ and $TP + FP = |P|$. For recall, the true positive equals the set of ground truth crosswalks with minimal distance smaller than $\tau$ and $TP + FN = |G|$. We evaluate precision and recall at a $\tau$ of 20cm, 40cm, 60cm and 80cm. 
We also calculate the Intersection over Union (IoU) of the drawn crosswalks and the ground truth.

\subsubsection{Experimental Setup:}
We trained our models using a batch size of 1 and ADAM \cite{adam} with a learning rate of 1e-4 and a weight decay of 5e-4. We decrease the learning rate by a factor of 10 every 100000 training iterations. We then perform data augmentation when training by randomly flipping and rotating the images. The models are trained for 110 epochs over the entire training set. 

\subsubsection{Importance of Sensor Modality:} 
\begin{figure}[t]
\vskip 0.2in
\begin{center}
\centerline{\includegraphics[width=0.8\textwidth]{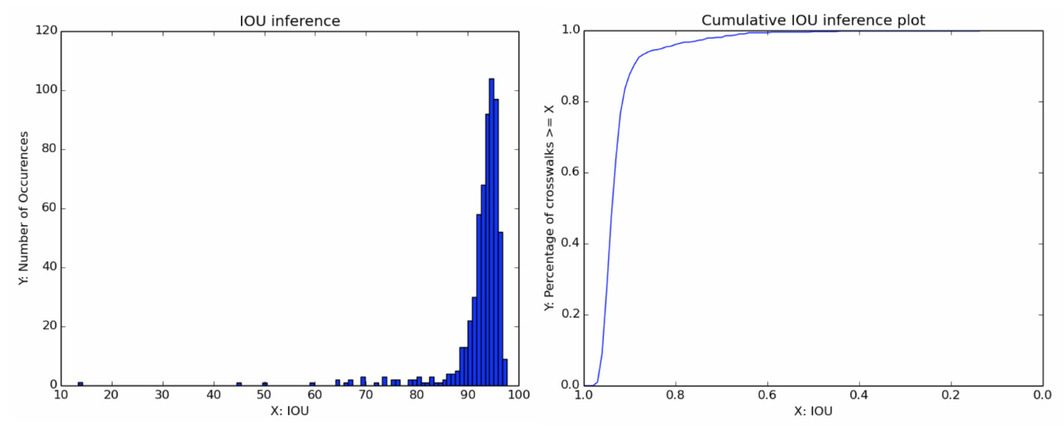}}
\vspace{-2mm}
\caption{A histogram of the IOUs (left) and cumulative IOU graph (right) using the LiDAR and camera as input.}
\label{fig:hist_cumul}\end{center}
\vskip -0.2in
\vspace{-5mm}
\end{figure}

We trained different models to use camera only, LiDAR only or a combination of both sensors. As shown in Table \ref{tab:results} using both sensors results in better performance. Note that the sensor type is encoded under C (camera) and L (LiDAR) in the table. Furthermore, a histogram of the IoUs using both LiDAR and camera images as input can be seen in Fig.  \ref{fig:hist_cumul} (left). We find that 94.1\% of the images have an IoU greater than 85.0\%.

\subsubsection{Online vs Offline maps:} 
\begin{figure}[t]
\centering
\begin{tabular}{cccccccr}
\includegraphics[width=.155\textwidth]{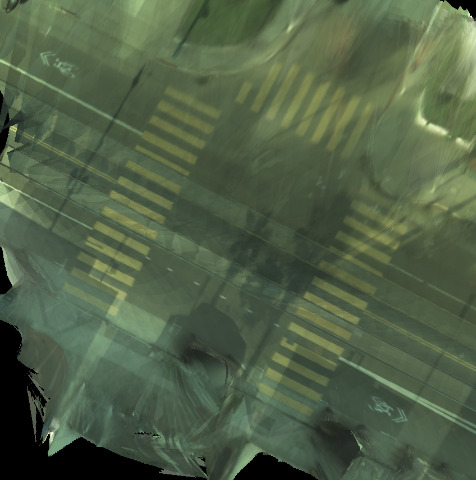} &
\includegraphics[width=.155\textwidth]{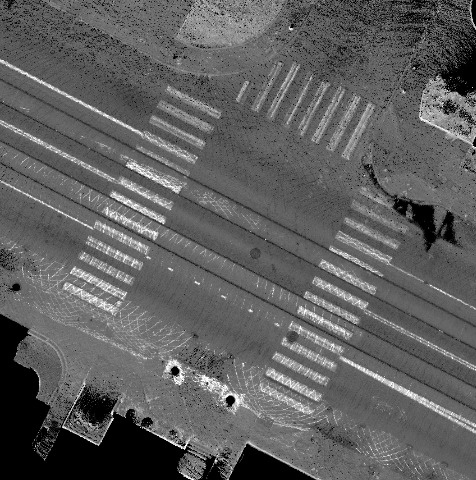} &
\includegraphics[width=.155\textwidth]{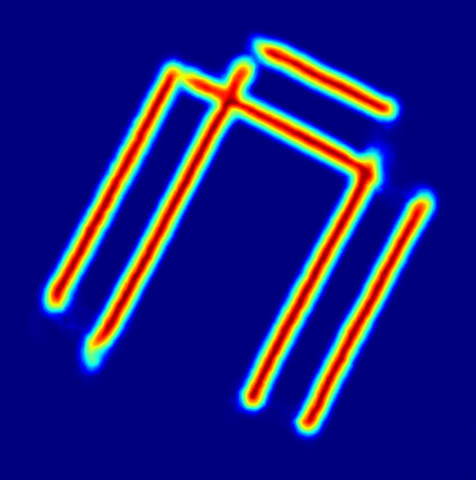} &
\includegraphics[width=.155\textwidth]{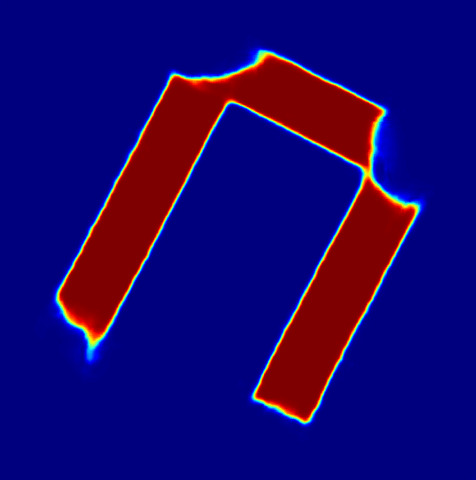} &
\includegraphics[width=.155\textwidth]{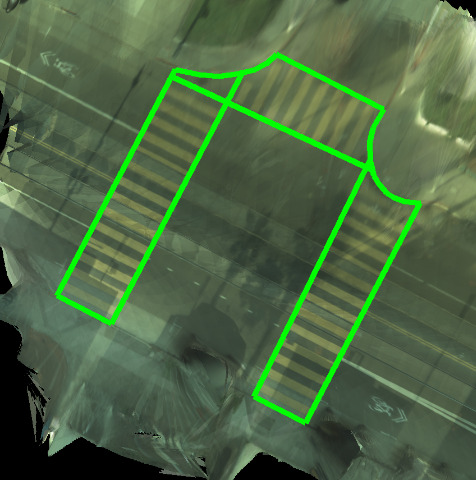} &
\includegraphics[width=.155\textwidth]{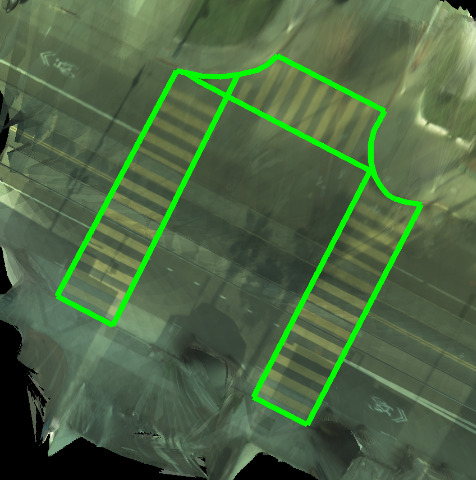} &\\

\includegraphics[width=.155\textwidth]{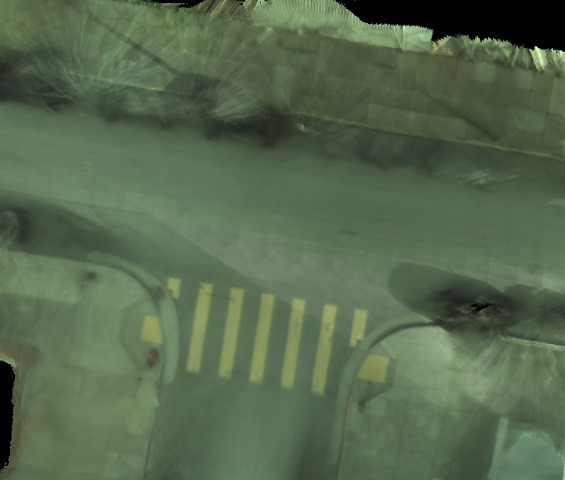} &
\includegraphics[width=.155\textwidth]{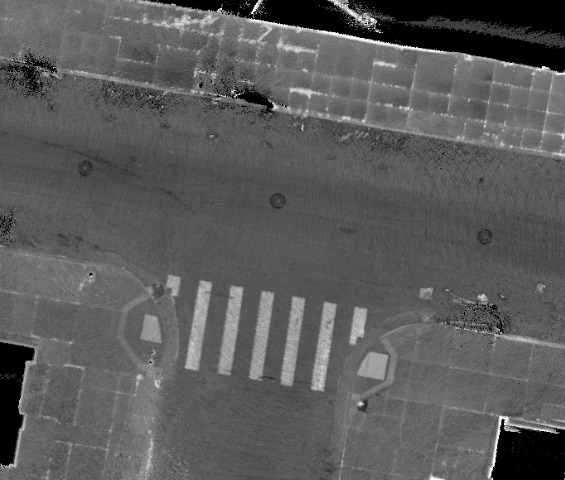} &
\includegraphics[width=.155\textwidth]{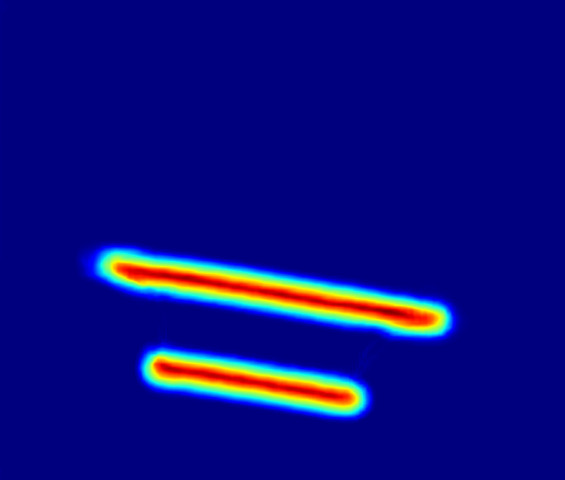} &
\includegraphics[width=.155\textwidth]{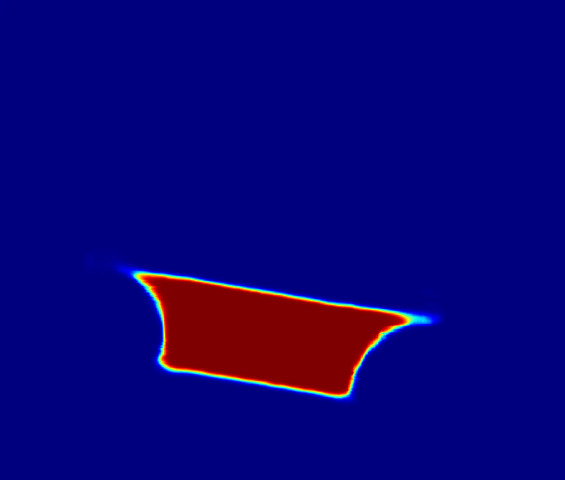} &
\includegraphics[width=.155\textwidth]{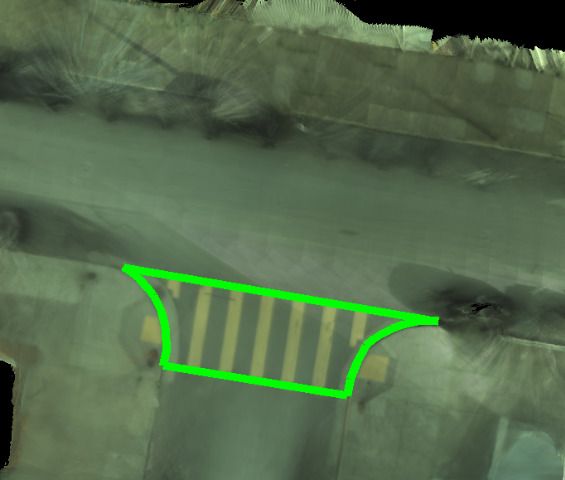} &
\includegraphics[width=.155\textwidth]{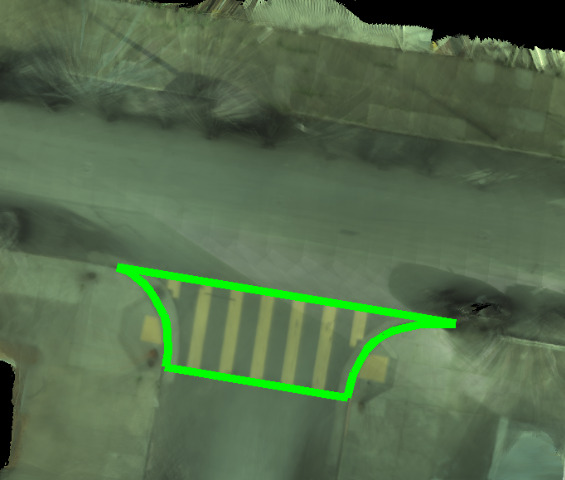} &\\

\includegraphics[width=.155\textwidth]{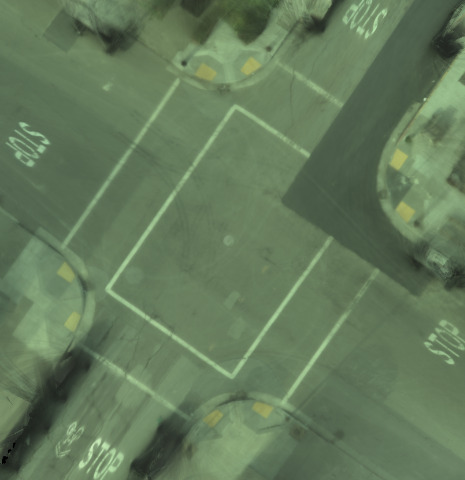} &
\includegraphics[width=.155\textwidth]{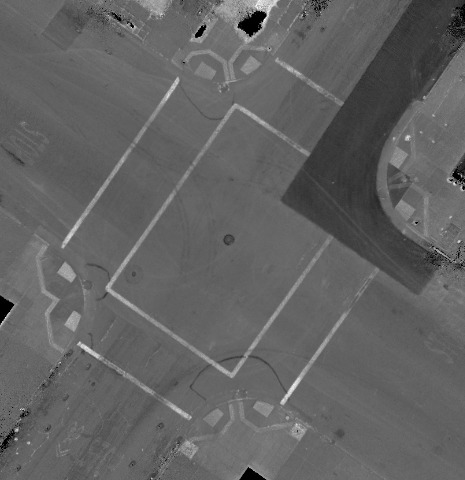} &
\includegraphics[width=.155\textwidth]{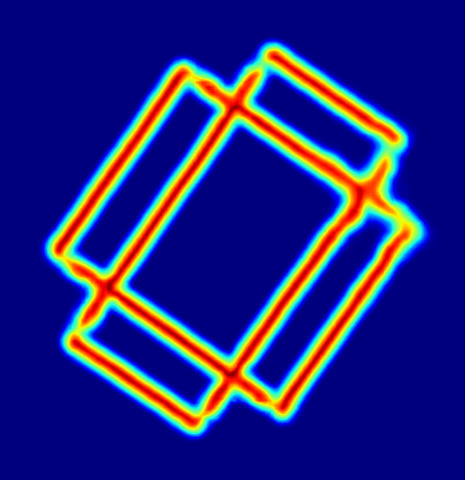} &
\includegraphics[width=.155\textwidth]{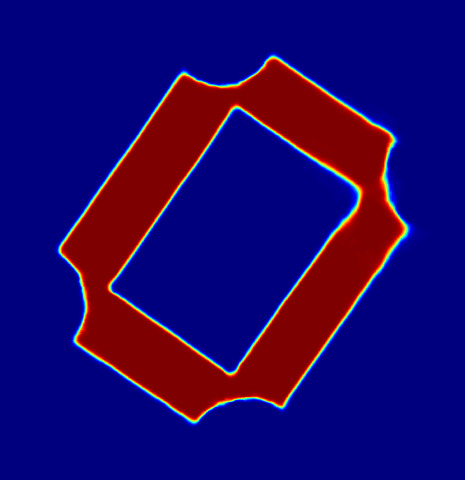} &
\includegraphics[width=.155\textwidth]{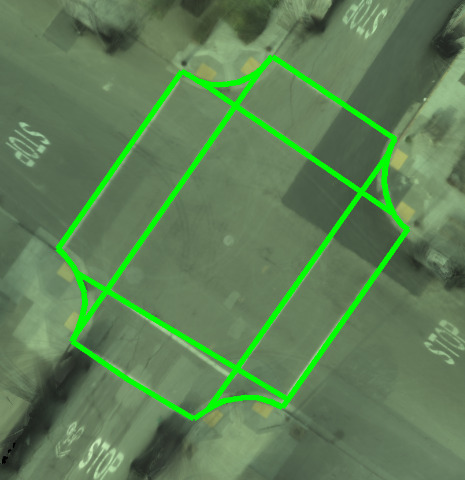} &
\includegraphics[width=.155\textwidth]{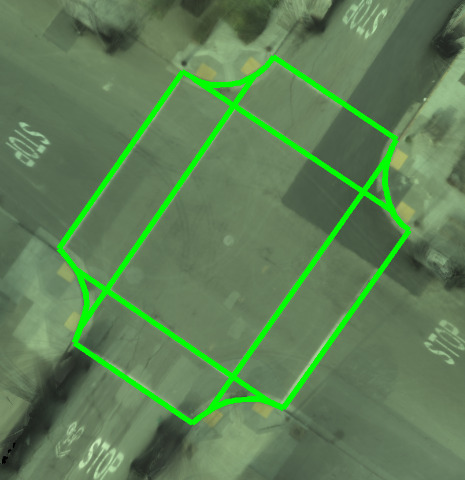} &\\

\includegraphics[width=.155\textwidth]{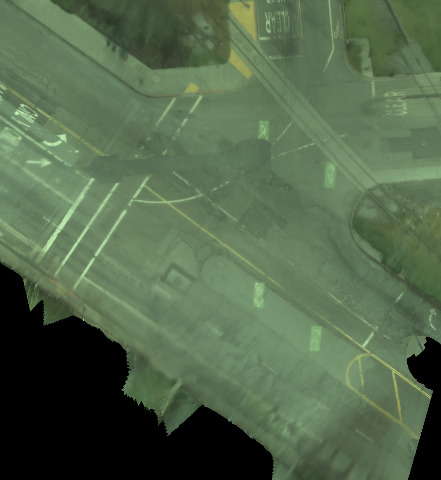} &
\includegraphics[width=.155\textwidth]{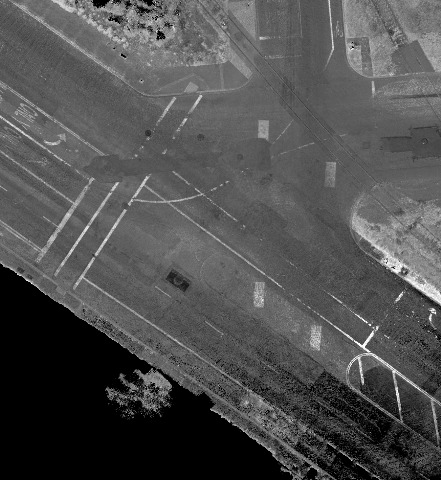} &
\includegraphics[width=.155\textwidth]{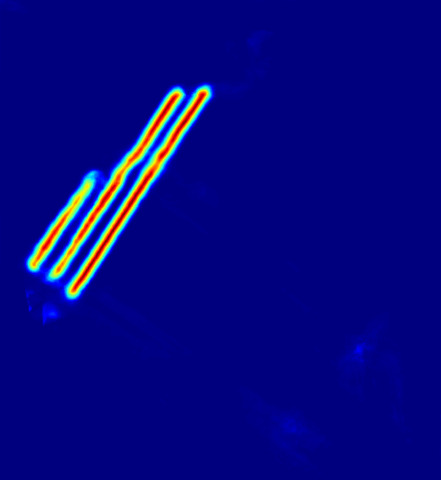} &
\includegraphics[width=.155\textwidth]{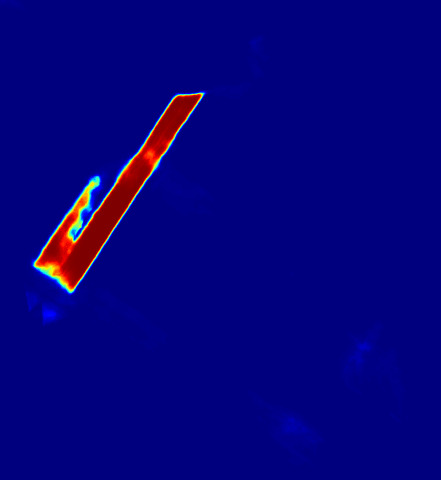} &
\includegraphics[width=.155\textwidth]{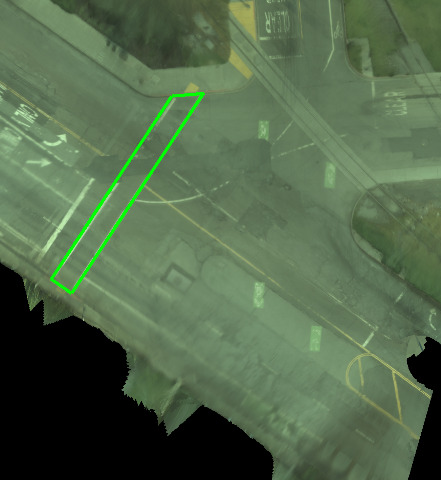} &
\includegraphics[width=.155\textwidth]{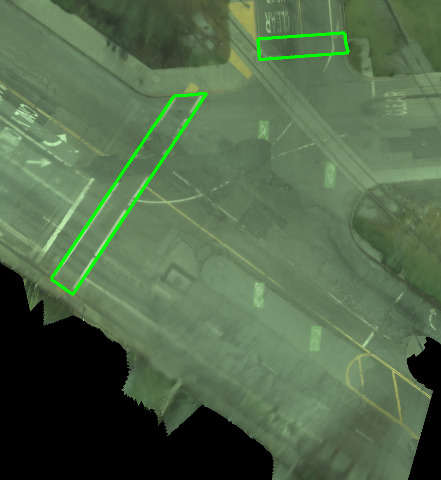} &\\

\includegraphics[width=.155\textwidth]{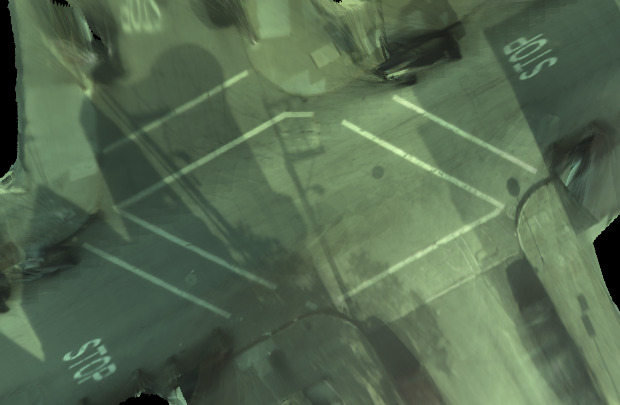} &
\includegraphics[width=.155\textwidth]{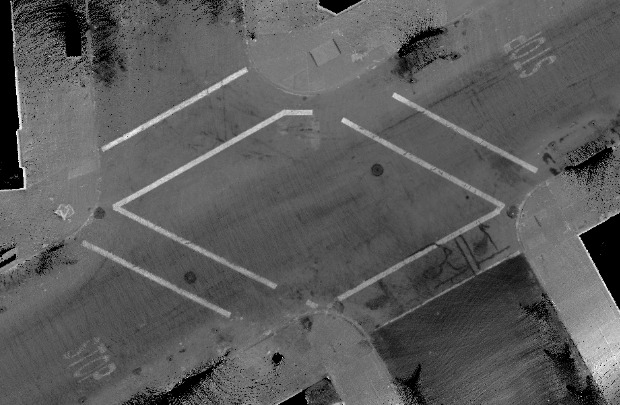} &
\includegraphics[width=.155\textwidth]{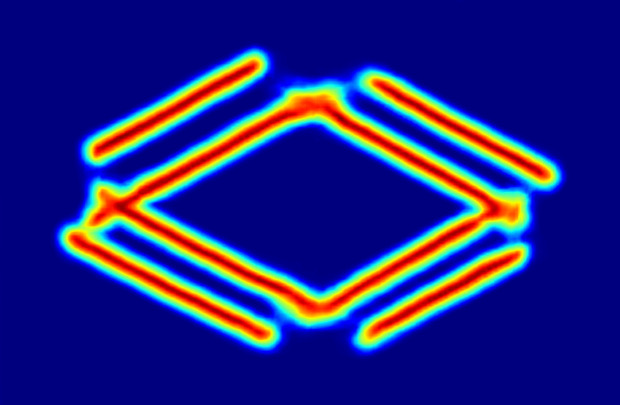} &
\includegraphics[width=.155\textwidth]{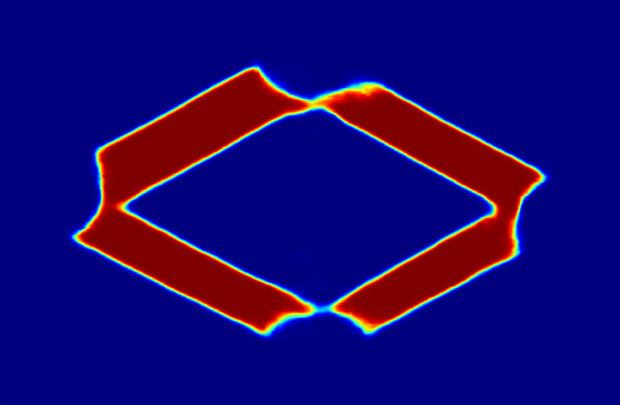} &
\includegraphics[width=.155\textwidth]{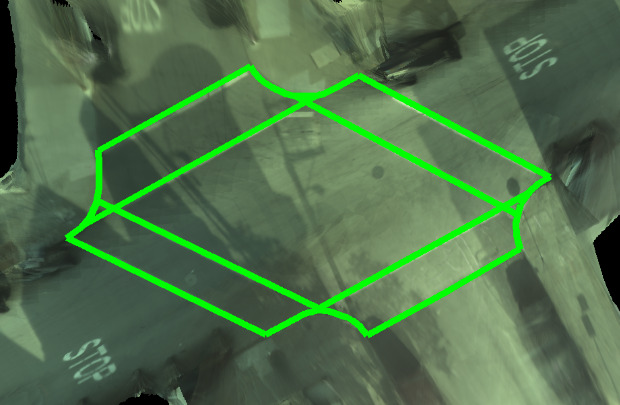} &
\includegraphics[width=.155\textwidth]{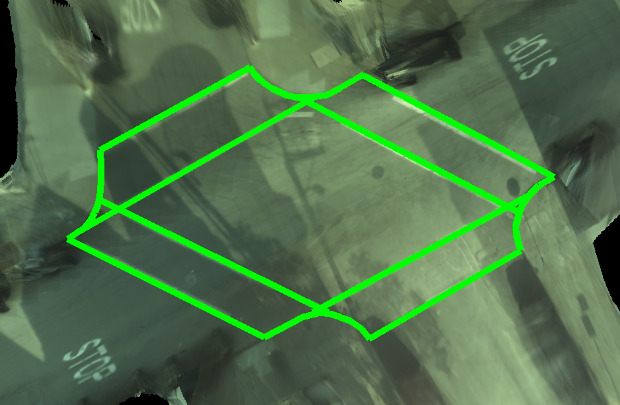} &\\

\includegraphics[width=.155\textwidth]{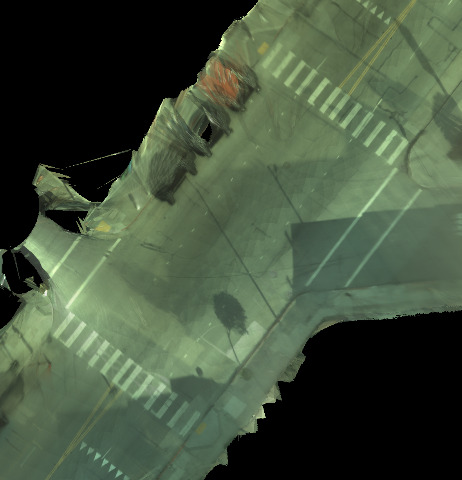} &
\includegraphics[width=.155\textwidth]{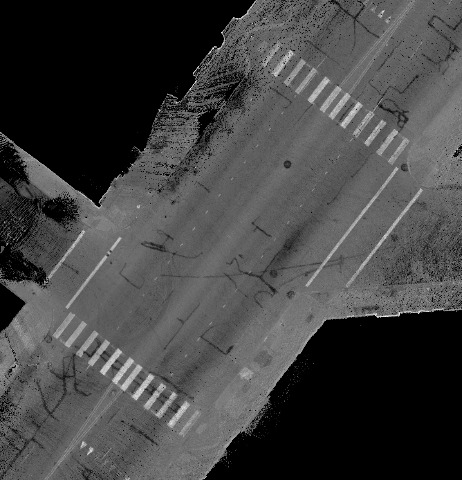} &
\includegraphics[width=.155\textwidth]{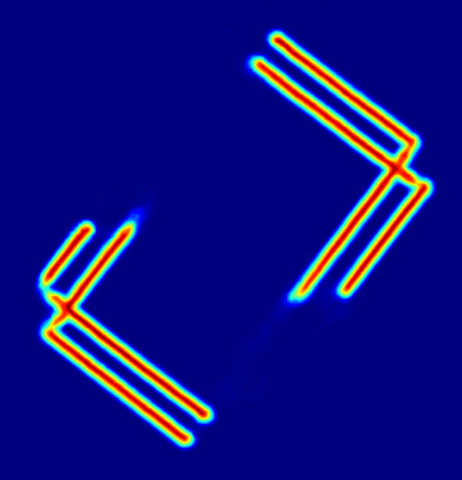} &
\includegraphics[width=.155\textwidth]{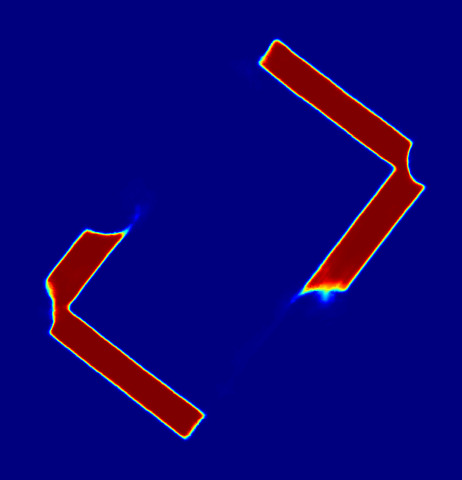} &
\includegraphics[width=.155\textwidth]{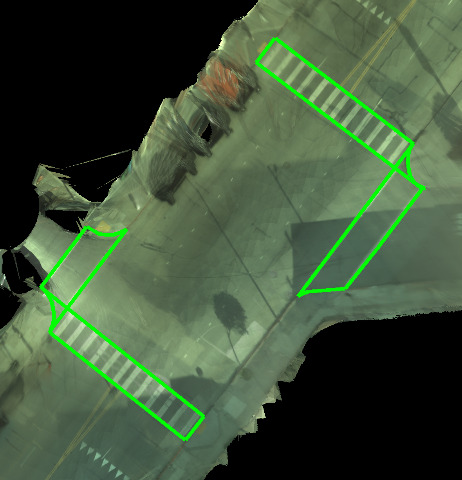} &
\includegraphics[width=.155\textwidth]{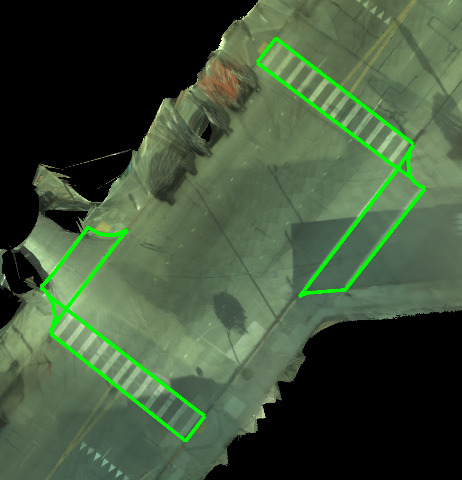} &\\
\end{tabular}

\vspace{-2mm}
\caption{Offline map model results using the the model trained on both camera and LiDAR imagery). Comparisons between col 1) ground camera, 2) ground lidar, 3) predicted inverse distance transform, 4) predicted segmentation, 5) predicted crosswalk polygons after inference and 6) gt crosswalk polygons.}
\label{fig:results}
\vspace{-5mm}

\end{figure}

\begin{figure}[t]
\centering
\begin{tabular}{ccccccr}
\includegraphics[width=.155\textwidth]{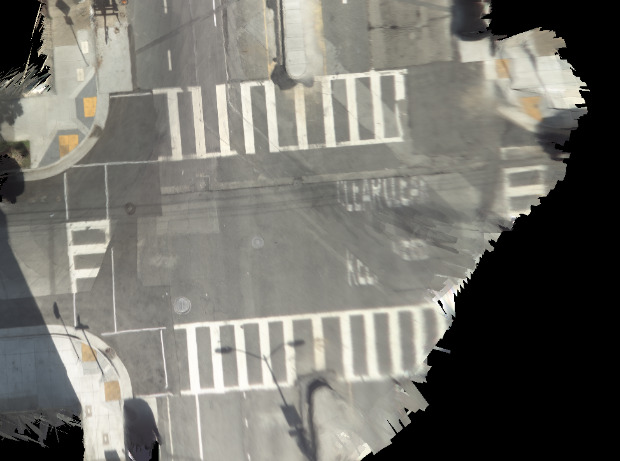} &
\includegraphics[width=.155\textwidth]{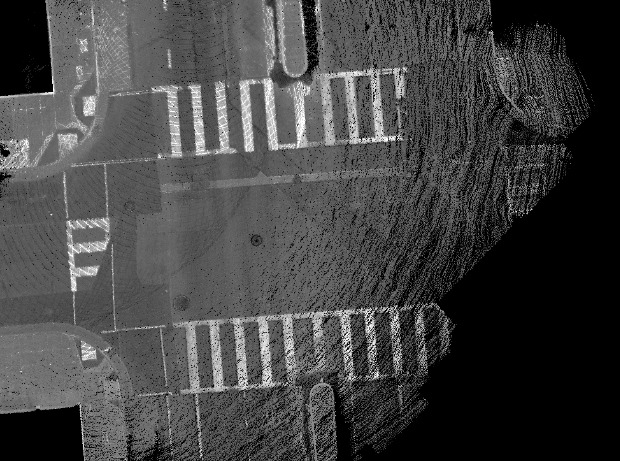} &
\includegraphics[width=.155\textwidth]{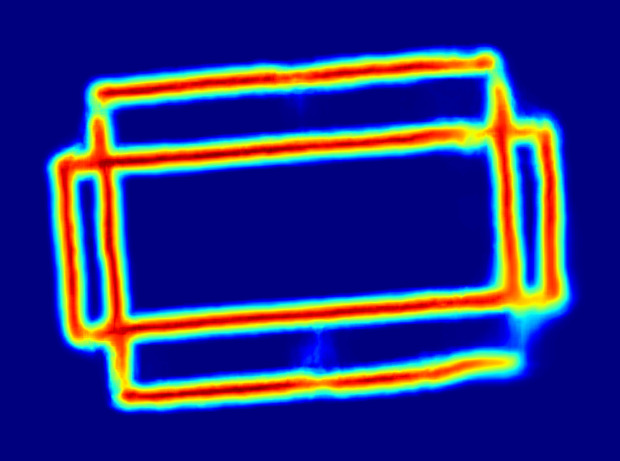} &
\includegraphics[width=.155\textwidth]{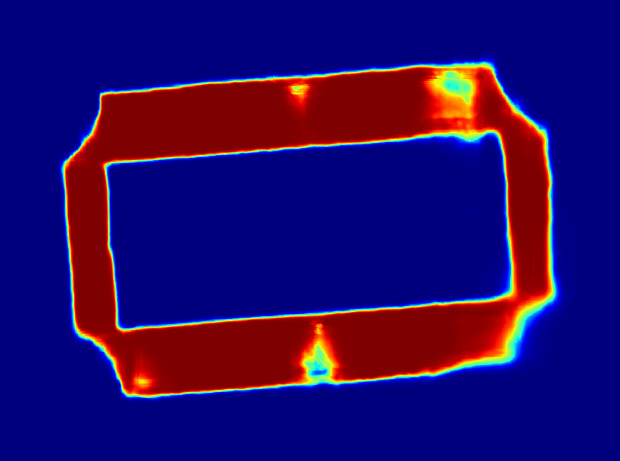} &
\includegraphics[width=.155\textwidth]{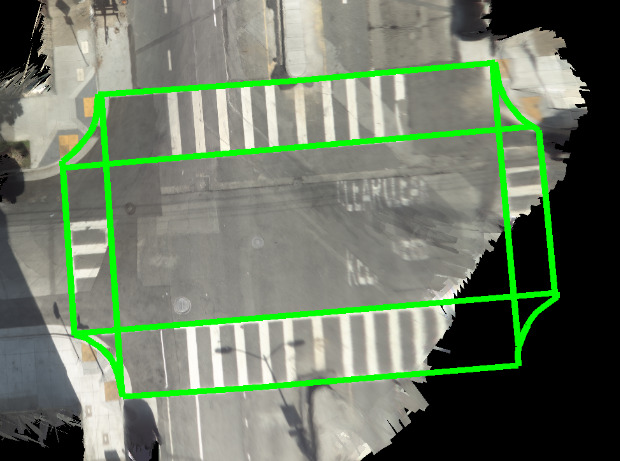} &
\includegraphics[width=.155\textwidth]{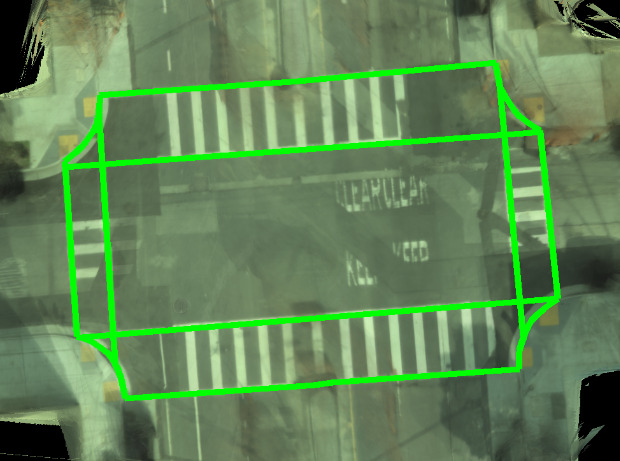} &\\

\includegraphics[width=.155\textwidth]{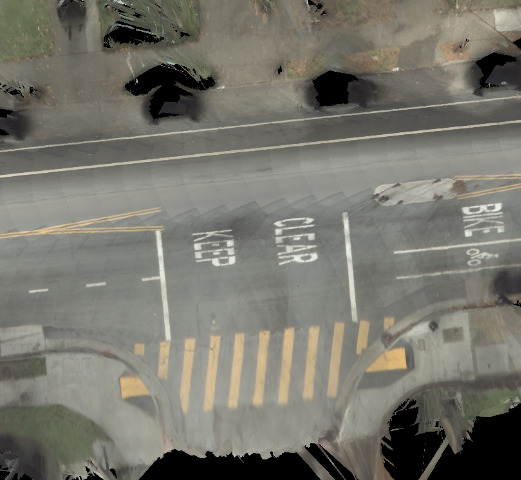} &
\includegraphics[width=.155\textwidth]{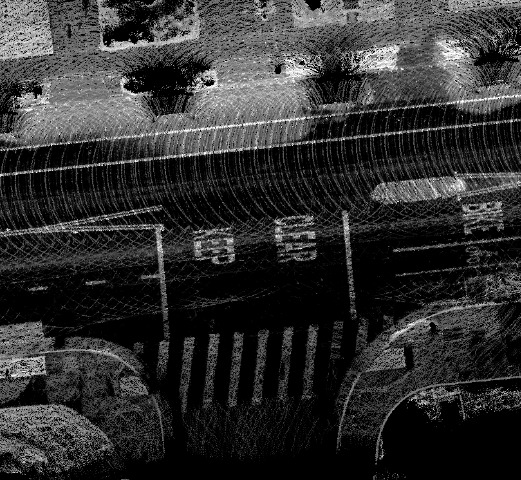} &
\includegraphics[width=.155\textwidth]{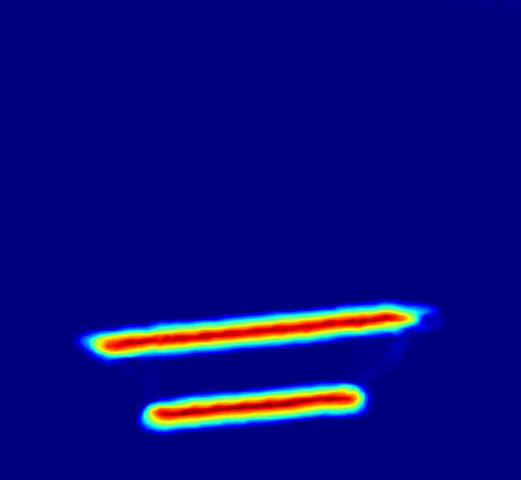} &
\includegraphics[width=.155\textwidth]{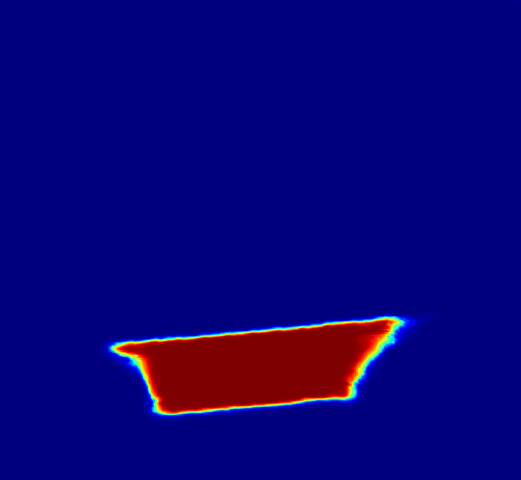} &
\includegraphics[width=.155\textwidth]{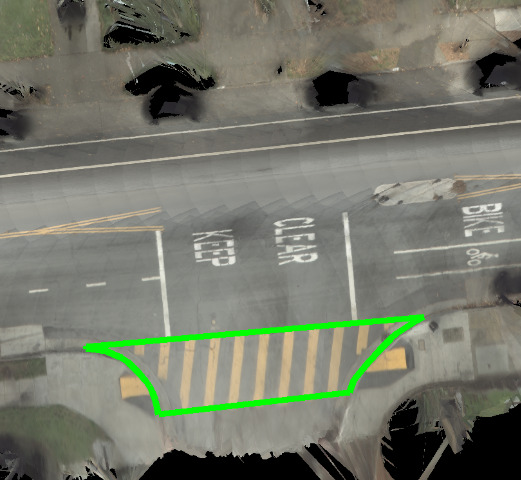} &
\includegraphics[width=.155\textwidth]{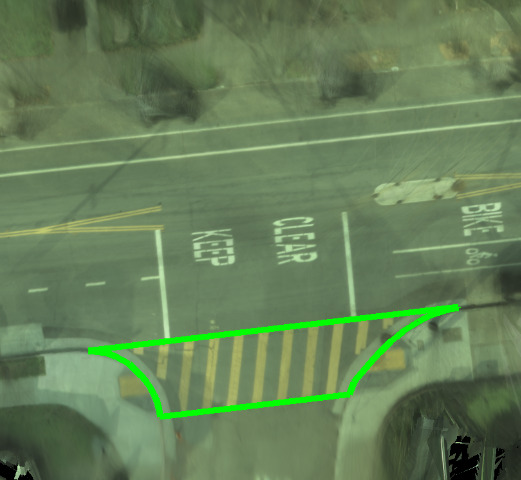} &\\

\includegraphics[width=.155\textwidth]{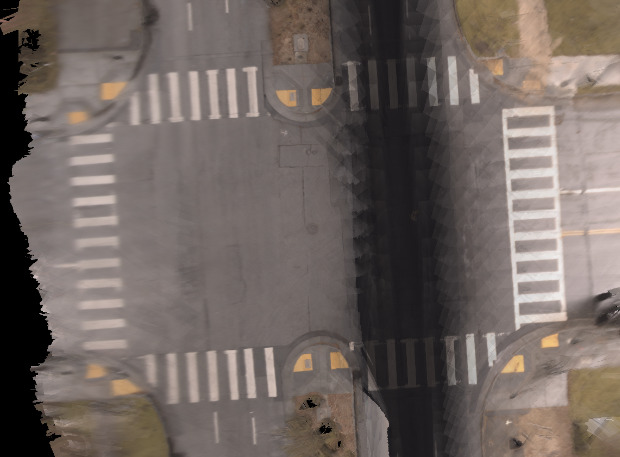} &
\includegraphics[width=.155\textwidth]{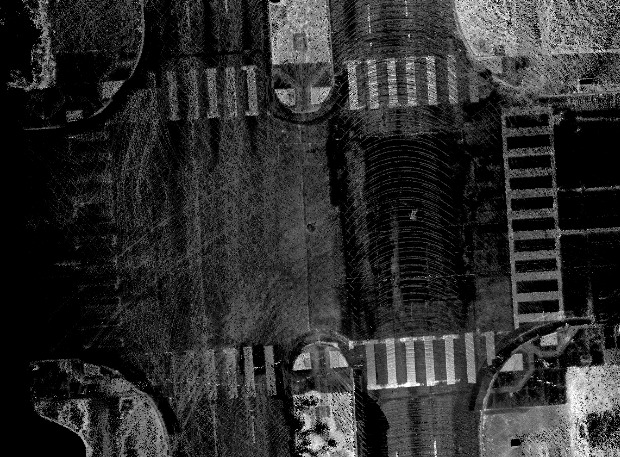} &
\includegraphics[width=.155\textwidth]{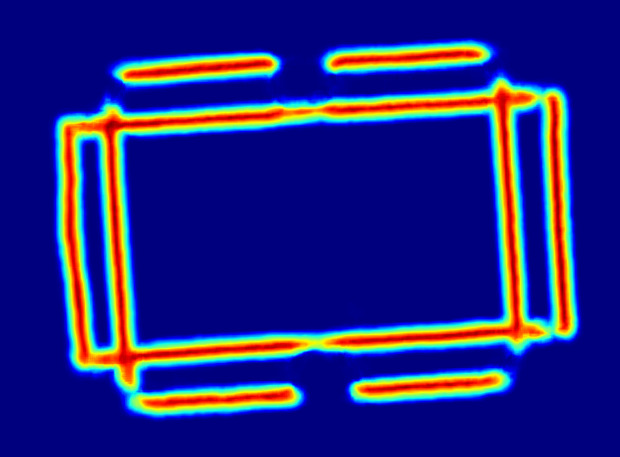} &
\includegraphics[width=.155\textwidth]{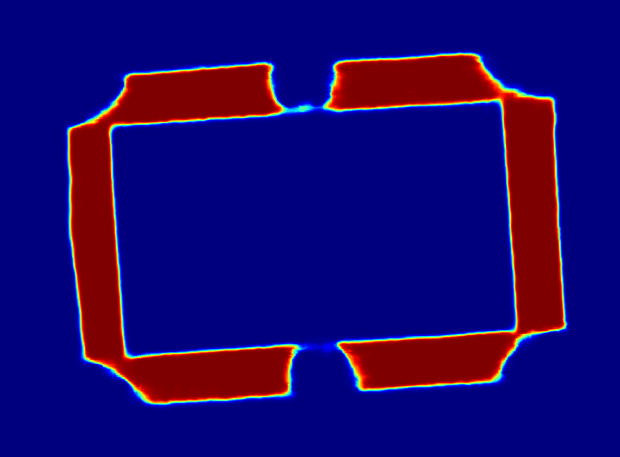} &
\includegraphics[width=.155\textwidth]{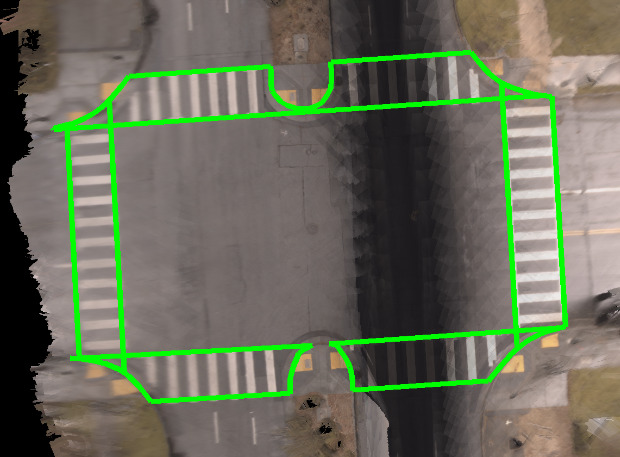} &
\includegraphics[width=.155\textwidth]{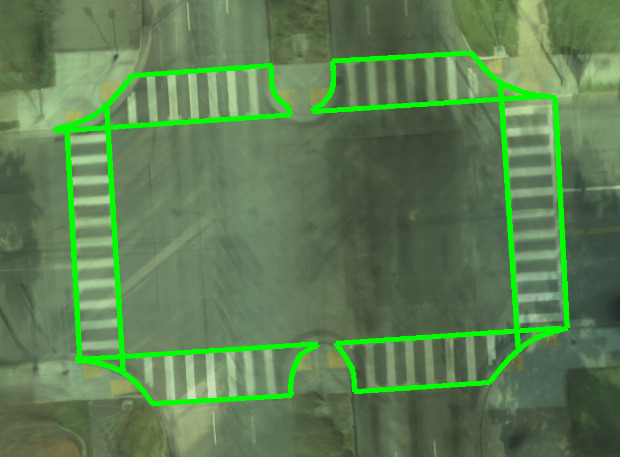} &\\

\includegraphics[width=.155\textwidth, height=1.5cm]{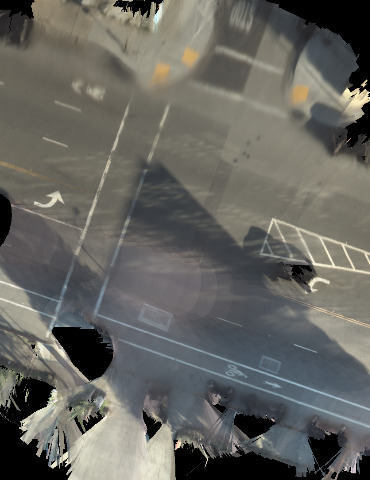} &
\includegraphics[width=.155\textwidth, height=1.5cm]{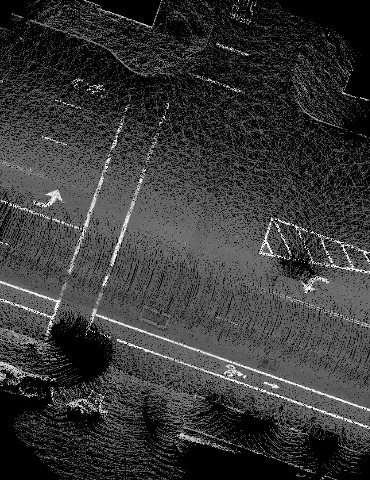} &
\includegraphics[width=.155\textwidth, height=1.5cm]{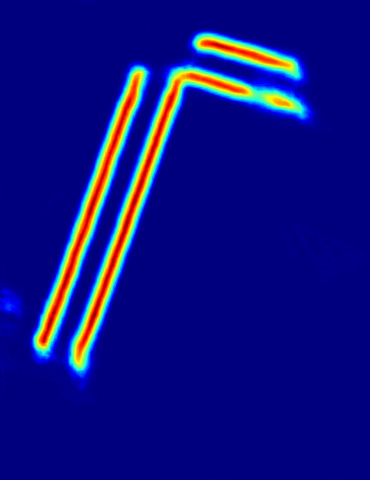} &
\includegraphics[width=.155\textwidth, height=1.5cm]{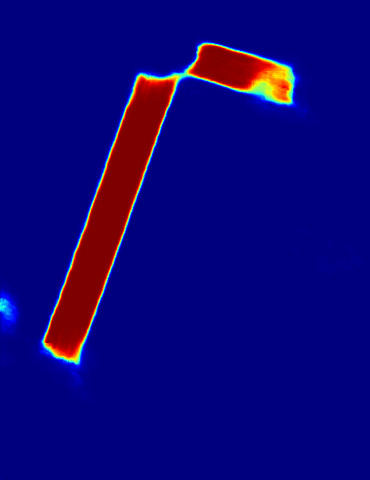} &
\includegraphics[width=.155\textwidth, height=1.5cm]{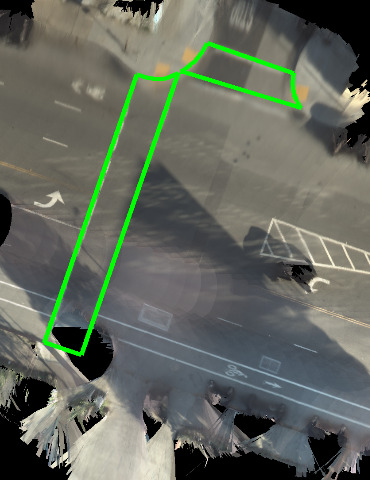} &
\includegraphics[width=.155\textwidth, height=1.5cm]{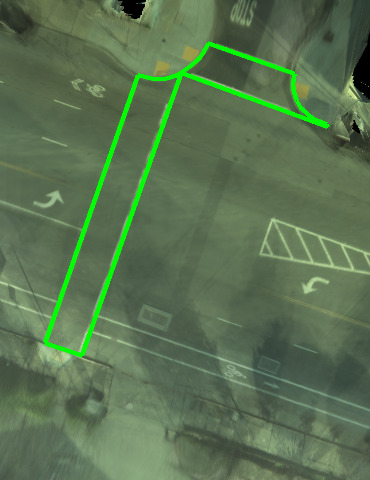} &\\

\includegraphics[width=.155\textwidth]{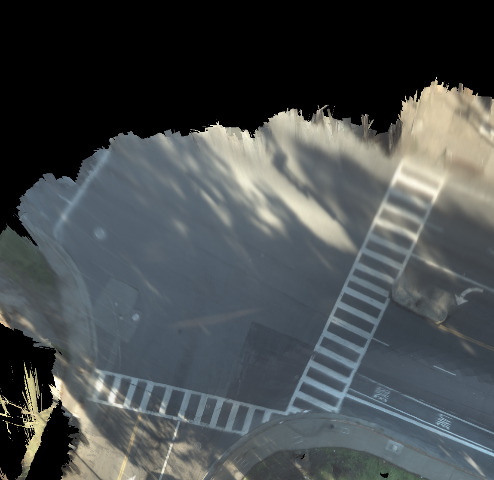} &
\includegraphics[width=.155\textwidth]{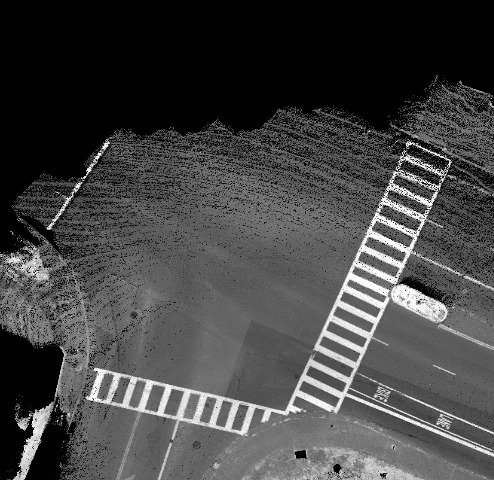} &
\includegraphics[width=.155\textwidth]{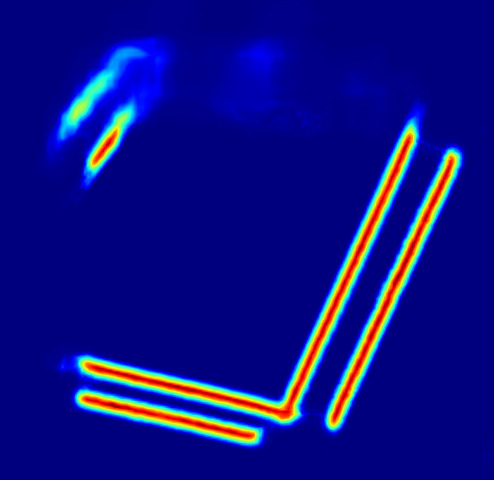} &
\includegraphics[width=.155\textwidth]{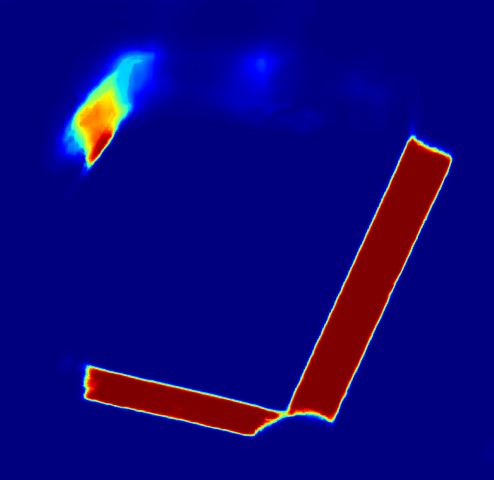} &
\includegraphics[width=.155\textwidth]{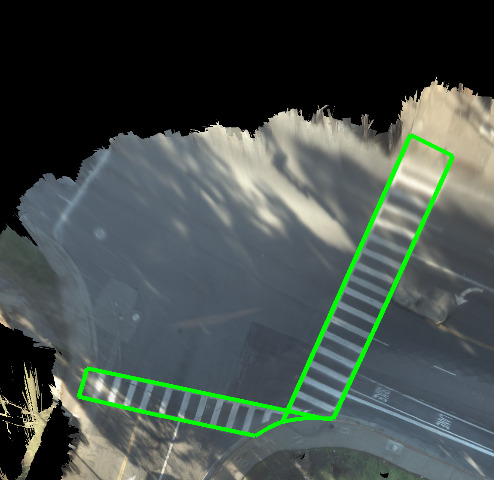} &
\includegraphics[width=.155\textwidth]{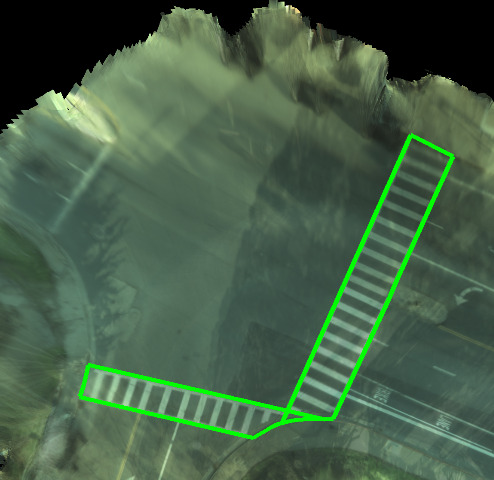} &\\

\includegraphics[width=.155\textwidth]{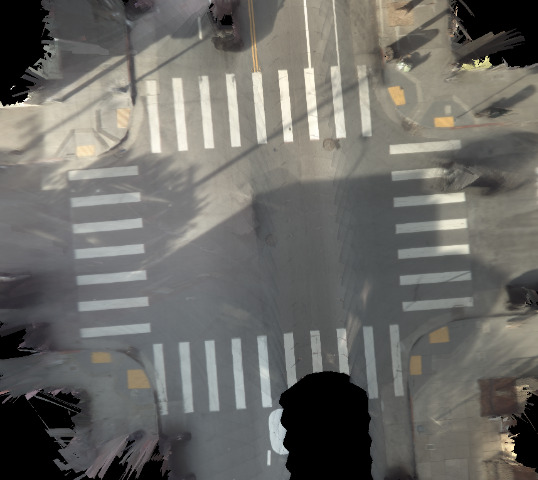} &
\includegraphics[width=.155\textwidth]{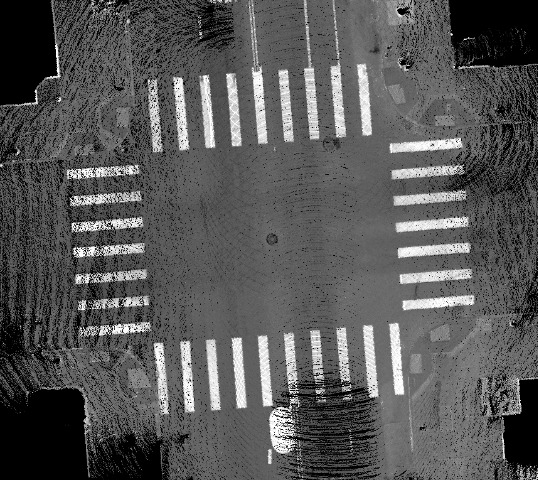} &
\includegraphics[width=.155\textwidth]{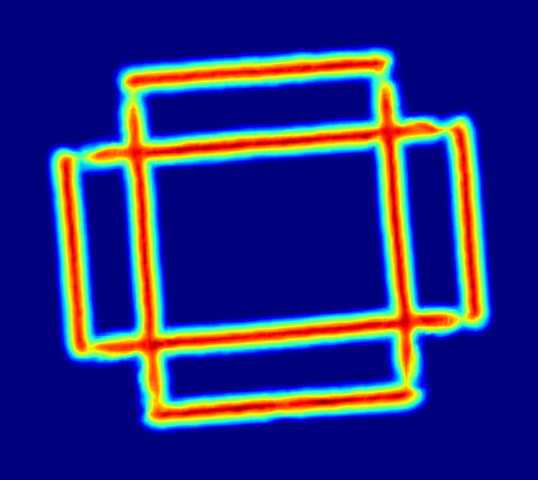} &
\includegraphics[width=.155\textwidth]{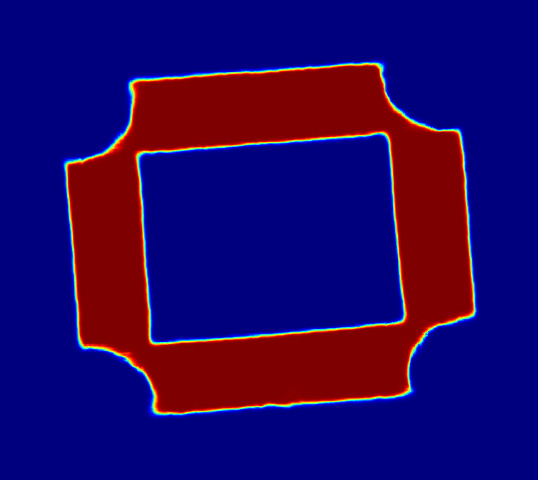} &
\includegraphics[width=.155\textwidth]{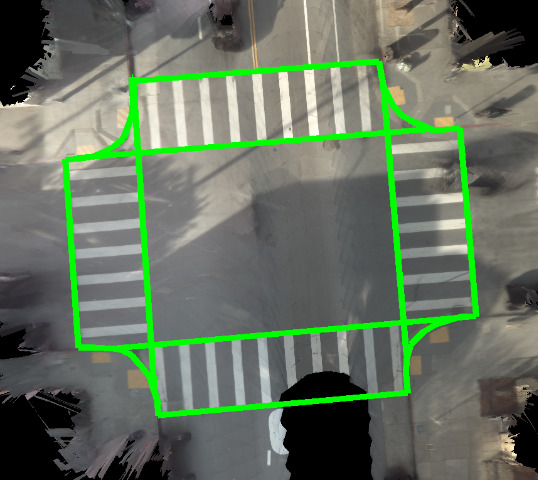} &
\includegraphics[width=.155\textwidth]{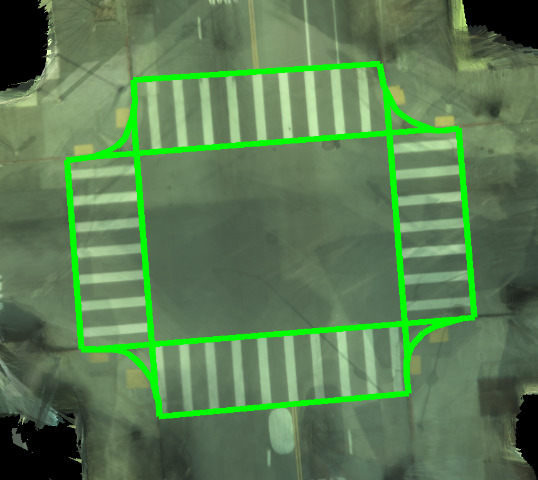} &\\

\end{tabular}

\vspace{-2mm}
\caption{Online map model results using the model trained on both camera and LiDAR imagery. Comparisons between col 1) ground camera (online map), 2) ground LiDAR (online map), 3) predicted inverse distance transform, 4) predicted segmentation, 5) predicted crosswalk polygons after inference and 6) gt crosswalk polygons overlayed on the ground camera (offline map).}
\label{fig:results_1}
\vspace{-5mm}
\end{figure}

Table \ref{tab:results} depicts results obtained when using a single pass (online mapping) vs using several passes of driving to create the input feature map (offline mapping). As expected, using multiple passes for offline mapping results in better performance with 96.6\% (row 7, prec $@$ 40cm), but 91.5\% (row 4, prec $@$ 40cm) automation can be reached in the online setting. We visualize some of the results from the model trained on both camera and LiDAR in an offline map setting in Figure \ref{fig:results}, while Figure \ref{fig:results_1} shows results of the online map setting. Our approach does a very good job at drawing crosswalks with very complex topology in both settings.

\subsubsection{Importance of Structured Representation:} The first entry in Table \ref{tab:results} shows the results of using a nearest neighbor algorithm on top of VGG features extracted from both the camera and LiDAR. However, this only achieves 24.8\% automation (precision $@$ 40 cm). The second entry in Table \ref{tab:results} shows the results of using only the output of the CNN model's semantic segmentation branch for the final prediction. As shown, the network is doing a great job but only 93.1\% automation (precision $@$ 40cm) can be achieved in the offline setting. 

\subsubsection{Speed:} The CNN forward pass runs at 50 ms per image. The unoptimized structured prediction step runs at 0.75s on a single core CPU. Optimizing the code would significantly improve the speed.

\subsubsection{Qualitative Results:} We refer the reader to Figure \ref{fig:results} and Figure \ref{fig:results_1} for an illustration of results for both offline and online settings. Despite the complex topology, our approach is still able to accurately draw the crosswalks.

\subsubsection{Human Disagreement:}
We compare the noise in human annotation of the ground truth by annotating 100 intersections with several annotators. Here we calculate the precision, recall and IoU. As shown in the last row in  Table \ref{tab:results}, there is about a 4.7\% error in IOU, and a 11.7\% and 12.7\% error in the precision and recall at 20cm between different individuals.

\subsubsection{Crosswalk Angle Analysis:}
\begin{figure}[t]
\vskip 0.2in
\begin{center}
\centerline{\includegraphics[width=0.8\textwidth]{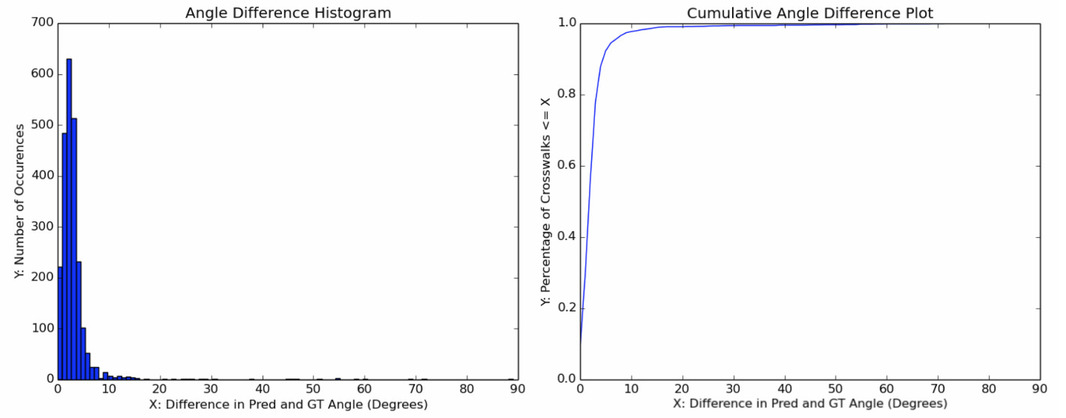}}
\vspace{-2mm}
\caption{A histogram of the angle differences (left) and cumulative angle difference graph (right) using the offline model trained on both camera and LiDAR.}
\label{fig:angle_analysis}
\end{center}
\vskip -0.2in
\vspace{-5mm}
\end{figure}

Having the correct crosswalk angle is crucial to achieving a high performance on our results. Thus, we perform analysis on the combination of the predicted alignment and centreline angle and compare it to the ground truth. That is, we find the difference between the angle used in inference with the ground truth angle. We plot a histogram and cumulative graph of the differences in Figure \ref{fig:angle_analysis}. The model we analyze is the model trained on both the camera and LiDAR imagery from the offline maps. We find that 89\% of the crosswalk angles used are within $\pm5^{\circ}$ from the ground truth. After the structured prediction step (which searches over additional angles) this becomes 98\%. 

\subsubsection{Intersection Complexity:}
\begin{figure}[t]
\vskip 0.2in
\begin{center}
\centerline{\includegraphics[width=\textwidth]{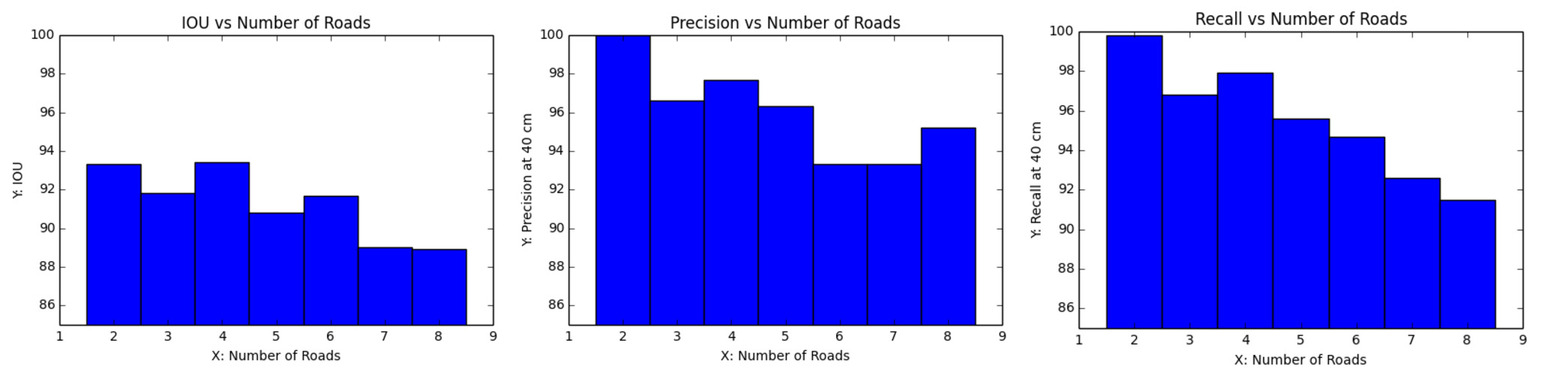}}
\vspace{-2mm}
\caption{We visualize the effect of the number of neighbouring roads on the results of the model trained on offline maps using both camera and LiDAR. Histograms of the IoUs vs. number of neighbouring roads (left), Precision at 40 cm vs. number of neighbouring roads (middle) and Recall at 40 cm vs. number of neighbouring roads (right) are plotted here. We use the offline model trained on both camera and LiDAR for this analysis.}
\label{fig:neighbours}\end{center}
\vskip -0.2in
\vspace{-2mm}
\end{figure}

We analyze the effect of the number of neighboring roads on our results. A neighboring road is defined as one of the connecting roads to the intersection that provides a road centerline for our structured prediction algorithm. If a street has a divider in the middle, then we split the street into 2 roads. Hence, it is possible for a 4 way intersection to have 8 roads, that is, 2 roads for each approach to the intersection. As shown in  Figure \ref{fig:neighbours}, as the number of roads increases, the performance decreases. This is expected, as those intersections are more complex.

\subsubsection{Ablation Studies:}
We perform ablation studies to analyze the contributions of different components in our model in the context of offline mapping with cameras and LiDAR. 
The results are shown on Table \ref{tab:ablation}.
We first explore the effect of removing certain components of the model. We remove the angle search of $\pm$ 2 $^{\circ}$ and $\pm$ 5 $^{\circ}$ in row (2) and remove the usage of the centerline angle in row (3). Both result in a slight decrease in performance. In row (4) we do not use the predicted angle when drawing the crosswalks; we see a significant drop of more than 10\% for all the performance metrics. This suggests that having the alignment prediction is very important for good inference results.

\begin{table}[h]
\centering
  \begin{tabular}{|l|*{9}{c|}}
  \cline{2-10}
  \multicolumn{1}{c|}{} &  \multicolumn{4}{|c}{Precision at (cm)} & \multicolumn{4}{|c|}{Recall at (cm)} & \multicolumn{1}{|c|}{IOU} \\ 
  \cline{2-10}
  \multicolumn{1}{c|}{} & \multicolumn{1}{|c}{20} & \multicolumn{1}{|c}{40} & \multicolumn{1}{|c}{60} & \multicolumn{1}{|c}{80} & \multicolumn{1}{|c}{20} & \multicolumn{1}{|c}{40} &   \multicolumn{1}{|c}{60} & \multicolumn{1}{|c}{80} & \multicolumn{1}{|c|}{} \\ 
  \hline 

  \textrm{Ours} &$85.6\%$ &$96.6\%$ &$98.1\%$  &$98.8\%$ 
    &$86.1\%$ &$96.8\%$ &$98.2\%$ &$98.7\%$ 
    &$92.4\%$ \\ 
  \textrm{No Ang Search} &$82.2\%$ &$94.4\%$ &$97.1\%$  &$98.1\%$ 
    &$82.7\%$ &$94.7\%$ &$97.2\%$ &$98.2\%$ 
    &$91.3 \%$ \\     
  \textrm{No Cent Ang} &$84.5\%$ &$96.3\%$ &$98.1\%$  &$98.8\%$ 
    &$84.9\%$ &$96.4\%$ &$98.0\%$ &$98.6\%$ 
    &$92.1 \%$ \\   
  \textrm{No Pred Ang} &$74.0\%$ &$85.3\%$ &$88.9\%$  &$91.4\%$ 
    &$73.8\%$ &$84.8\%$ &$88.3\%$ &$90.5\%$ 
    &$83.7 \%$ \\   
  \textrm{GT DT} &$88.5\%$ &$96.6\%$ &$97.8\%$  &$98.3\%$ 
    &$89.5\%$ &$97.3\%$ &$98.4\%$ &$98.8\%$ 
    &$92.9 \%$ \\  
  \textrm{GT Seg} &$\textbf{94.1}\%$ &$\textbf{97.8}\%$ &$\textbf{98.7\%}$  &$\textbf{99.2\%}$ 
    &$94.7\%$ &$98.1\%$ &$98.8\%$ &$99.1\%$ 
    &$94.9 \%$ \\  
  \textrm{GT Ang} &$85.5\%$ &$96.5\%$ &$98.1\%$  &$98.7\%$ 
    &$85.7\%$ &$96.4\%$ &$97.9\%$ &$98.4\%$ 
    &$92.2 \%$ \\  
  \textrm{GT DT+S+A} &$93.9\%$ &$97.5\%$ &$98.5\%$  &$99.0\%$ 
    &$\textbf{94.9\%}$ &$\textbf{98.1\%}$ &$\textbf{98.9\%}$ &$\textbf{99.2\%}$ 
    &$\textbf{94.9\%}$ \\   
    
  \hline 
    
  \end{tabular}
  \caption{We report the ablation studies and performance using oracle information in this table. For the ablation studies we analyze the effect of the angle search, road centreline angles and predicted angles in rows (2-4). For the oracle information we inject GT distance transform, segmentation and angles and analyze the results in rows (5-8).}
  \label{tab:ablation}
  \vspace{-5mm}
\end{table}

\subsubsection{Oracle Performance:} 
We analyze the upper bound performance of our system by introducing oracle information. 
Comparing rows (5), (6) and (7) in Table \ref{tab:ablation}  we see that having ground truth segmentation significantly increases the performance of the model. On the other hand, having ground truth distance transform only slightly increases the performance.  Interestingly, using ground truth angle performance slightly worse than our result in row (1). This is likely due to the fact that our predicted angles are very accurate. Our angle analysis shows that without ground truth angles we can already achieve 98\% angle accuracy. On row (8), we use the ground truth distance transform, segmentation and angle and see that this performs around the same as using just ground truth segmentation. This suggests that improvements to the semantic segmentation in future models will yield the greatest impact.

\begin{figure}[t]
\centering
\begin{tabular}{ccccccr}
\includegraphics[width=.155\textwidth, height=1.75cm]{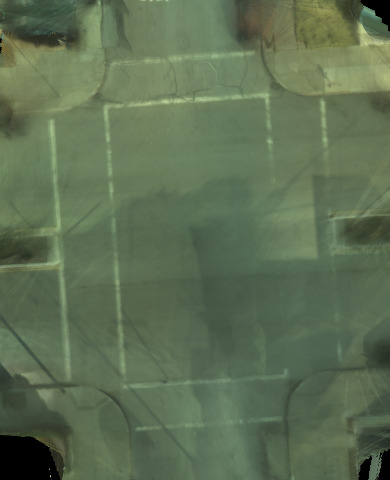} &
\includegraphics[width=.155\textwidth, height=1.75cm]{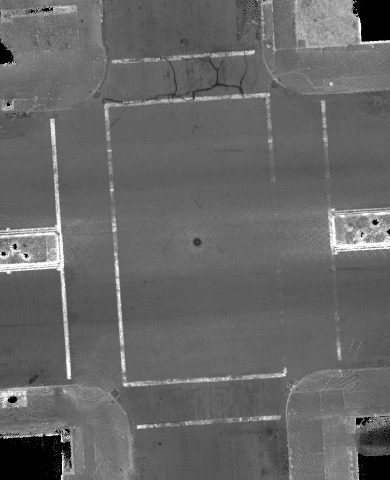} &
\includegraphics[width=.155\textwidth, height=1.75cm]{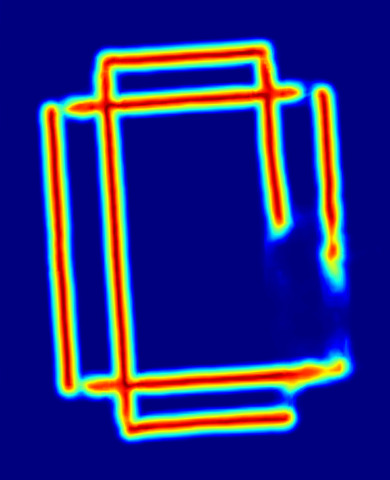} &
\includegraphics[width=.155\textwidth, height=1.75cm]{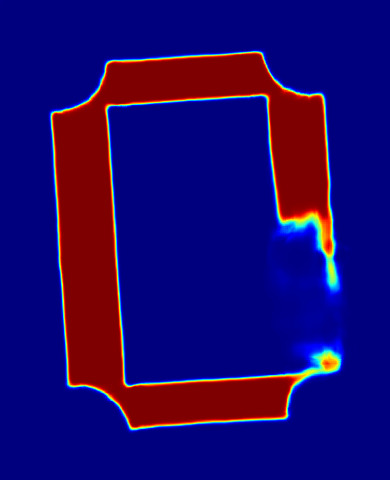} &
\includegraphics[width=.155\textwidth, height=1.75cm]{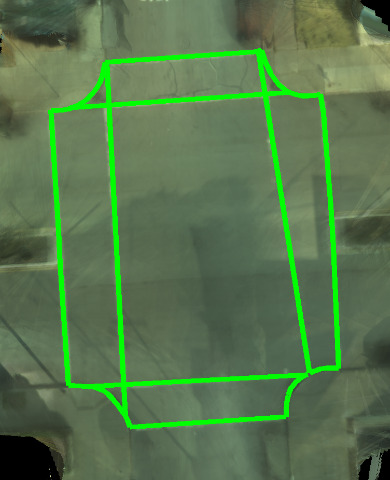} &
\includegraphics[width=.155\textwidth, height=1.75cm]{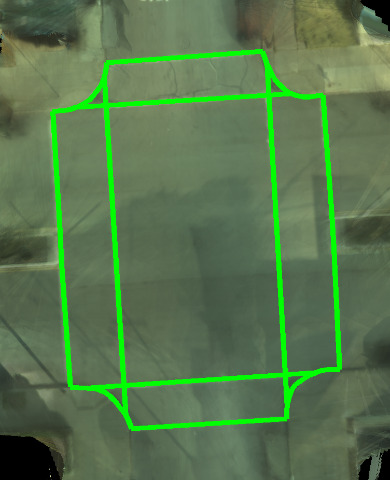} &\\

\includegraphics[width=.155\textwidth]{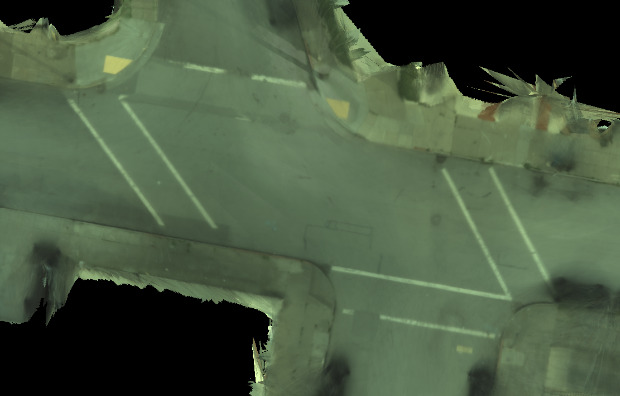} &
\includegraphics[width=.155\textwidth]{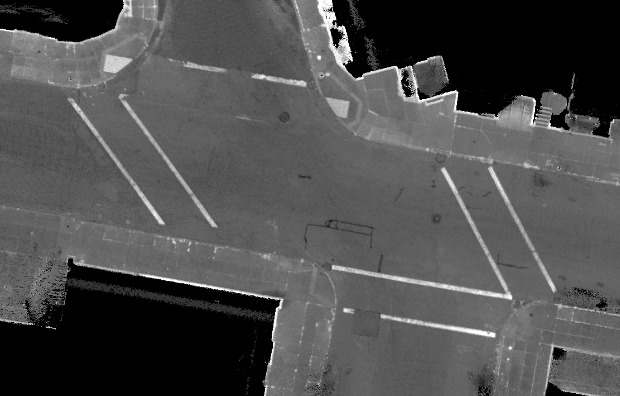} &
\includegraphics[width=.155\textwidth]{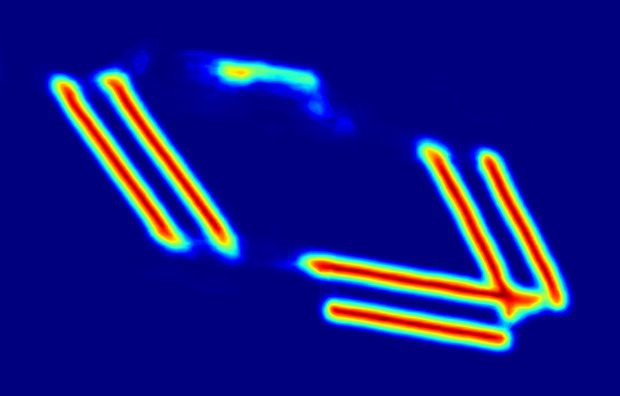} &
\includegraphics[width=.155\textwidth]{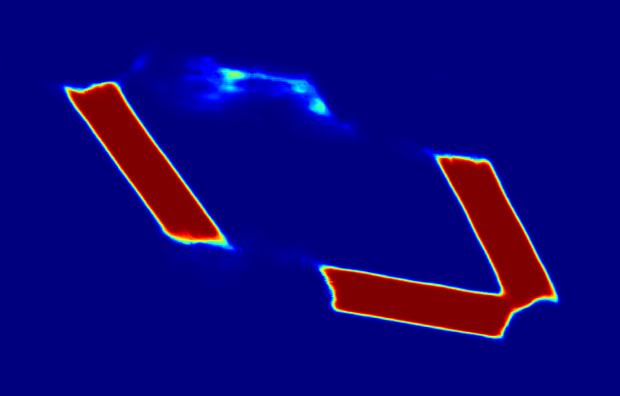} &
\includegraphics[width=.155\textwidth]{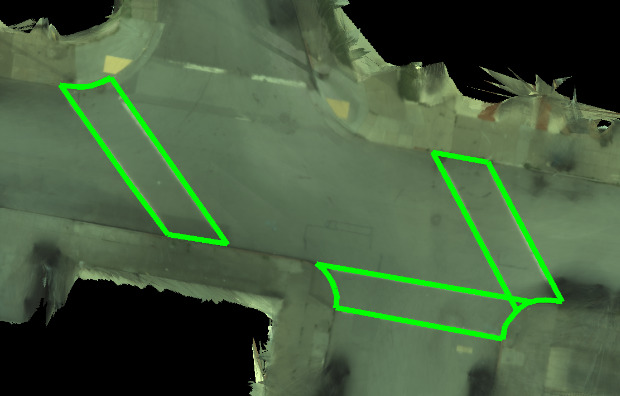} &
\includegraphics[width=.155\textwidth]{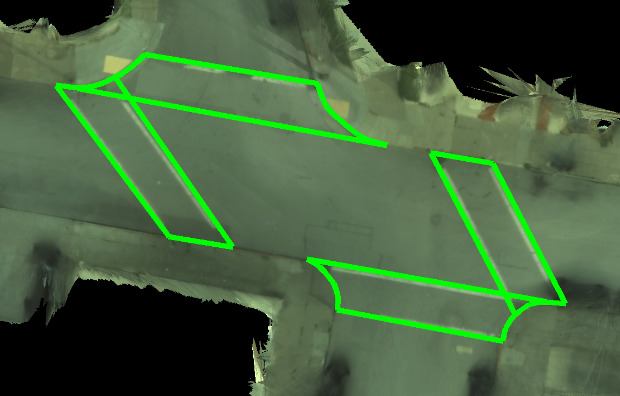} &\\

\end{tabular}

\vspace{-2mm}
\caption{The main failure modes are caused by the trade off between segmentation and distance transform weights (top) and poor image/paint quality (bottom). Here we show comparisons between col 1) ground camera, 2) ground LiDAR, 3) predicted inverse distance transform, 4) predicted segmentation, 5) predicted crosswalk polygons after inference and 6) gt crosswalk polygons.}
\label{fig:failure_modes}
\vspace{-3mm}
\end{figure}

\begin{figure}[!h]
\centering
\begin{tabular}{ccccccr}
\includegraphics[width=.155\textwidth, height=2cm]{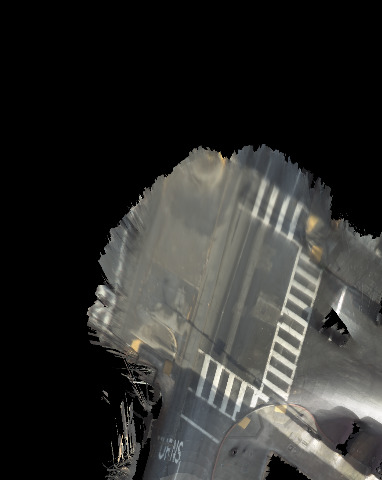} &
\includegraphics[width=.155\textwidth, height=2cm]{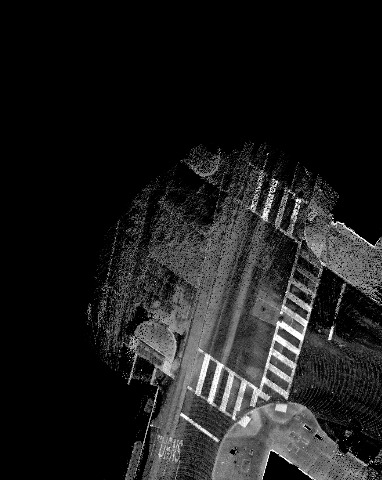} &
\includegraphics[width=.155\textwidth, height=2cm]{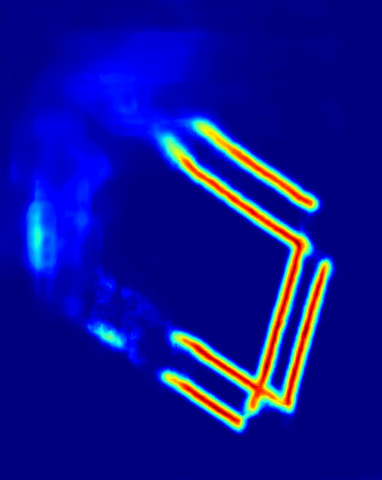} &
\includegraphics[width=.155\textwidth, height=2cm]{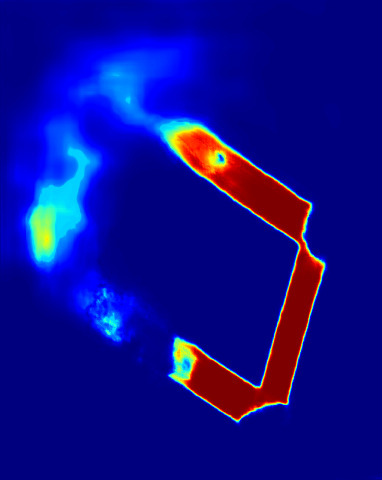} &
\includegraphics[width=.155\textwidth, height=2cm]{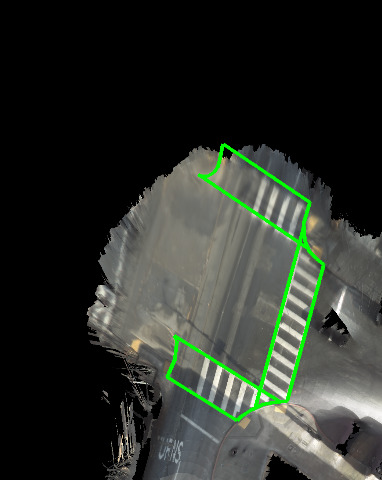} &
\includegraphics[width=.155\textwidth, height=2cm]{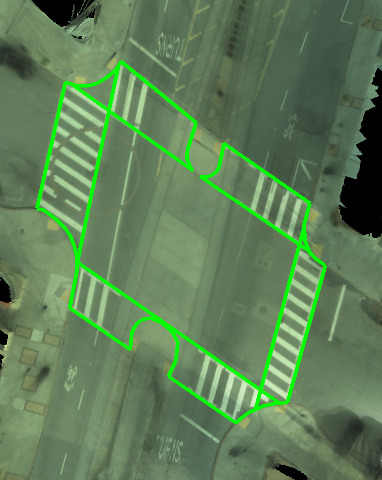} &\\

\includegraphics[width=.155\textwidth]{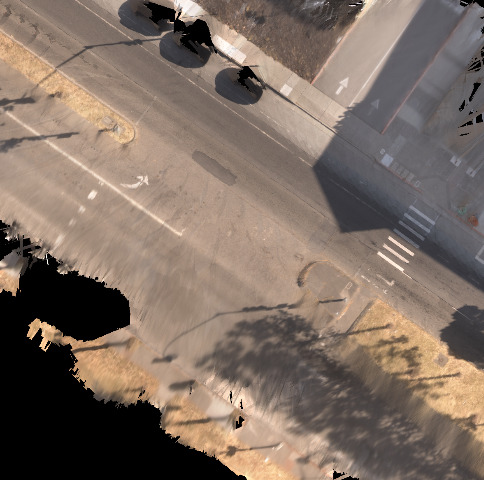} &
\includegraphics[width=.155\textwidth]{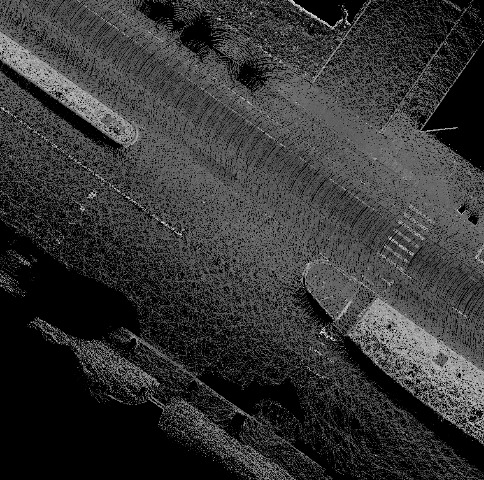} &
\includegraphics[width=.155\textwidth]{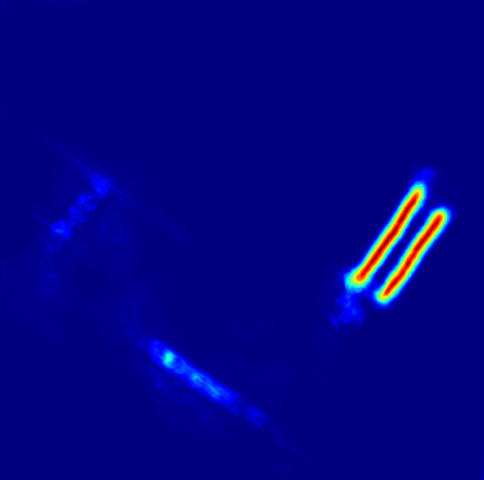} &
\includegraphics[width=.155\textwidth]{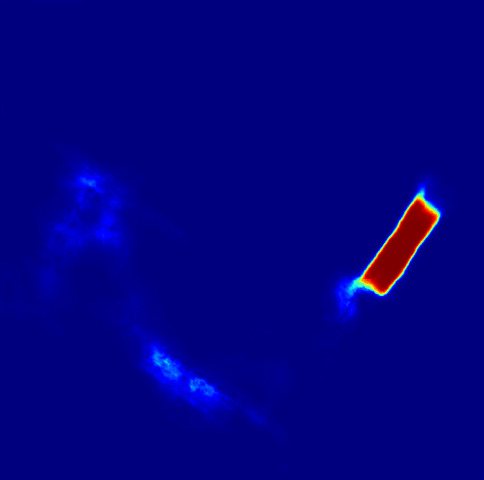} &
\includegraphics[width=.155\textwidth]{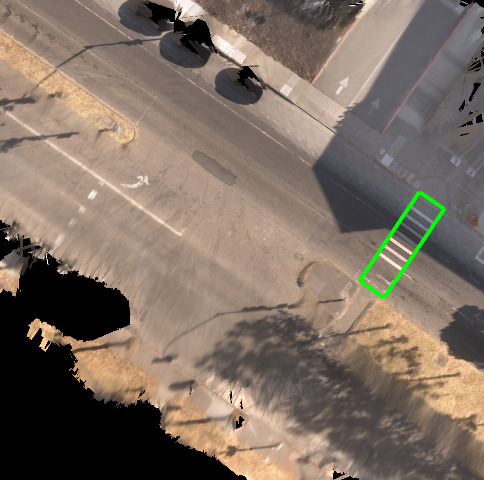} &
\includegraphics[width=.155\textwidth]{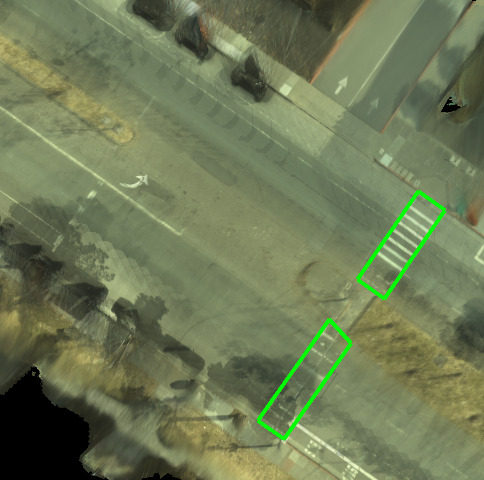} &\\

\end{tabular}

\caption{The main failure mode for the online models is caused by poor data collection when mapping the roads. This poor data collection causes holes and/or poor image quality as seen in this figure. Here we show comparisons between col 1) ground camera (online map), 2) ground LiDAR (online map), 3) predicted inverse distance transform, 4) predicted segmentation, 5) predicted crosswalk polygons after inference and 6) gt crosswalk polygons overlayed on the ground camera (offline map).}
\label{fig:failure_modes_1}
\vspace{-5mm}
\end{figure}

\subsubsection{Failure Modes:}
Since we use the weight $\lambda_I$ to weigh between maximizing segmentation or distance transform energies in our energy formulation, we may at times choose the wrong weighting for a particular input. As seen in Figure \ref{fig:failure_modes} (top), since almost half of the crosswalk boundary in the right crosswalk is missing, our model predicts the wrong segmentation. In this case, our model shows that predicting a boundary that focuses on the segmentation energy gives a larger value and thus produces the wrong inference. The second failure mode can be seen in the bottom image. Here, the paint quality in the ground imagery (although not shown, this is also true for the LiDAR imagery) is of poor quality. Thus, our model mistakes the crosswalk for a stop line at an intersection, and does not predict its presence for the segmentation output. For the online mapping scenario, the major failure mode are holes in the map, as shown in  Fig. \ref{fig:failure_modes_1}.

\subsubsection{False Positives:}
Our dataset was composed of images that contain
crosswalks. Without retraining, our approach produces 5.7\% false positives. When retrained with images that do not contain crosswalks (45\% added images) the false positive rate is 0.04\%. The performance of the retrained model is around the same as our result from Table \ref{tab:results} row (8). Examples of the retrained model results can be seen in Fig \ref{fig:fp}.

\section{Conclusion}
\label{conclusion}

\begin{figure}[t]
\centering
\centering
\begin{tabular}{cccccccr}
\includegraphics[width=.155\textwidth]{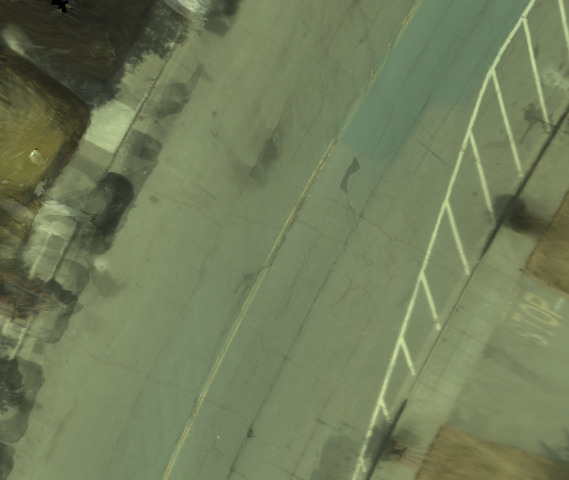} &
\includegraphics[width=.155\textwidth]{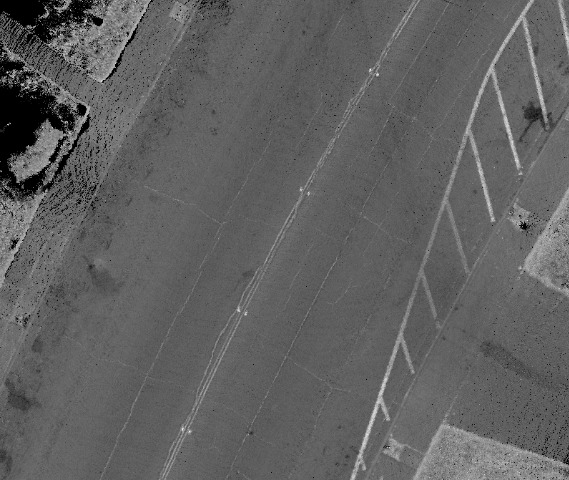} &
\includegraphics[width=.155\textwidth]{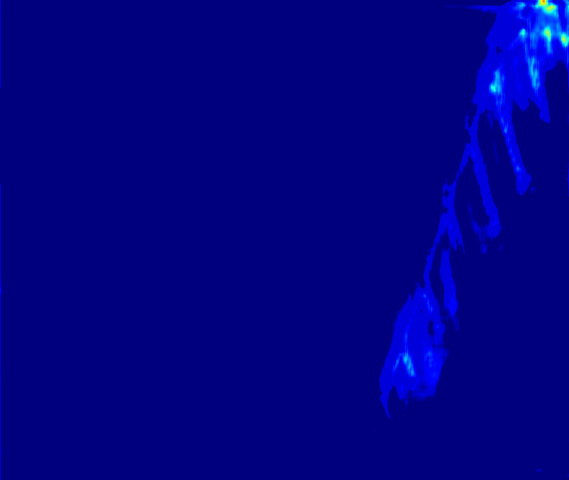} &
\includegraphics[width=.155\textwidth]{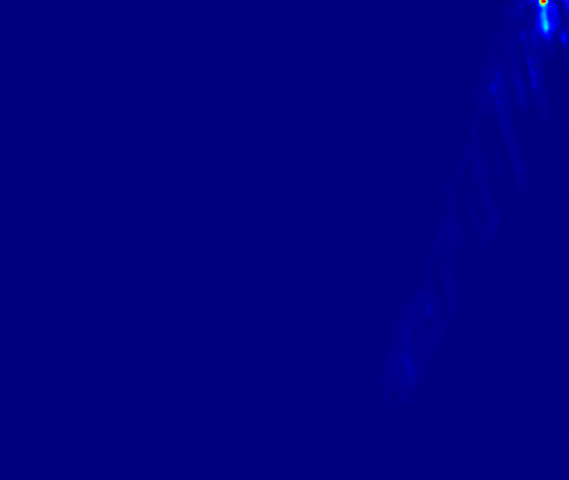} &
\includegraphics[width=.155\textwidth]{images/no_crosswalks/nan_sfo_1c8dcf28-e369-4f6a-c89e-328dbc135032_60b4efcb-e267-4a53-e8f1-59e9af05ac1c_ground.jpg} &
\includegraphics[width=.155\textwidth]{images/no_crosswalks/nan_sfo_1c8dcf28-e369-4f6a-c89e-328dbc135032_60b4efcb-e267-4a53-e8f1-59e9af05ac1c_ground.jpg} &\\

\includegraphics[width=.155\textwidth]{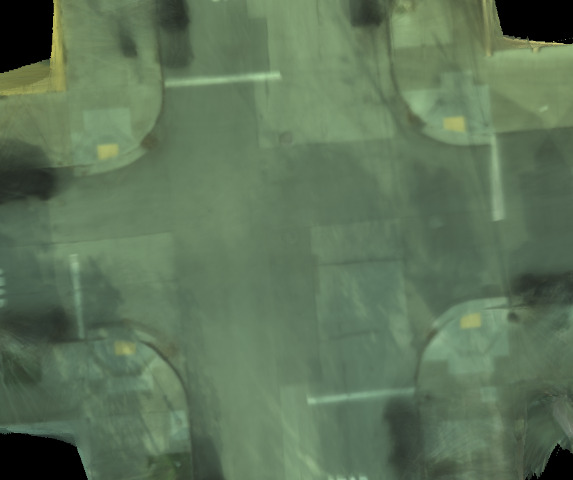} &
\includegraphics[width=.155\textwidth]{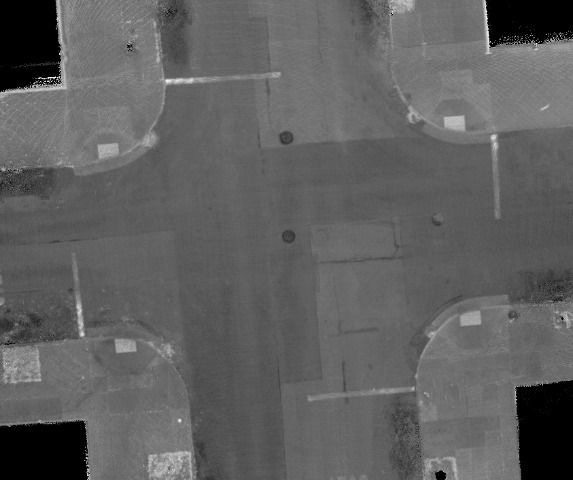} &
\includegraphics[width=.155\textwidth]{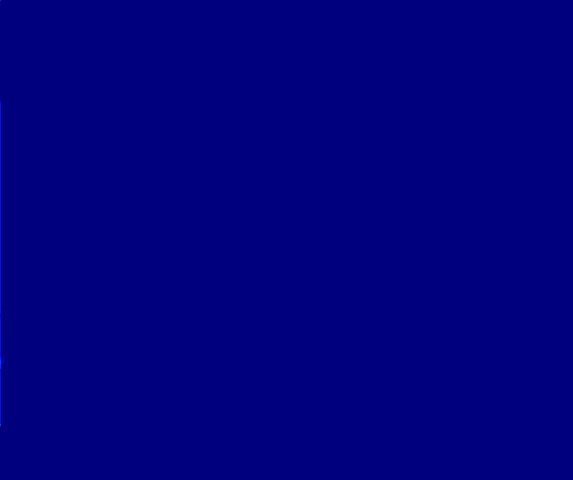} &
\includegraphics[width=.155\textwidth]{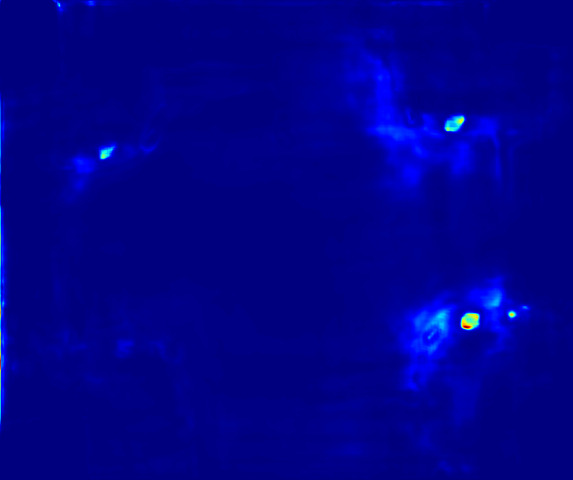} &
\includegraphics[width=.155\textwidth]{images/no_crosswalks/nan_sfo_1ffddf16-b2b2-4fa8-ceb3-9dfe23e47fcb_84088aac-1543-4c61-d407-2d531de79695_ground.jpg} &
\includegraphics[width=.155\textwidth]{images/no_crosswalks/nan_sfo_1ffddf16-b2b2-4fa8-ceb3-9dfe23e47fcb_84088aac-1543-4c61-d407-2d531de79695_ground.jpg} &\\

\end{tabular}
\vspace{-0.2cm}%
\caption{Examples with no crosswalks.  1) ground camera, 2) ground  LiDAR, 3) predicted inverse distance transform, 4) predicted segmentation, 5) predicted crosswalk polygons, 6) gt polygons. }%
\vspace{0.2cm}%
\label{fig:fp}

\includegraphics[width=\textwidth]{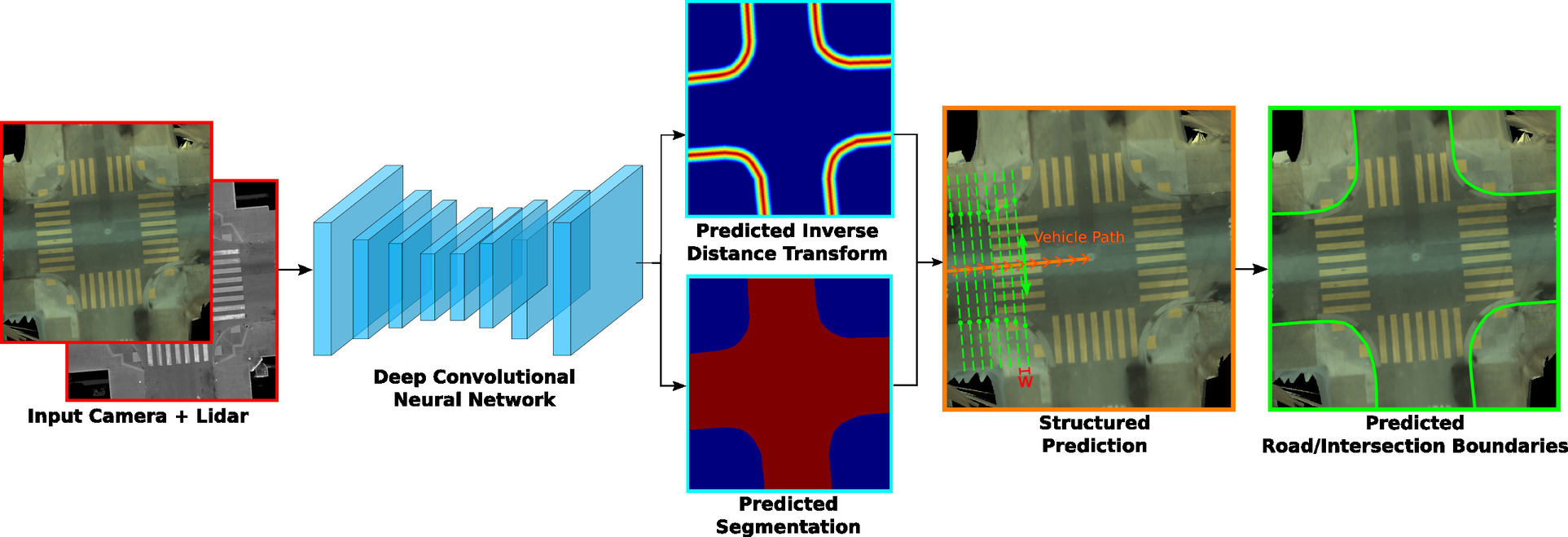}
\vspace{-4mm}
\caption{Generalization of our method to road/intersection boundary prediction.}
\label{fig:generalization}
\vspace{-5mm}
\end{figure}

In this paper we have proposed a deep structured model that  can leverage LiDAR and camera imagery to draw structured crosswalks.  Our experiments in a large city has shown that 96.6\% automation can be achieved for offline mapping while 91.5\% for online mapping. In the future we plan to extend our approach to estimate crosswalks from satellite images. We also plan to extend our approach to predict other semantic elements present in modern HD maps. For example, we can draw stop lines if we predict one boundary instead of two. We can also use this general approach to tackle road/intersection boundaries as seen in Fig \ref{fig:generalization}. Here the CNN outputs both an inverse distance transform and predicted segmentation. We can use the vehicle’s driving path and at every interval we perform a search perpendicular to the vehicle path for the left and right points of the boundary. This can be further extended to draw the lane boundaries.

\clearpage

\bibliographystyle{splncs04}
\bibliography{ref}

\newpage
\renewcommand{\thesection}{Appendix}
\section{}

\begin{figure}
\begin{tabular}{cccccccr}
\includegraphics[width=.1595\textwidth]{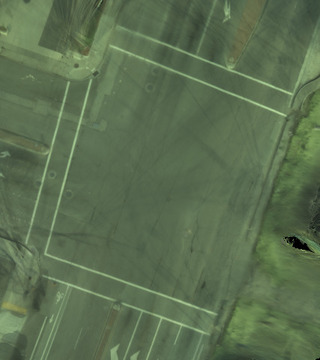} &
\includegraphics[width=.1595\textwidth]{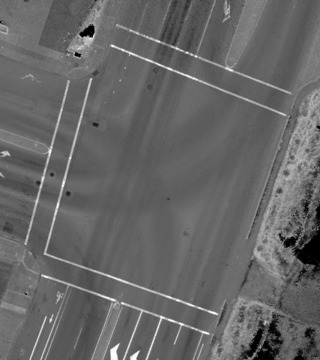} &
\includegraphics[width=.1595\textwidth]{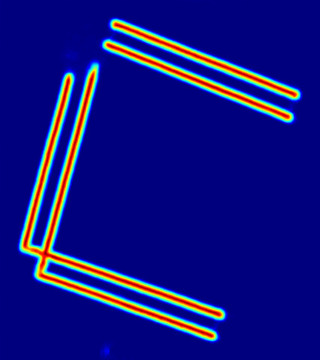} &
\includegraphics[width=.1595\textwidth]{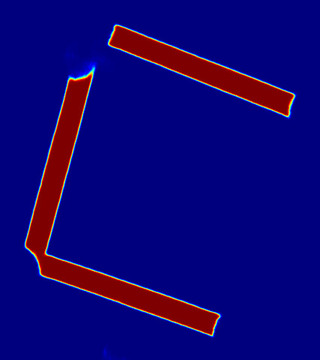} &
\includegraphics[width=.1595\textwidth]{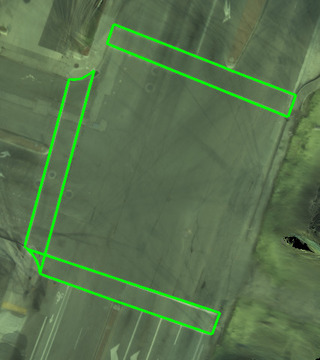} &
\includegraphics[width=.1595\textwidth]{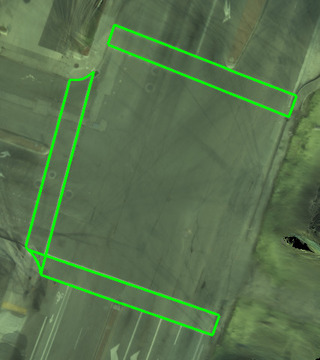} &\\

\includegraphics[width=.1595\textwidth]{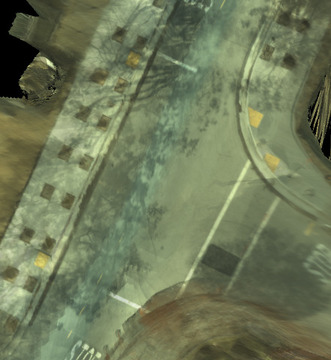} &
\includegraphics[width=.1595\textwidth]{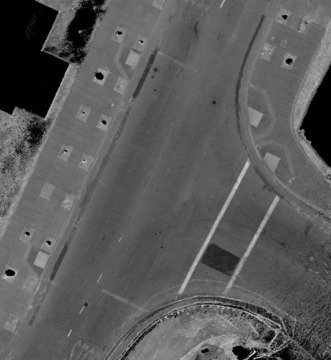} &
\includegraphics[width=.1595\textwidth]{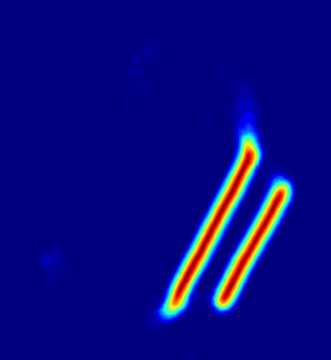} &
\includegraphics[width=.1595\textwidth]{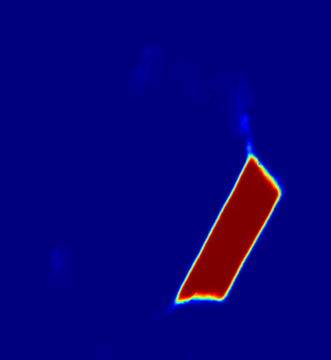} &
\includegraphics[width=.1595\textwidth]{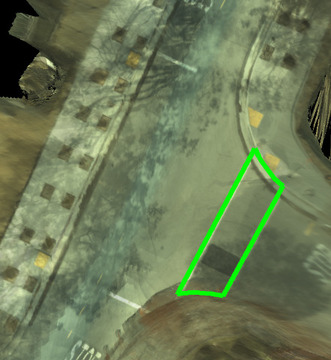} &
\includegraphics[width=.1595\textwidth]{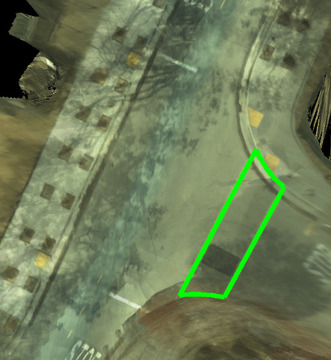} &\\

\includegraphics[width=.1595\textwidth]{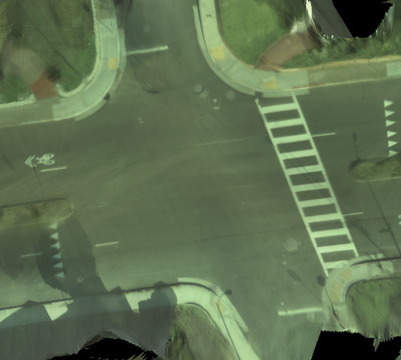} &
\includegraphics[width=.1595\textwidth]{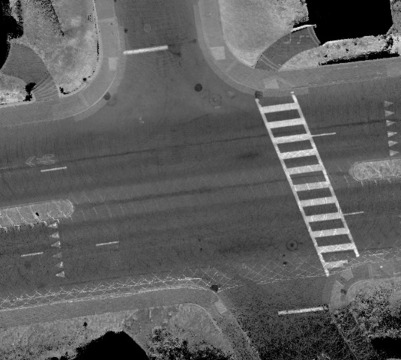} &
\includegraphics[width=.1595\textwidth]{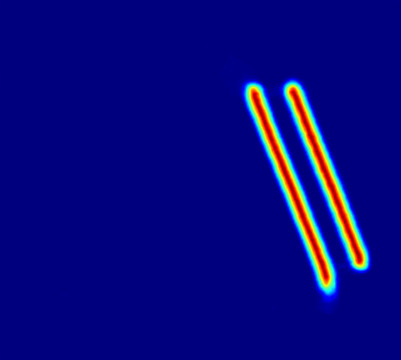} &
\includegraphics[width=.1595\textwidth]{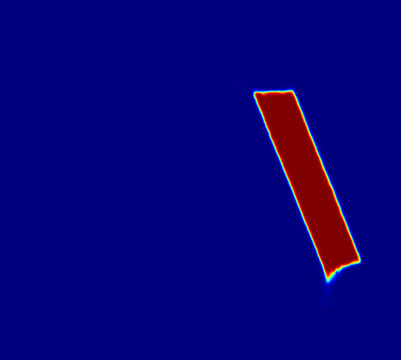} &
\includegraphics[width=.1595\textwidth]{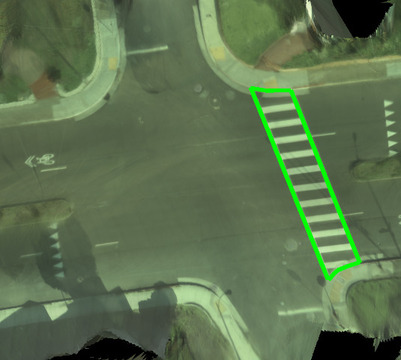} &
\includegraphics[width=.1595\textwidth]{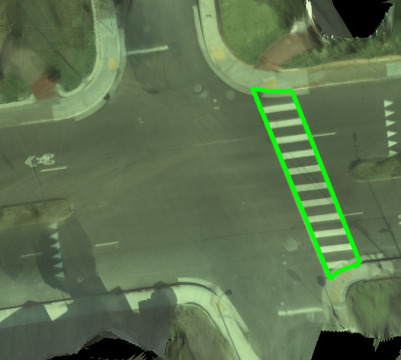} &\\

\includegraphics[width=.1595\textwidth]{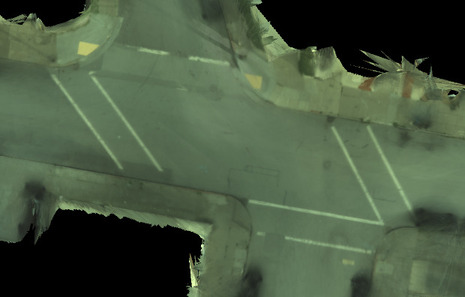} &
\includegraphics[width=.1595\textwidth]{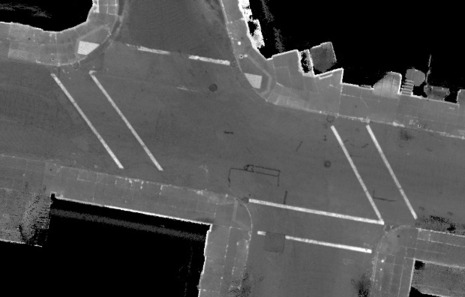} &
\includegraphics[width=.1595\textwidth]{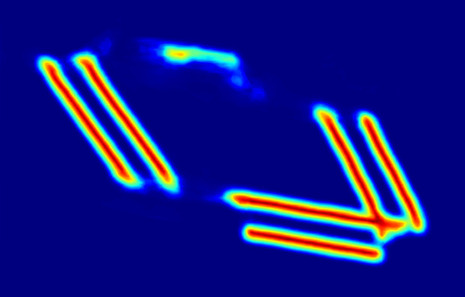} &
\includegraphics[width=.1595\textwidth]{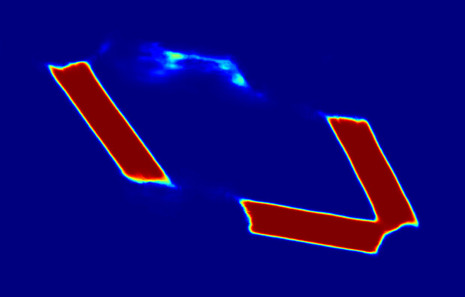} &
\includegraphics[width=.1595\textwidth]{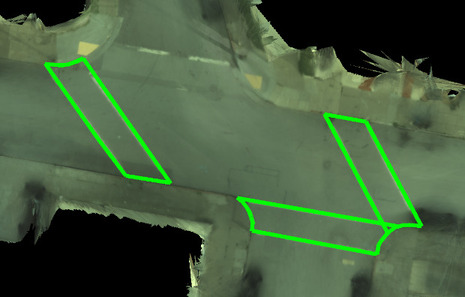} &
\includegraphics[width=.1595\textwidth]{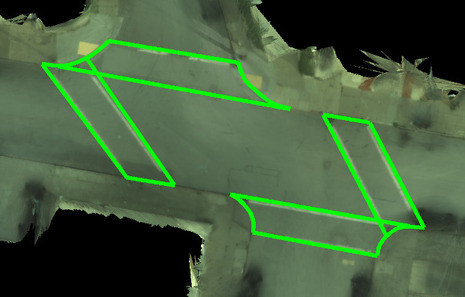} &\\

\includegraphics[width=.1595\textwidth]{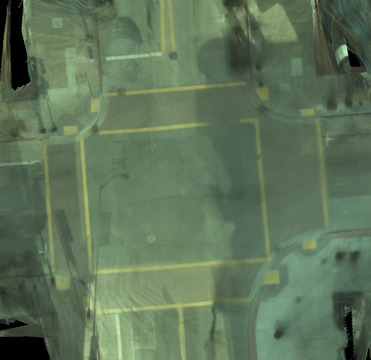} &
\includegraphics[width=.1595\textwidth]{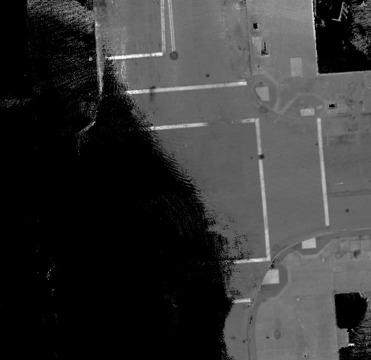} &
\includegraphics[width=.1595\textwidth]{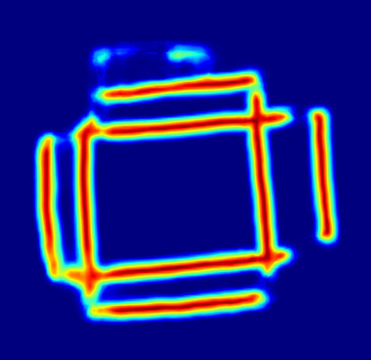} &
\includegraphics[width=.1595\textwidth]{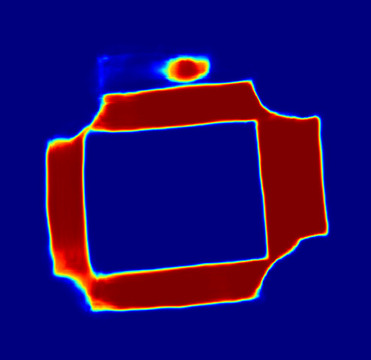} &
\includegraphics[width=.1595\textwidth]{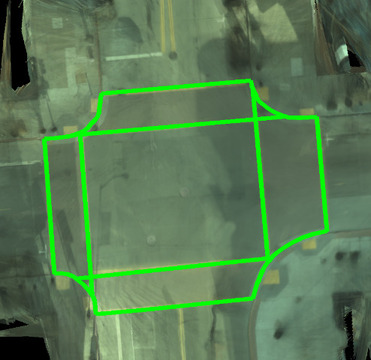} &
\includegraphics[width=.1595\textwidth]{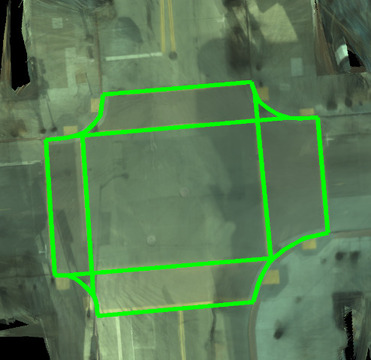} &\\

\includegraphics[width=.1595\textwidth]{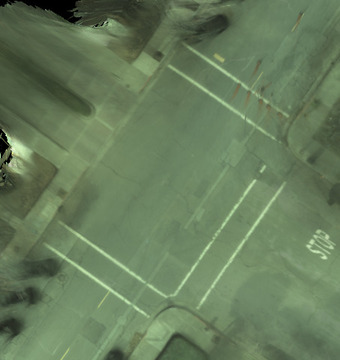} &
\includegraphics[width=.1595\textwidth]{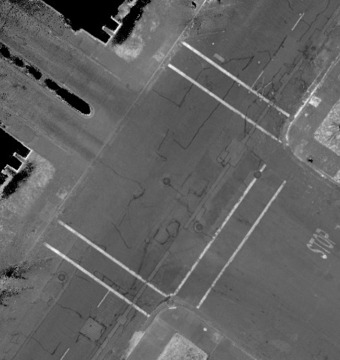} &
\includegraphics[width=.1595\textwidth]{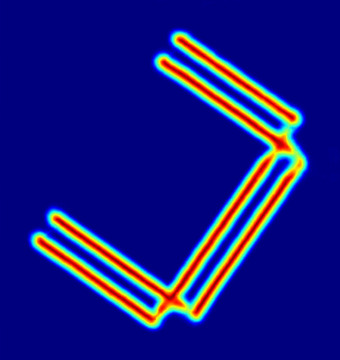} &
\includegraphics[width=.1595\textwidth]{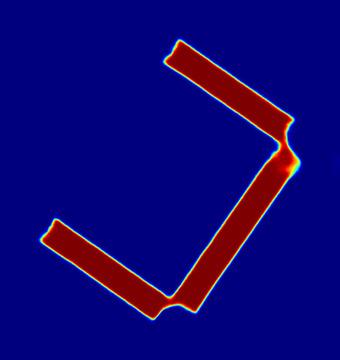} &
\includegraphics[width=.1595\textwidth]{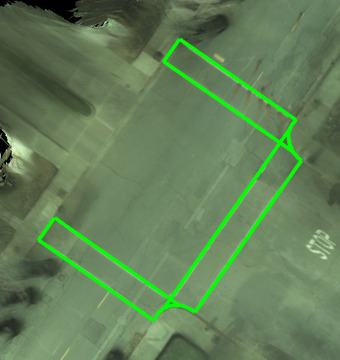} &
\includegraphics[width=.1595\textwidth]{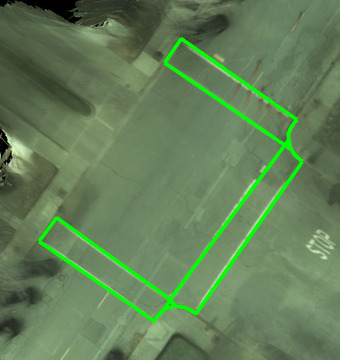} &\\

\includegraphics[width=.1595\textwidth]{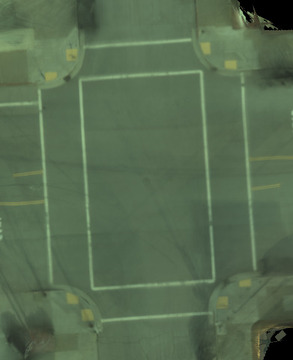} &
\includegraphics[width=.1595\textwidth]{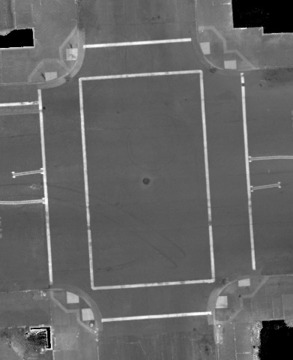} &
\includegraphics[width=.1595\textwidth]{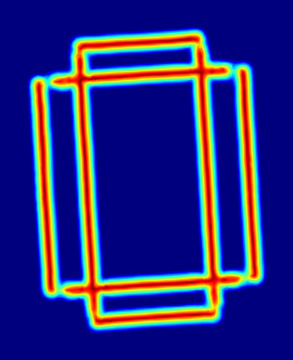} &
\includegraphics[width=.1595\textwidth]{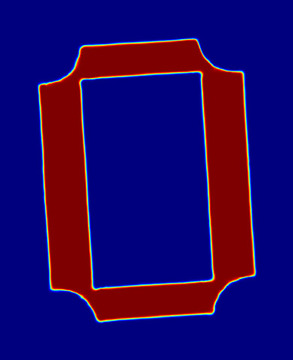} &
\includegraphics[width=.1595\textwidth]{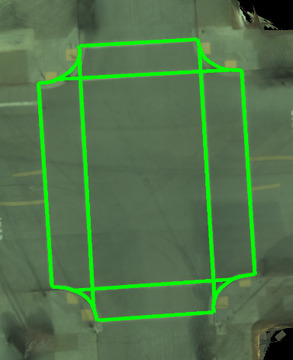} &
\includegraphics[width=.1595\textwidth]{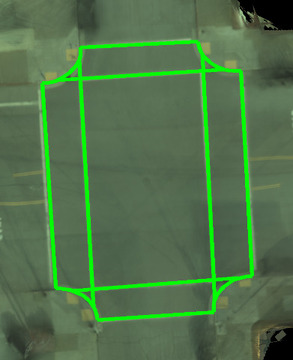} &\\

\includegraphics[width=.1595\textwidth]{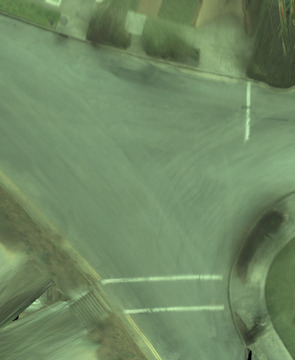} &
\includegraphics[width=.1595\textwidth]{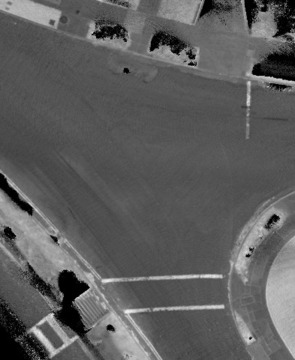} &
\includegraphics[width=.1595\textwidth]{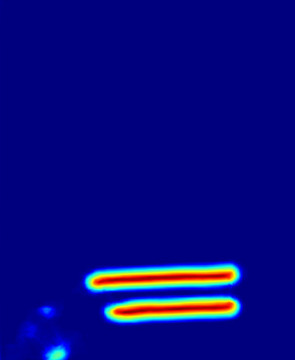} &
\includegraphics[width=.1595\textwidth]{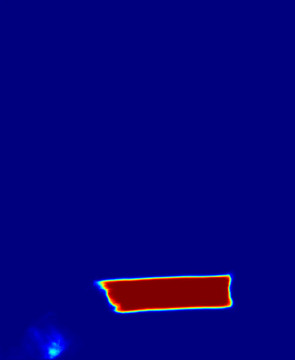} &
\includegraphics[width=.1595\textwidth]{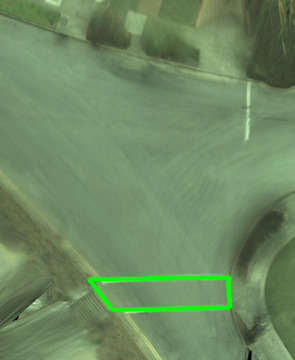} &
\includegraphics[width=.1595\textwidth]{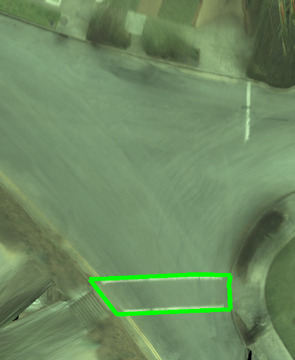} &\\

a) & b) & c) & d) & e) & f) & \\
\end{tabular}

\caption{Offline map model (multiple passes) results using the the model trained on both camera and LiDAR. Comparisons between col a) input ground camera, b) input ground lidar, c) predicted inverse distance transform, d) predicted segmentation, e) our predicted crosswalk polygons after inference and f) ground truth crosswalk polygons.}
\label{fig:offline_1}

\end{figure}

\begin{figure}
\begin{tabular}{cccccccr}

\includegraphics[width=.1595\textwidth]{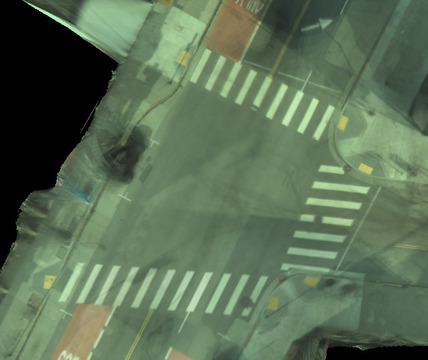} &
\includegraphics[width=.1595\textwidth]{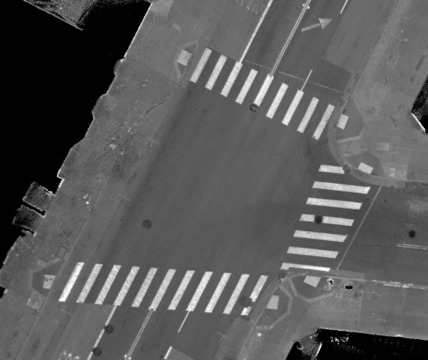} &
\includegraphics[width=.1595\textwidth]{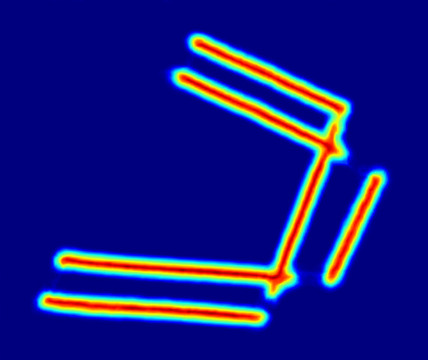} &
\includegraphics[width=.1595\textwidth]{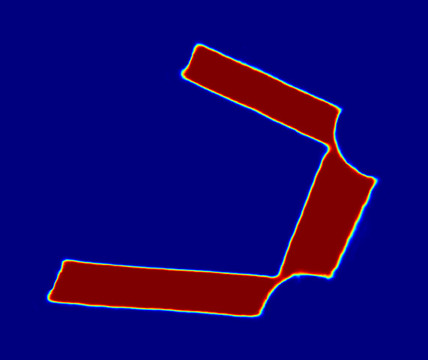} &
\includegraphics[width=.1595\textwidth]{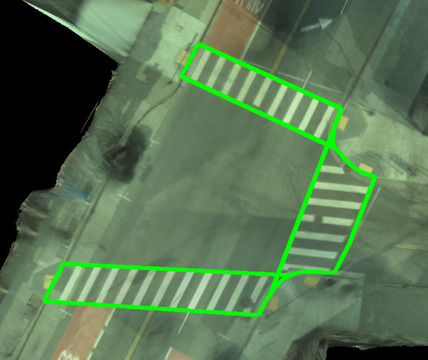} &
\includegraphics[width=.1595\textwidth]{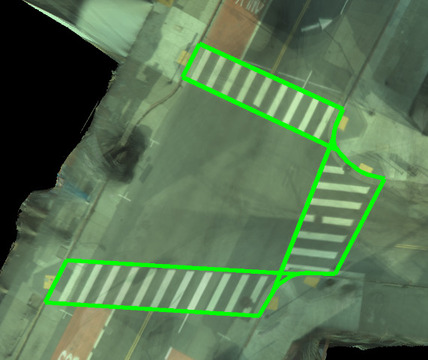} &\\

\includegraphics[width=.1595\textwidth]{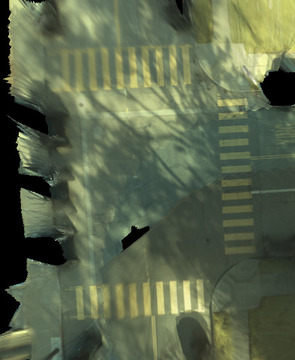} &
\includegraphics[width=.1595\textwidth]{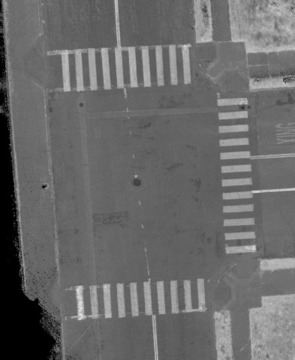} &
\includegraphics[width=.1595\textwidth]{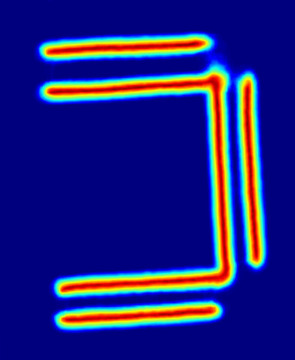} &
\includegraphics[width=.1595\textwidth]{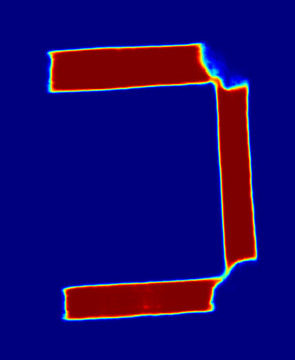} &
\includegraphics[width=.1595\textwidth]{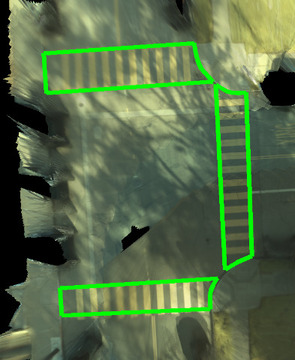} &
\includegraphics[width=.1595\textwidth]{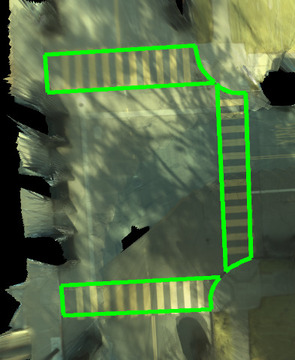} &\\

\includegraphics[width=.1595\textwidth]{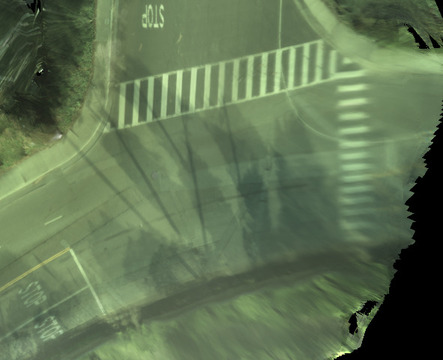} &
\includegraphics[width=.1595\textwidth]{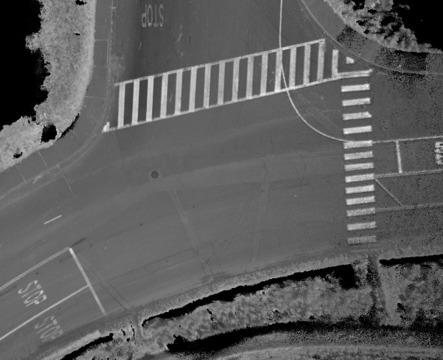} &
\includegraphics[width=.1595\textwidth]{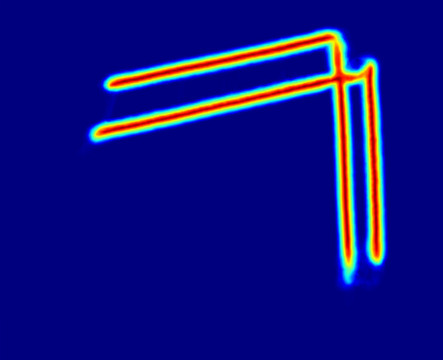} &
\includegraphics[width=.1595\textwidth]{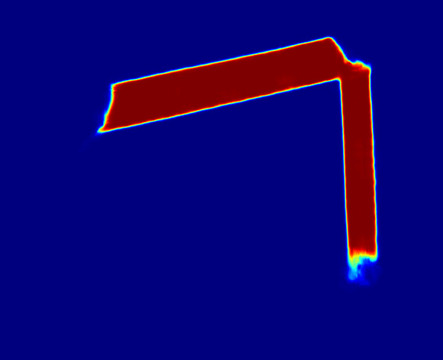} &
\includegraphics[width=.1595\textwidth]{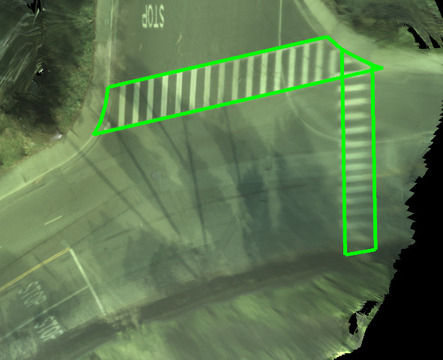} &
\includegraphics[width=.1595\textwidth]{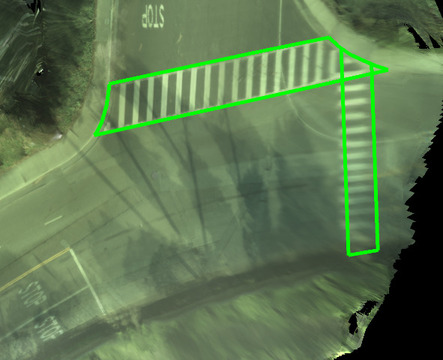} &\\

\includegraphics[width=.1595\textwidth]{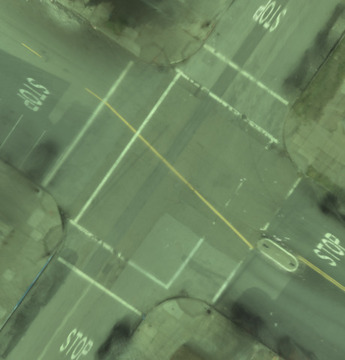} &
\includegraphics[width=.1595\textwidth]{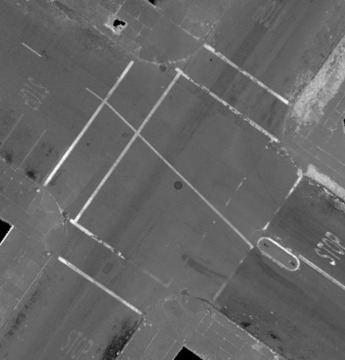} &
\includegraphics[width=.1595\textwidth]{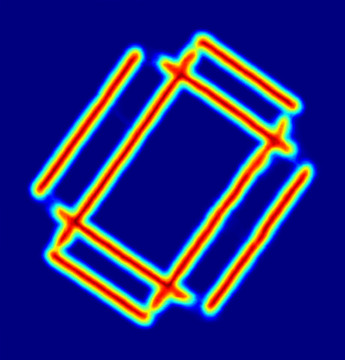} &
\includegraphics[width=.1595\textwidth]{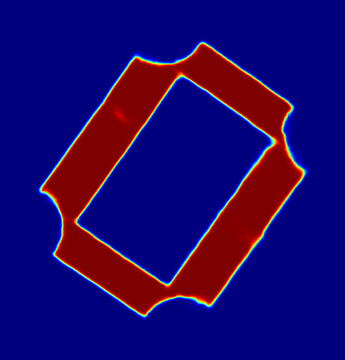} &
\includegraphics[width=.1595\textwidth]{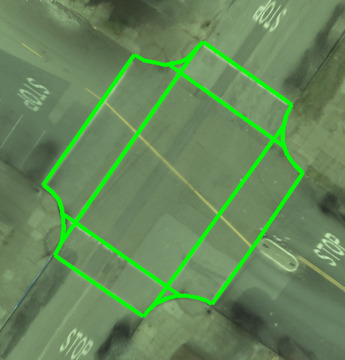} &
\includegraphics[width=.1595\textwidth]{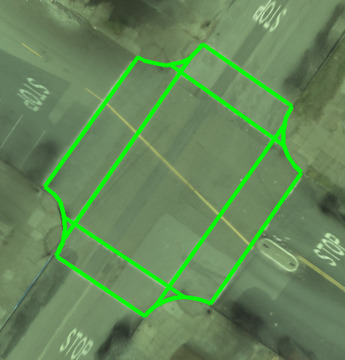} &\\

\includegraphics[width=.1595\textwidth]{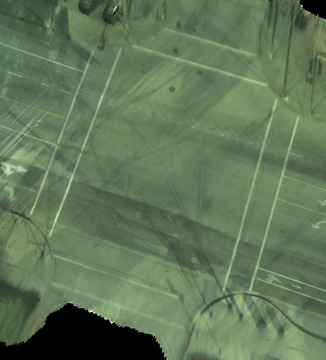} &
\includegraphics[width=.1595\textwidth]{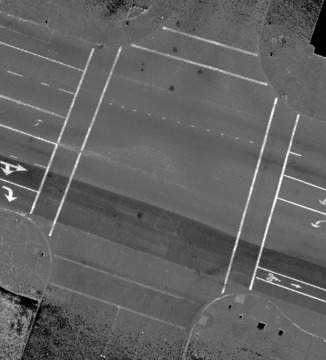} &
\includegraphics[width=.1595\textwidth]{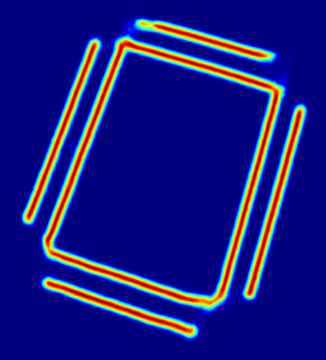} &
\includegraphics[width=.1595\textwidth]{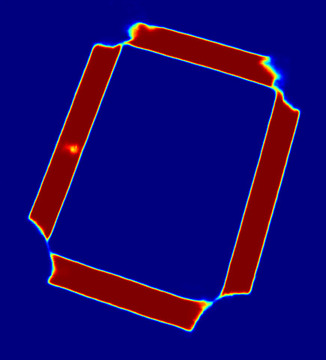} &
\includegraphics[width=.1595\textwidth]{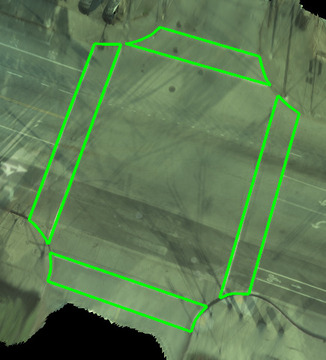} &
\includegraphics[width=.1595\textwidth]{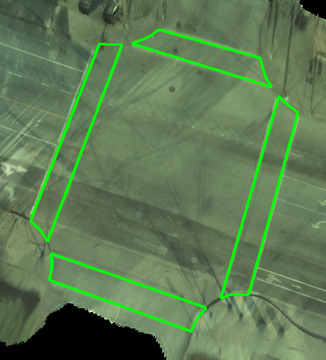} &\\

\includegraphics[width=.1595\textwidth]{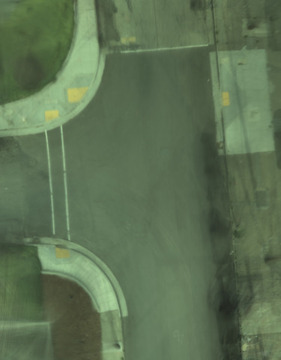} &
\includegraphics[width=.1595\textwidth]{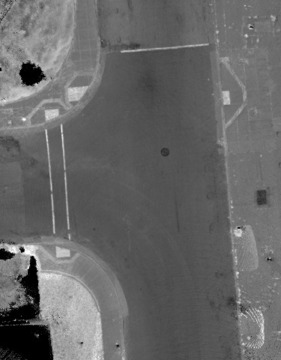} &
\includegraphics[width=.1595\textwidth]{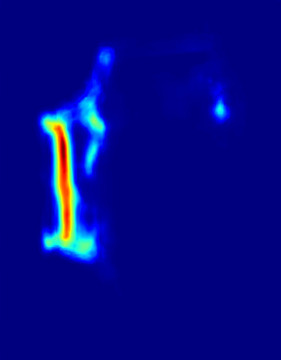} &
\includegraphics[width=.1595\textwidth]{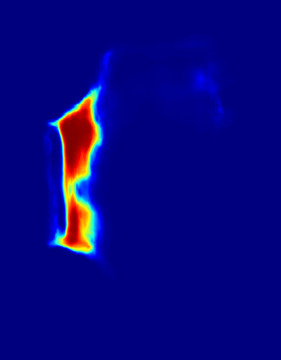} &
\includegraphics[width=.1595\textwidth]{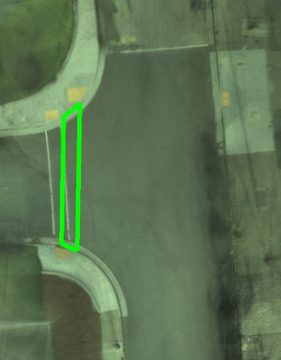} &
\includegraphics[width=.1595\textwidth]{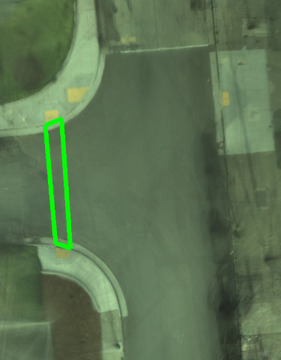} &\\

\includegraphics[width=.1595\textwidth]{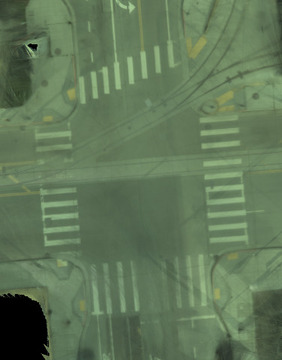} &
\includegraphics[width=.1595\textwidth]{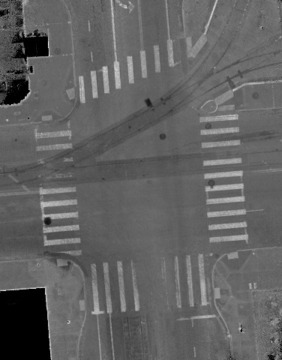} &
\includegraphics[width=.1595\textwidth]{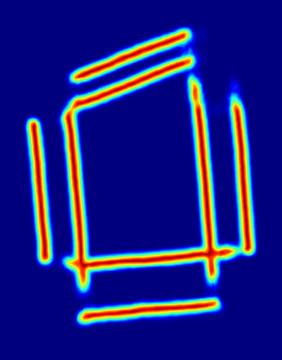} &
\includegraphics[width=.1595\textwidth]{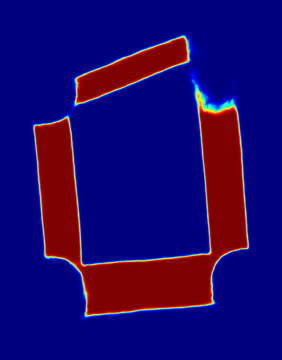} &
\includegraphics[width=.1595\textwidth]{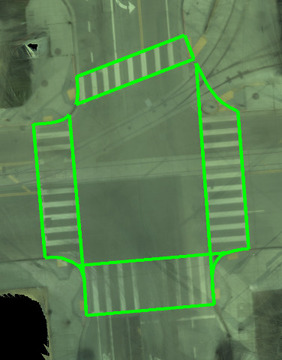} &
\includegraphics[width=.1595\textwidth]{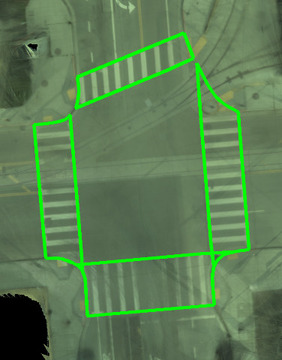} &\\

\includegraphics[width=.1595\textwidth]{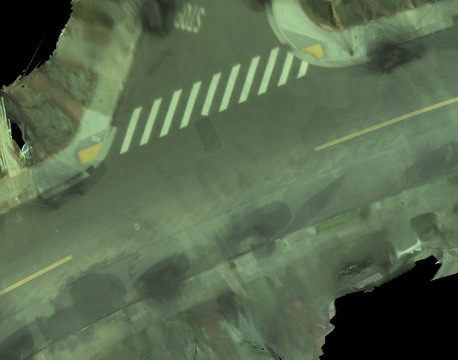} &
\includegraphics[width=.1595\textwidth]{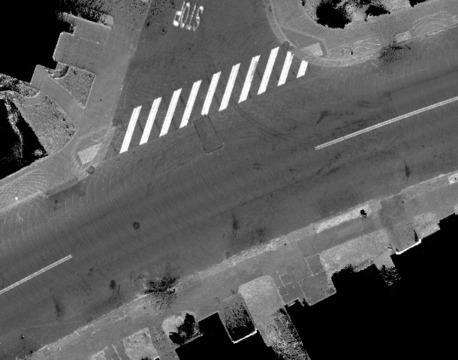} &
\includegraphics[width=.1595\textwidth]{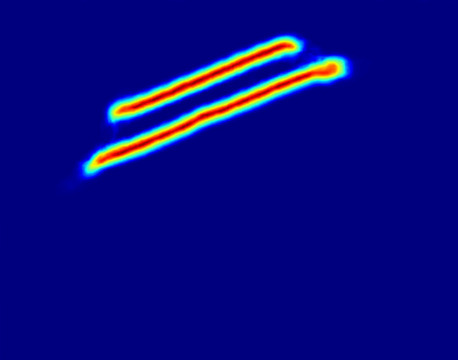} &
\includegraphics[width=.1595\textwidth]{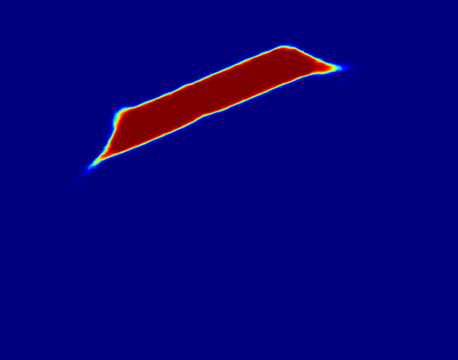} &
\includegraphics[width=.1595\textwidth]{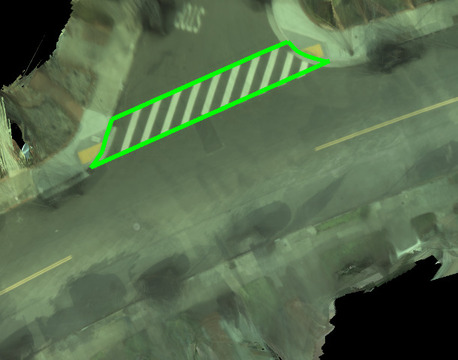} &
\includegraphics[width=.1595\textwidth]{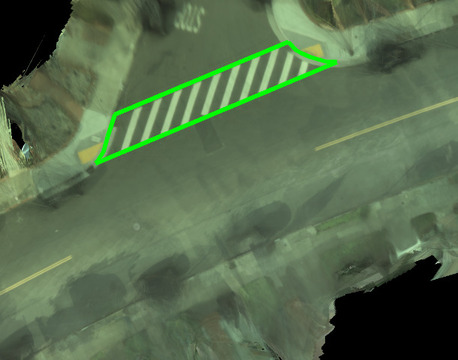} &\\

a) & b) & c) & d) & e) & f) & \\
\end{tabular}

\caption{Offline map model (multiple passes) results using the the model trained on both camera and LiDAR. Comparisons between col a) input ground camera, b) input ground lidar, c) predicted inverse distance transform, d) predicted segmentation, e) our predicted crosswalk polygons after inference and f) ground truth crosswalk polygons.}
\label{fig:offline_2}

\end{figure}

\begin{figure}
\begin{tabular}{cccccccr}
\includegraphics[width=.1595\textwidth]{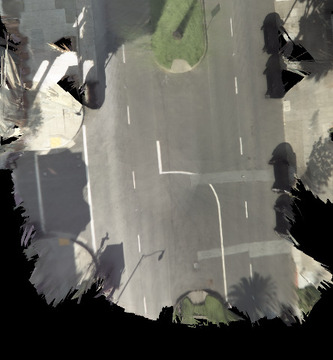} &
\includegraphics[width=.1595\textwidth]{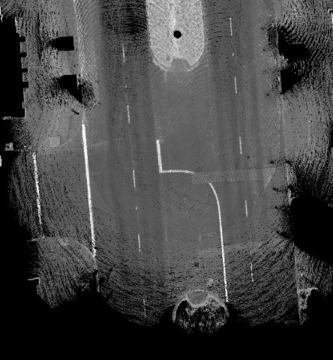} &
\includegraphics[width=.1595\textwidth]{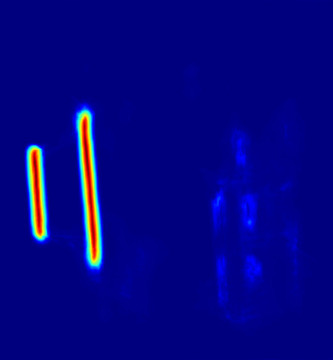} &
\includegraphics[width=.1595\textwidth]{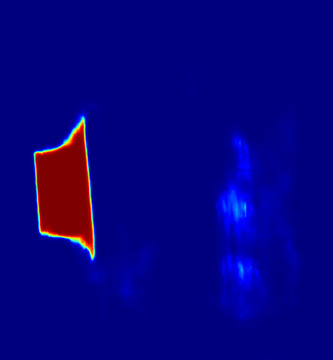} &
\includegraphics[width=.1595\textwidth]{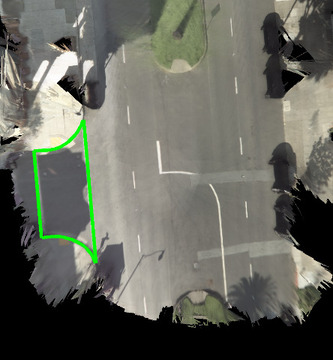} &
\includegraphics[width=.1595\textwidth]{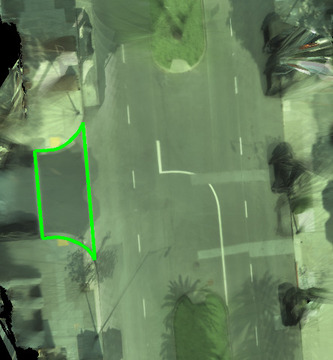} &\\

\includegraphics[width=.1595\textwidth]{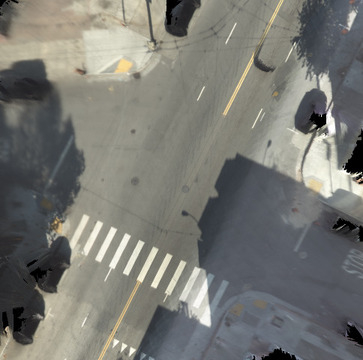} &
\includegraphics[width=.1595\textwidth]{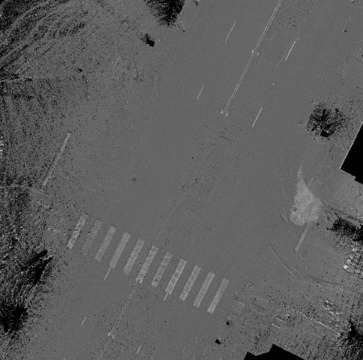} &
\includegraphics[width=.1595\textwidth]{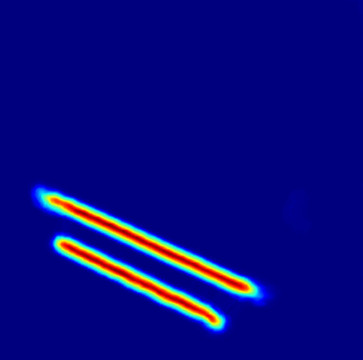} &
\includegraphics[width=.1595\textwidth]{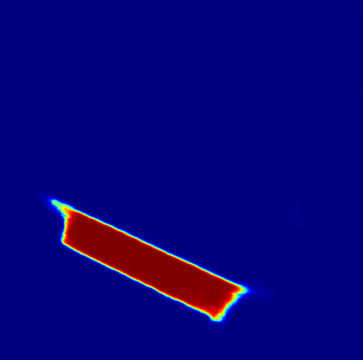} &
\includegraphics[width=.1595\textwidth]{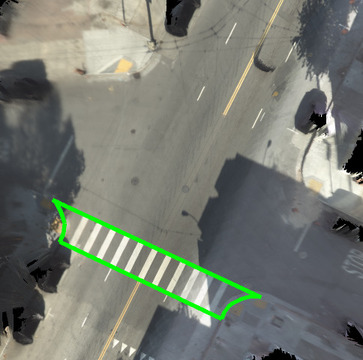} &
\includegraphics[width=.1595\textwidth]{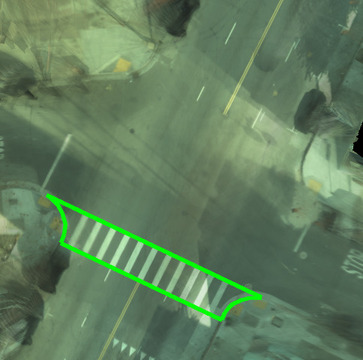} &\\

\includegraphics[width=.1595\textwidth]{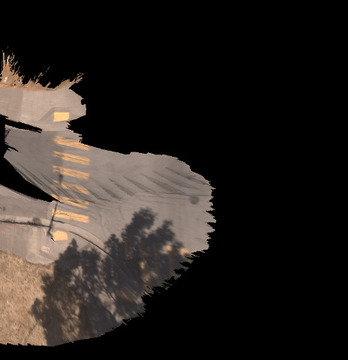} &
\includegraphics[width=.1595\textwidth]{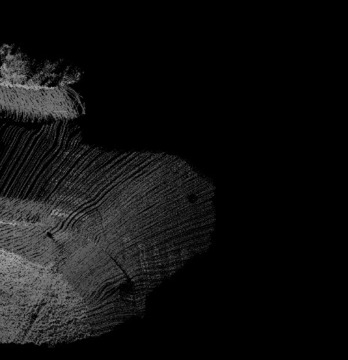} &
\includegraphics[width=.1595\textwidth]{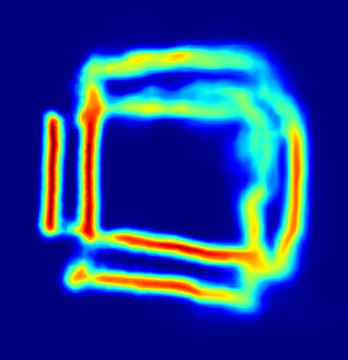} &
\includegraphics[width=.1595\textwidth]{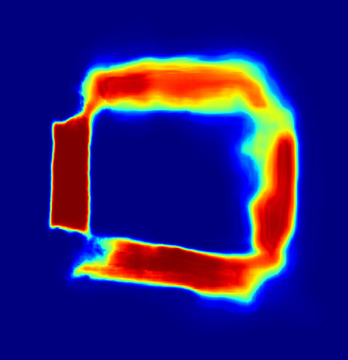} &
\includegraphics[width=.1595\textwidth]{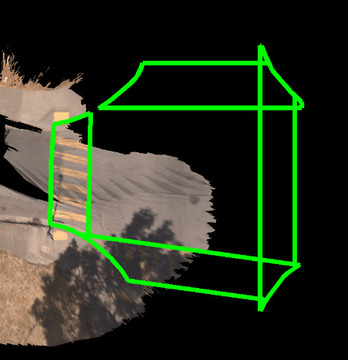} &
\includegraphics[width=.1595\textwidth]{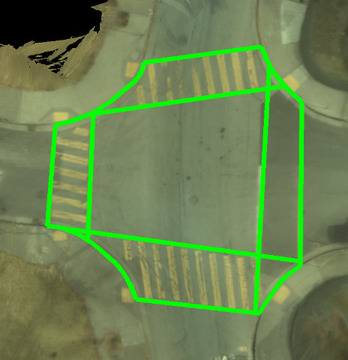} &\\

\includegraphics[width=.1595\textwidth]{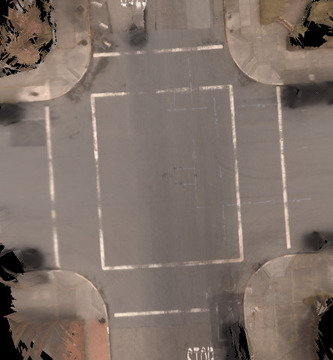} &
\includegraphics[width=.1595\textwidth]{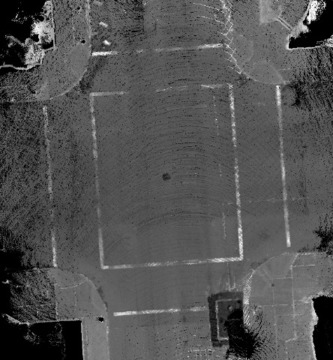} &
\includegraphics[width=.1595\textwidth]{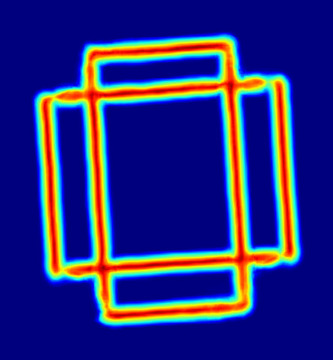} &
\includegraphics[width=.1595\textwidth]{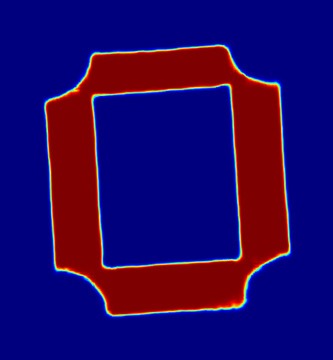} &
\includegraphics[width=.1595\textwidth]{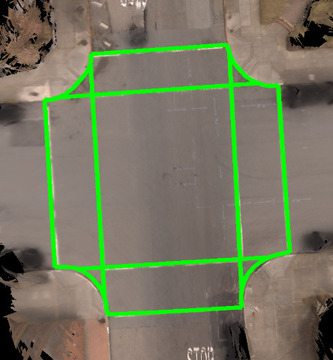} &
\includegraphics[width=.1595\textwidth]{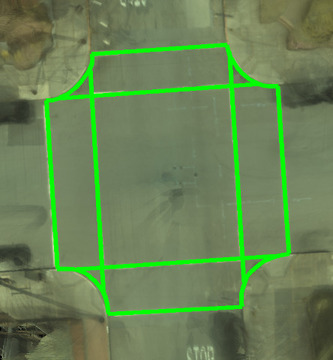} &\\

\includegraphics[width=.1595\textwidth]{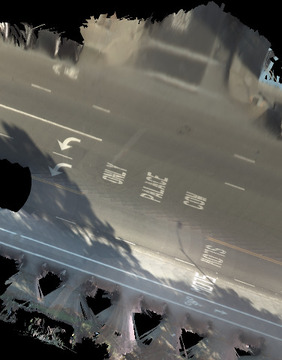} &
\includegraphics[width=.1595\textwidth]{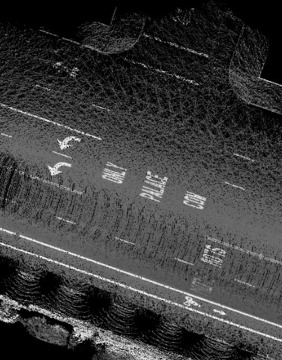} &
\includegraphics[width=.1595\textwidth]{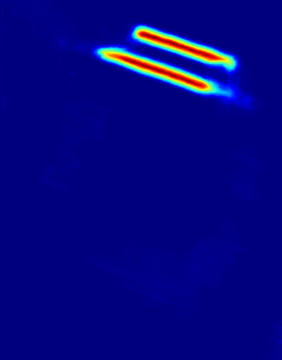} &
\includegraphics[width=.1595\textwidth]{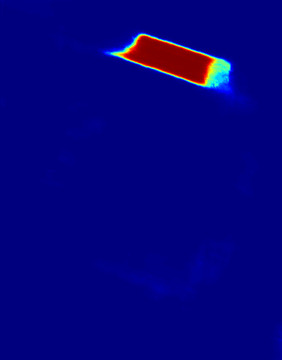} &
\includegraphics[width=.1595\textwidth]{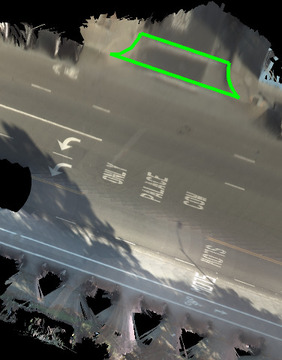} &
\includegraphics[width=.1595\textwidth]{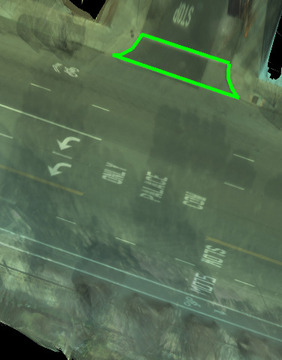} &\\

\includegraphics[width=.1595\textwidth]{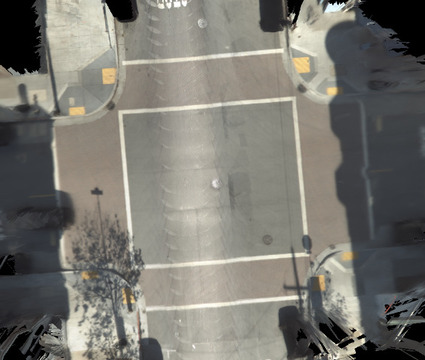} &
\includegraphics[width=.1595\textwidth]{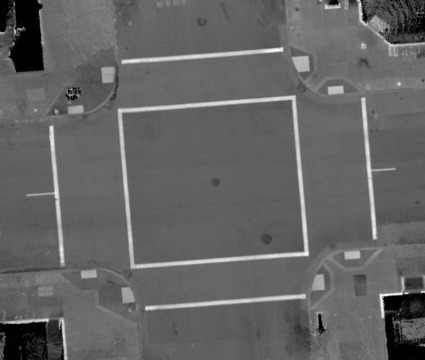} &
\includegraphics[width=.1595\textwidth]{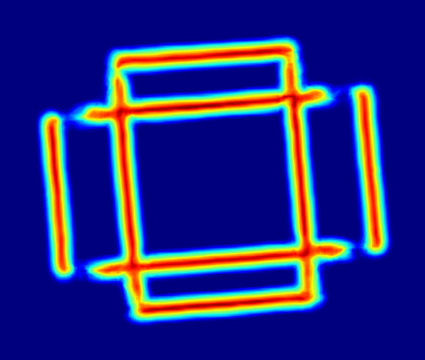} &
\includegraphics[width=.1595\textwidth]{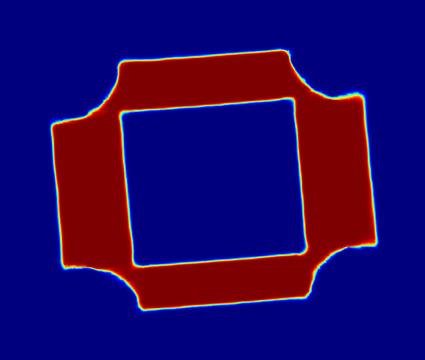} &
\includegraphics[width=.1595\textwidth]{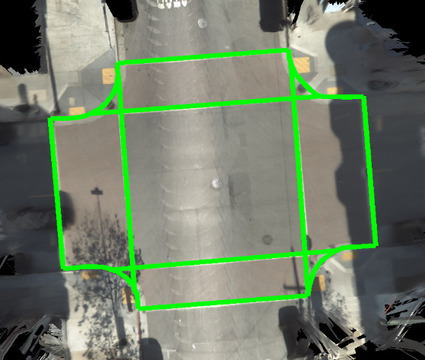} &
\includegraphics[width=.1595\textwidth]{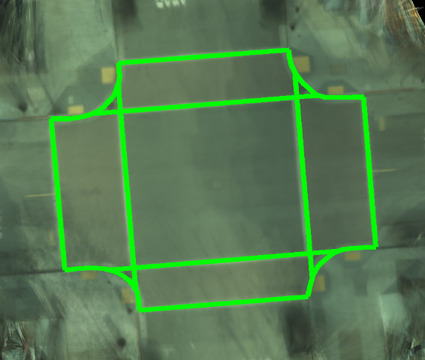} &\\

\includegraphics[width=.1595\textwidth]{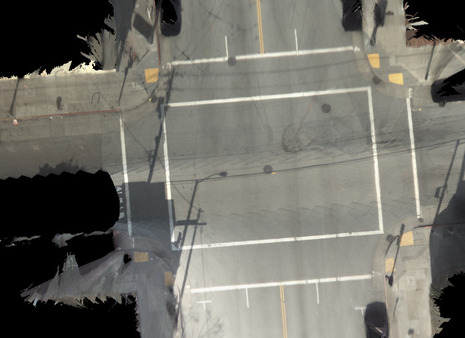} &
\includegraphics[width=.1595\textwidth]{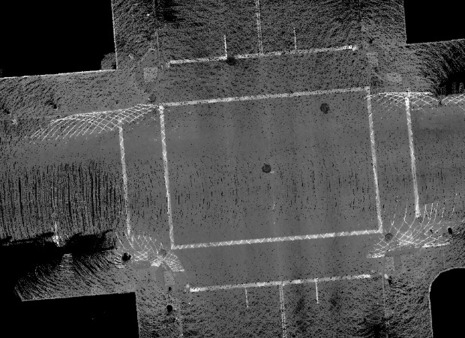} &
\includegraphics[width=.1595\textwidth]{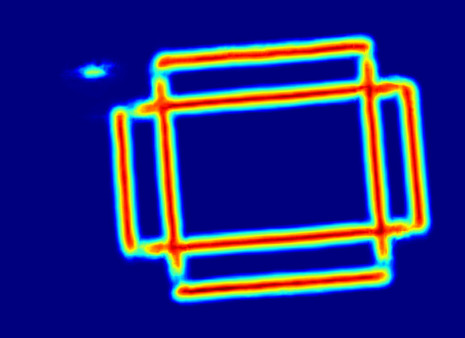} &
\includegraphics[width=.1595\textwidth]{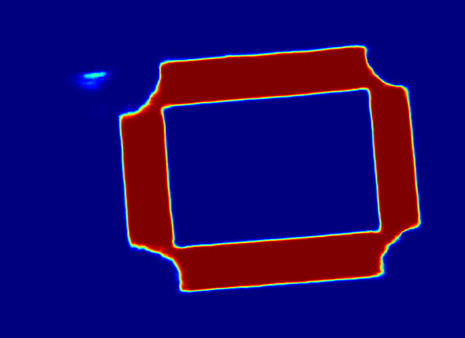} &
\includegraphics[width=.1595\textwidth]{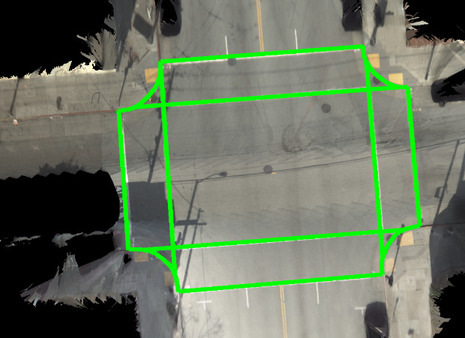} &
\includegraphics[width=.1595\textwidth]{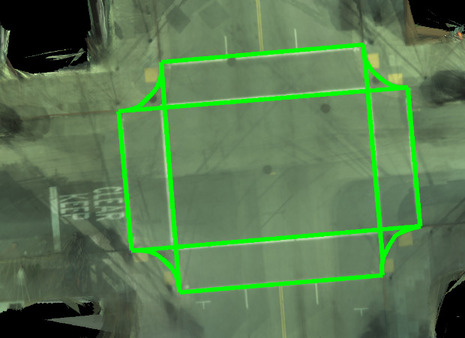} &\\

\includegraphics[width=.1595\textwidth]{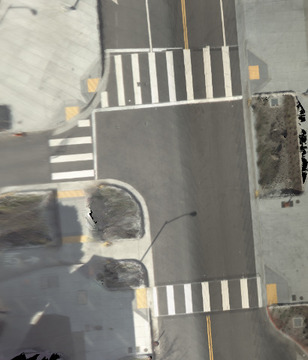} &
\includegraphics[width=.1595\textwidth]{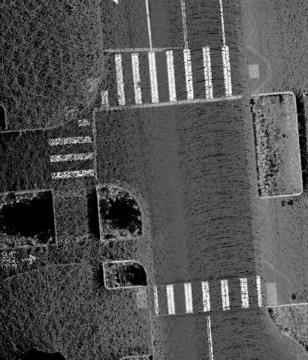} &
\includegraphics[width=.1595\textwidth]{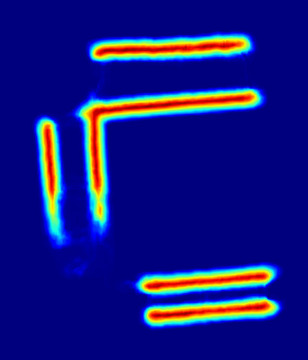} &
\includegraphics[width=.1595\textwidth]{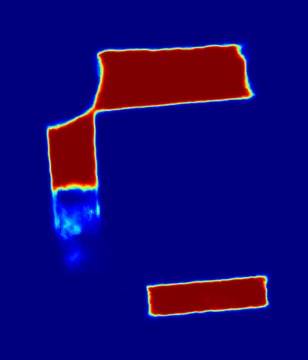} &
\includegraphics[width=.1595\textwidth]{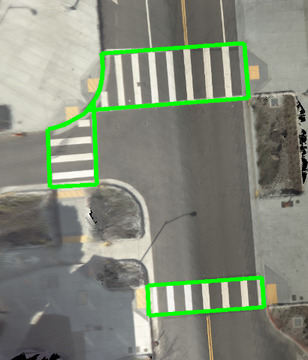} &
\includegraphics[width=.1595\textwidth]{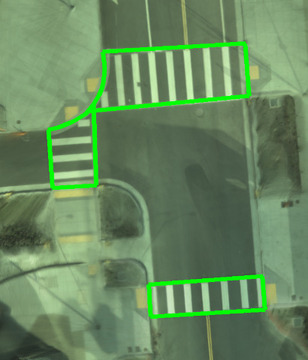} &\\

a) & b) & c) & d) & e) & f) & \\
\end{tabular}

\caption{Online map model (single pass) results using the model trained on both camera and LiDAR imagery. Comparisons between col a) input ground camera (online map), b) input ground LiDAR (online map), c) predicted inverse distance transform, d) predicted segmentation, e) our predicted crosswalk polygons after inference and f) ground truth crosswalk polygons overlayed on the ground RGB (offline map).}
\label{fig:online_1}

\end{figure}

\begin{figure}
\begin{tabular}{cccccccr}
\includegraphics[width=.1595\textwidth]{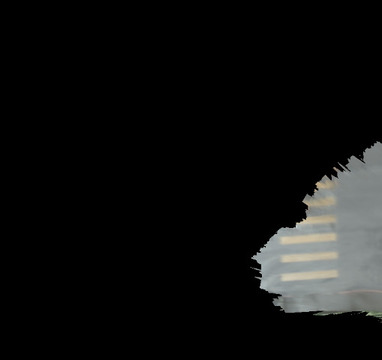} &
\includegraphics[width=.1595\textwidth]{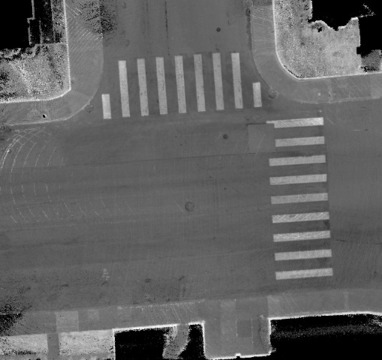} &
\includegraphics[width=.1595\textwidth]{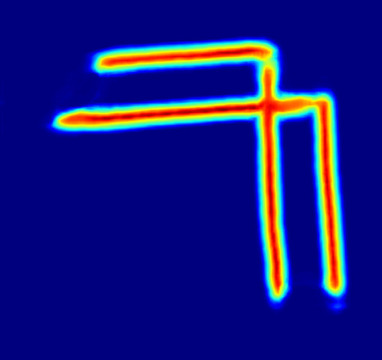} &
\includegraphics[width=.1595\textwidth]{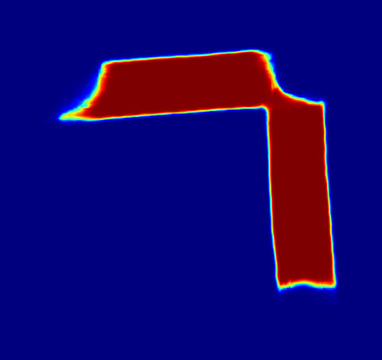} &
\includegraphics[width=.1595\textwidth]{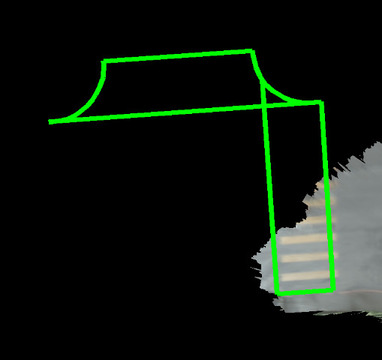} &
\includegraphics[width=.1595\textwidth]{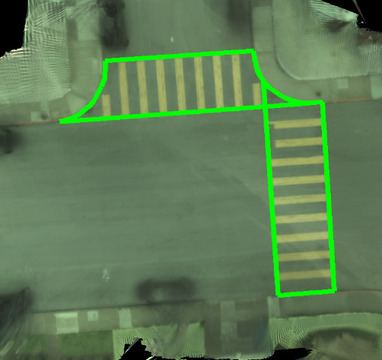} &\\

\includegraphics[width=.1595\textwidth]{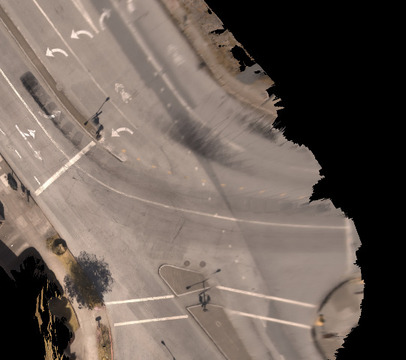} &
\includegraphics[width=.1595\textwidth]{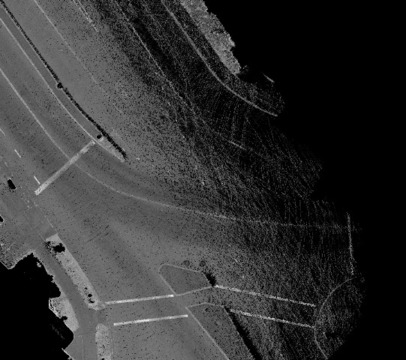} &
\includegraphics[width=.1595\textwidth]{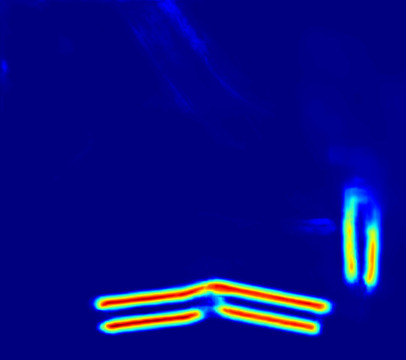} &
\includegraphics[width=.1595\textwidth]{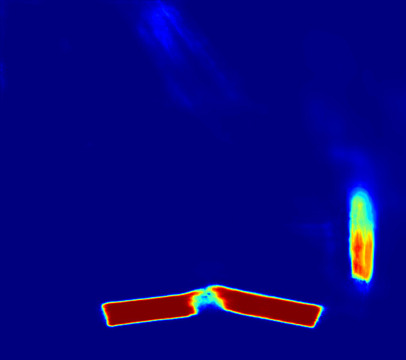} &
\includegraphics[width=.1595\textwidth]{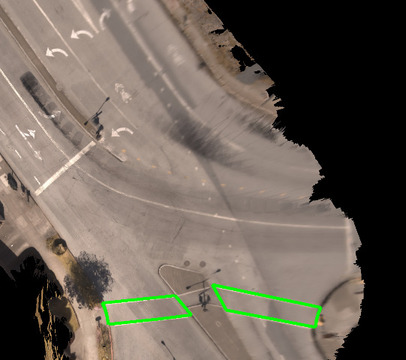} &
\includegraphics[width=.1595\textwidth]{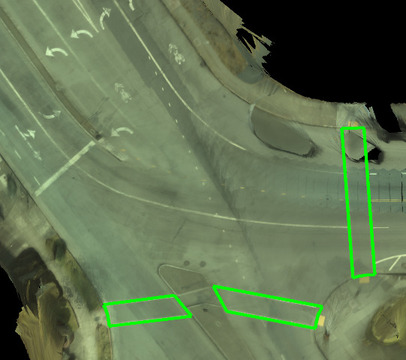} &\\

\includegraphics[width=.1595\textwidth]{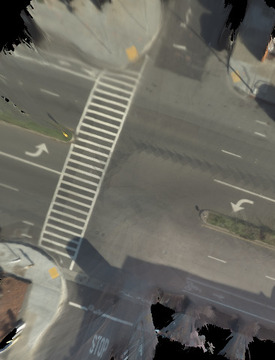} &
\includegraphics[width=.1595\textwidth]{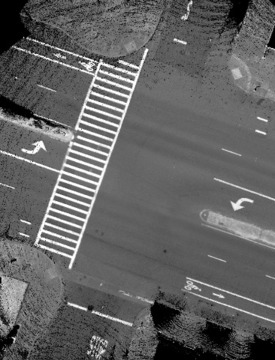} &
\includegraphics[width=.1595\textwidth]{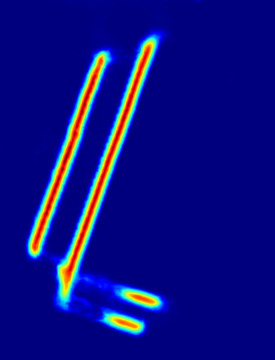} &
\includegraphics[width=.1595\textwidth]{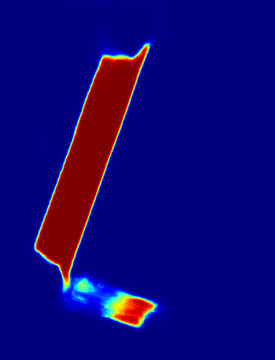} &
\includegraphics[width=.1595\textwidth]{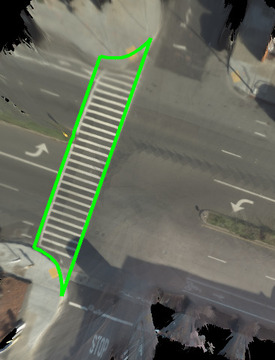} &
\includegraphics[width=.1595\textwidth]{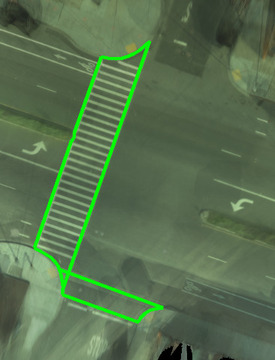} &\\

\includegraphics[width=.1595\textwidth]{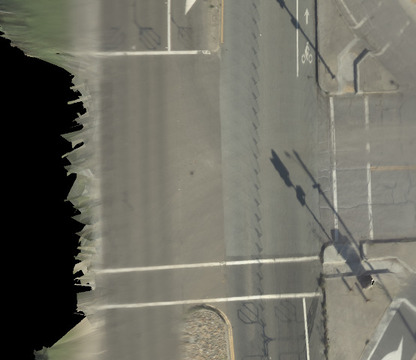} &
\includegraphics[width=.1595\textwidth]{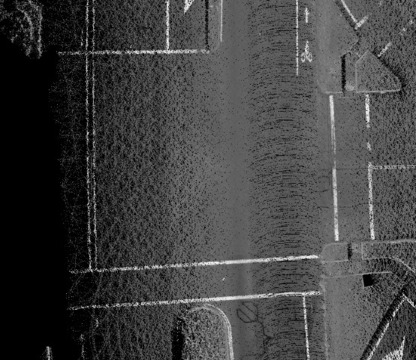} &
\includegraphics[width=.1595\textwidth]{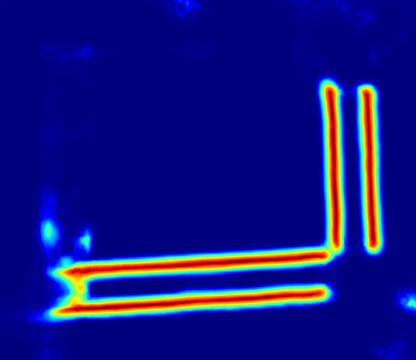} &
\includegraphics[width=.1595\textwidth]{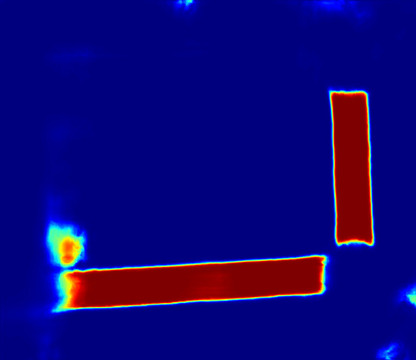} &
\includegraphics[width=.1595\textwidth]{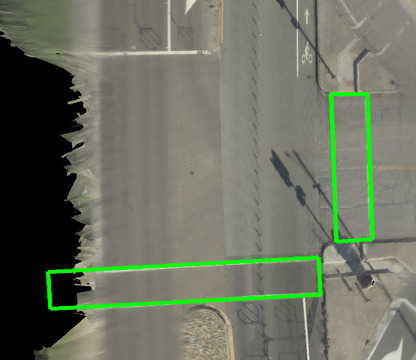} &
\includegraphics[width=.1595\textwidth]{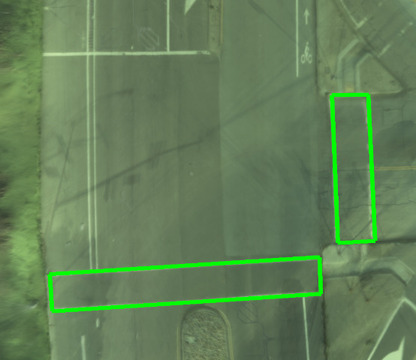} &\\

\includegraphics[width=.1595\textwidth]{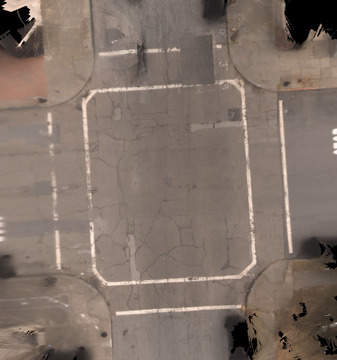} &
\includegraphics[width=.1595\textwidth]{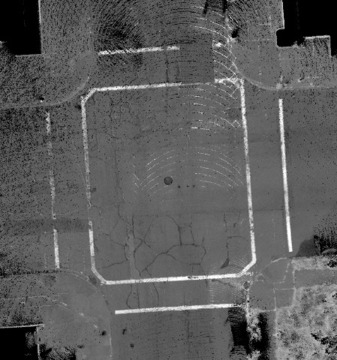} &
\includegraphics[width=.1595\textwidth]{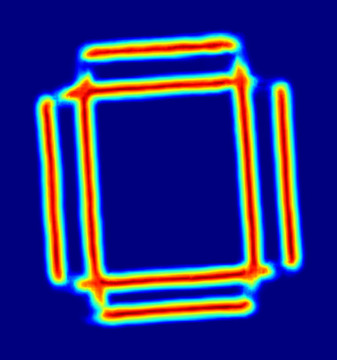} &
\includegraphics[width=.1595\textwidth]{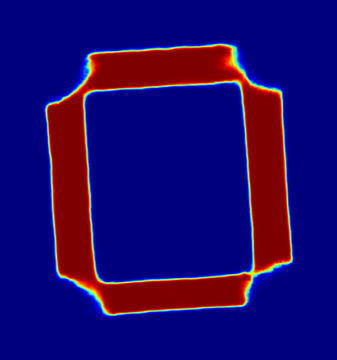} &
\includegraphics[width=.1595\textwidth]{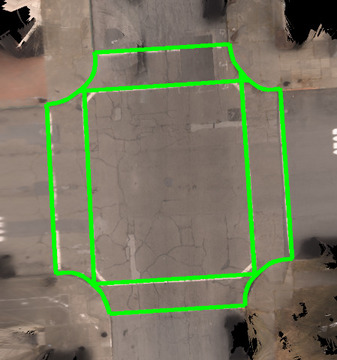} &
\includegraphics[width=.1595\textwidth]{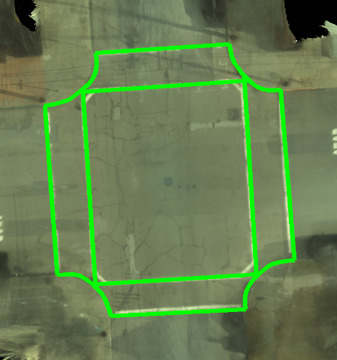} &\\

\includegraphics[width=.1595\textwidth]{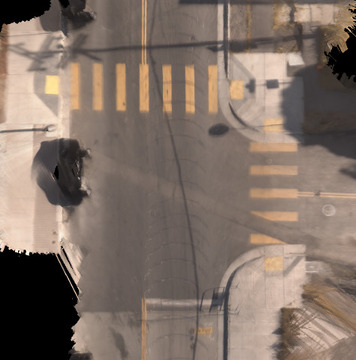} &
\includegraphics[width=.1595\textwidth]{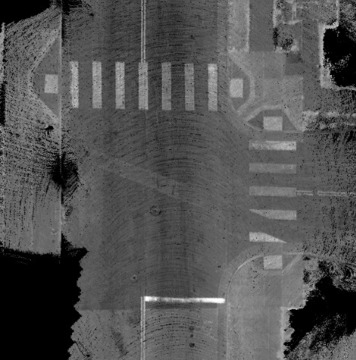} &
\includegraphics[width=.1595\textwidth]{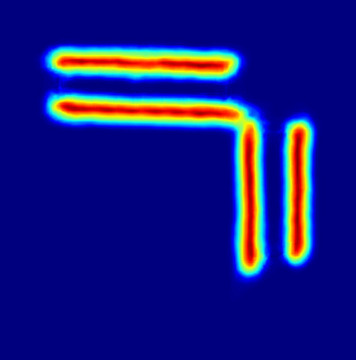} &
\includegraphics[width=.1595\textwidth]{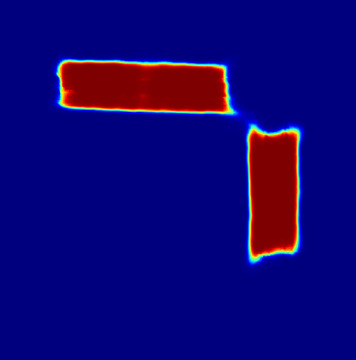} &
\includegraphics[width=.1595\textwidth]{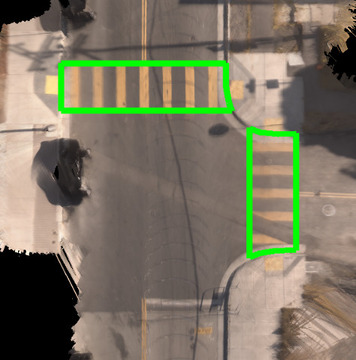} &
\includegraphics[width=.1595\textwidth]{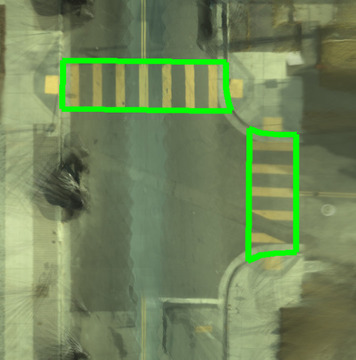} &\\

\includegraphics[width=.1595\textwidth]{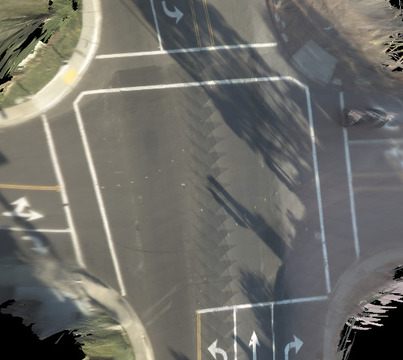} &
\includegraphics[width=.1595\textwidth]{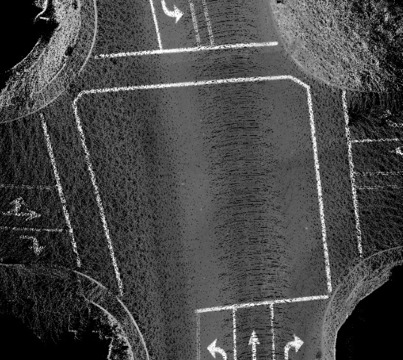} &
\includegraphics[width=.1595\textwidth]{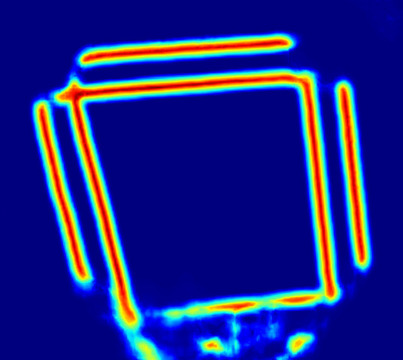} &
\includegraphics[width=.1595\textwidth]{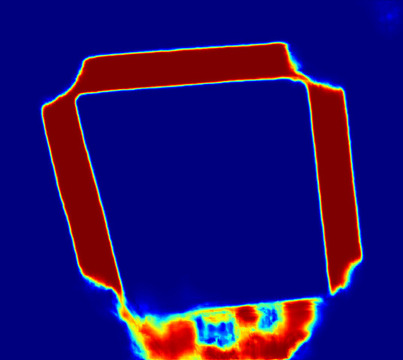} &
\includegraphics[width=.1595\textwidth]{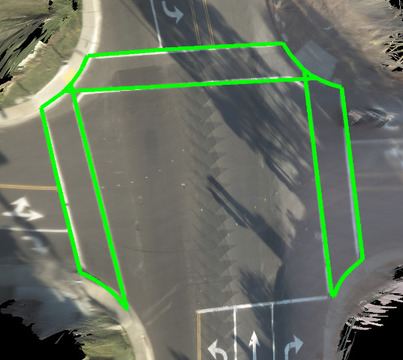} &
\includegraphics[width=.1595\textwidth]{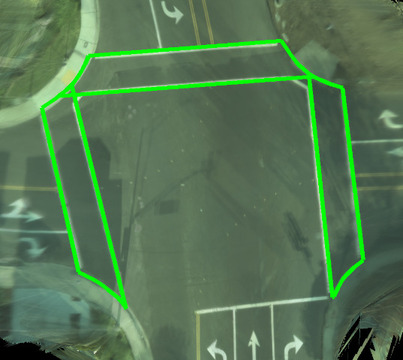} &\\

\includegraphics[width=.1595\textwidth]{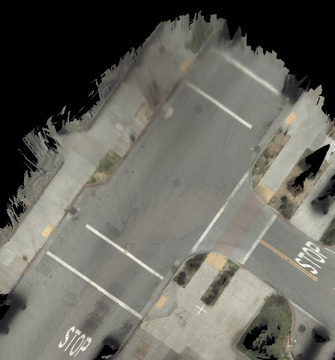} &
\includegraphics[width=.1595\textwidth]{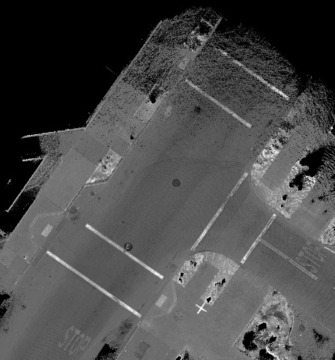} &
\includegraphics[width=.1595\textwidth]{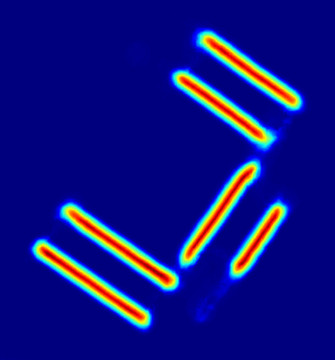} &
\includegraphics[width=.1595\textwidth]{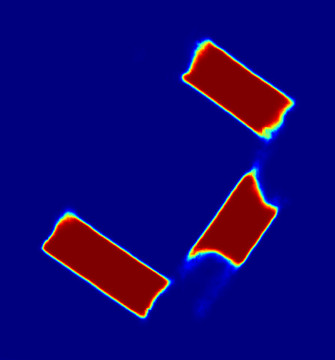} &
\includegraphics[width=.1595\textwidth]{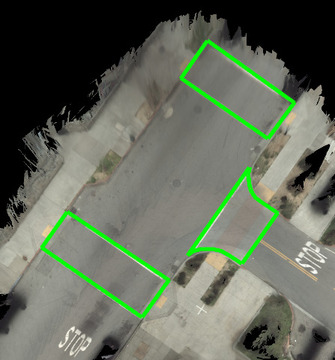} &
\includegraphics[width=.1595\textwidth]{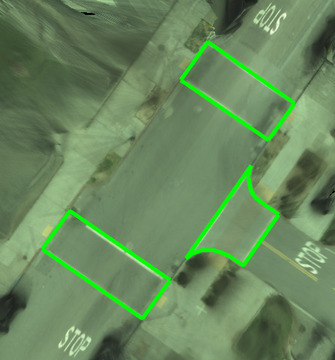} &\\

a) & b) & c) & d) & e) & f) & \\
\end{tabular}

\caption{Online map model (single pass) results using the model trained on both camera and LiDAR imagery. Comparisons between col a) input ground camera (online map), b) input ground LiDAR (online map), c) predicted inverse distance transform, d) predicted segmentation, e) our predicted crosswalk polygons after inference and f) ground truth crosswalk polygons overlayed on the ground RGB (offline map).}
\label{fig:online_2}

\end{figure}

We visualize more qualitative results of our structured prediction model for drawing crosswalks. In particular:

\begin{enumerate}
  \item In Figures \ref{fig:offline_1} and \ref{fig:offline_2}, we show more examples of our model trained on both ground camera and LiDAR imagery from the offline maps (multiple passes). 
  \item In Figures \ref{fig:online_1} and \ref{fig:online_2}, we show more examples of our model trained on both ground camera and LiDAR imagery from the online maps (single pass). 
\end{enumerate}

\end{document}